\documentclass[journal]{IEEEtran}
\usepackage{graphicx}
\usepackage{amsmath,amssymb} 
\usepackage{color}
\usepackage{times}
\usepackage{epsfig}
\usepackage{graphicx}
\usepackage{amsmath}
\usepackage{amssymb}
\usepackage{epstopdf}
\usepackage{algorithm}
\usepackage{algorithmic}
\usepackage{subfigure}
\usepackage{booktabs}
\usepackage{bm}
\usepackage{url}
\ifCLASSINFOpdf
\else
\fi
\newcommand{\twosmallfigwidth}{41}

\newcommand{\fourfigwidth}{42}

\newcommand{\tablecolumnwidth}{38}
\newcommand{\smalltablecolumnwidth}{12}
\newcommand{\mediumtablecolumnwidth}{25}

\DeclareMathOperator{\tr}{tr}

\DeclareMathOperator{\sign}{sign}
\DeclareMathOperator{\chol}{chol}

\newcommand{\fMRItwofigwidths}{40}
\newcommand{\fMRIfigheights}{40}

\newcommand{\figwidths}{28}
\newcommand{\figheights}{25}

\newcommand{\twomedfigwidths}{90}

\begin{document}
%
\title{Learning Discriminative Stein Kernel for SPD Matrices and Its Applications}
%
%
%
\author{Authors
\thanks{Authors Info}}
\author{Jianjia~Zhang,
        Lei~Wang,
        Luping~Zhou,
        and Wanqing Li
\thanks{Jianjia~Zhang, Lei~Wang, Luping Zhou, and Wanqing Li are with the School of Computing and Information Technology. University of Wollongong, Wollongong 2522 Australia (e-mail: jz163@uowmail.edu.au, \{leiw, lupingz, wanqing\}@uow.edu.au).}}

\maketitle

\begin{abstract}
Stein kernel has recently shown promising performance on classifying images represented by symmetric positive definite (SPD) matrices. It evaluates the similarity between two SPD matrices through their eigenvalues. In this paper, we argue that directly using the original eigenvalues may be problematic because: i) Eigenvalue estimation becomes biased when the number of samples is inadequate, which may lead to unreliable kernel evaluation; ii) More importantly, eigenvalues only reflect the property of an individual SPD matrix. They are not necessarily optimal for computing Stein kernel when the goal is to discriminate different classes of SPD matrices. To address the two issues, we propose a discriminative Stein kernel, in which an extra parameter vector is defined to adjust the eigenvalues of input SPD matrices. The optimal parameter values are sought by optimizing a proxy of classification performance. To show the generality of the proposed method, three kernel learning criteria that are commonly used in the literature are employed respectively as a proxy. A comprehensive experimental study is conducted on a variety of image classification tasks to compare the proposed discriminative Stein kernel with the original Stein kernel and other methods for evaluating the  similarity between SPD matrices. The results demonstrate that the discriminative Stein kernel can attain greater discrimination and better align with classification tasks by altering the eigenvalues. This makes it produce higher classification performance than the original Stein kernel and other commonly used methods. 
\end{abstract}


%
\IEEEpeerreviewmaketitle

\section{Introduction}\label{Introduction}

%
%
%
%
\IEEEPARstart{I}{n} the past several years, symmetric positive definite (SPD) matrices have been widely applied to pattern analysis and computer vision. For example, covariance descriptor~\cite{rc2006} is used in texture classification~\cite{ts2010,sc2012}, face recognition~\cite{gr2008}, action recognition~\cite{ar2013,ar2010}, pedestrian detection~\cite{pd2008,fp2008}, visual tracking~\cite{an2013} and 3D shape matching/retrieval~\cite{CovarianceHedi}. Also, based on functional magnetic resonance imaging, a SPD correlation matrix is used to model brain networks to discriminate patients with Alzheimer's disease from the healthy~\cite{wee2012resting}. The tensors obtained in diffusion tensor imaging (DTI)~\cite{basser1994mr} are SPD matrices too. Since the data in these tasks are either represented by or converted to SPD matrices, they are classified by the classification of the associated SPD matrices. 

A fundamental issue in the classification of SPD matrices is how to measure their similarity. Essentially, when a sufficiently good similarity measure is available, a simple $k$-nearest neighbor classifier will be able to achieve excellent classification performance. 
Considering that SPD matrices reside on a Riemannian manifold~\cite{km2013}, commonly used Euclidean-based measures may not be effective since they do not take the manifold structure into account. To circumvent this problem, affine-invariant Riemannian metric (AIRM) has been proposed in~\cite{forstner2003metric} for comparing  covariance matrices, and used in \cite{fletcher2004principal,lenglet2004statistics,pennec2006riemannian} for statistical analysis of DTI data. Although improving similarity measurement, AIRM involves matrix inverse and square rooting~\cite{lm2006}, which could result in high computational cost when the dimensions of SPD matrices are high. 
In the past decade, effectively and efficiently measuring the similarity between SPD matrices on the Riemannian manifold has attracted much attention. One attempt is to map the manifold to a Euclidean space~\cite{pd2008,veeraraghavan2005matching,fiori2012extended}, i.e. the tangent space at the mean point. However, 
these approaches suffer from two limitations: i) Mapping the points from manifold to the tangent space or vice-versa is also computationally expensive; ii) More importantly, the tangent space is merely a local approximation of the manifold at the mean point, leading to a suboptimal solution. 

To address these limitations, an alternative approach is to embed the manifold into a high-dimensional Reproducing Kernel Hilbert Space (RKHS) by using kernels. This can bring at least three advantages: i) The computational cost can be reduced by selecting an efficient kernel; ii) The manifold structure can be well incorporated in the embedding; iii) Many Euclidean algorithms, e.g. support vector machines (SVM), can be readily used.
Several kernel functions for SPD matrices have been defined via SPD-matrix-based distance functions~\cite{km2013}, including Log-Euclidean distance~\cite{lm2006,vemulapalli2013kernel}, Cholesky distance~\cite{ns2009}, and Power Euclidean distance~\cite{ns2009}. 
Recently, another kernel function called \textit{Stein kernel} has been proposed in~\cite{pd2011} with promising applications reported in~\cite{sc2012,salehianrecursive,alavi2013relational}. 
 Specifically, in~\cite{sc2012}, Stein kernel is applied to evaluate the similarity of covariance descriptors for image classification. As shown in that work, Stein kernel is able to achieve better performance than 
other kernels in texture classification, face recognition and person re-identification.

Stein kernel evaluates the similarity of SPD matrices through their eigenvalues. Although this is theoretically rigorous and elegant,  directly using the original eigenvalues may encounter the following two issues in practice. 
\begin{itemize}
\item \textit{Issue-I.} Some SPD matrices, e.g.  covariance descriptors~\cite{rc2006}, have to be estimated from a set of samples. Nevertheless, covariance matrix estimation is sensitive to the number of samples, and eigenvalue estimation becomes biased when the number of samples is inadequate~\cite{ec2008}. Such a biasness will affect Stein kernel evaluation. 
\item \textit{Issue-II.} Even if true eigenvalues could be obtained, when the goal is to discriminate different sets of SPD matrices, computing Stein kernel with these eigenvalues is not necessarily optimal. This is because eigenvalues only reflect the intrinsic property of an individual matrix and the eigenvalues of all the involved matrices have not been collectively manipulated toward greater discrimination.
\end{itemize}

In this paper, we propose a \textit{discriminative} Stein kernel (DSK in short) to address the two issues. Specifically, assuming that the eigenvalues of each SPD matrix have been sorted in a given order, a parameter is assigned to each eigenvalue to adjust its magnitude. Treating these parameters as extra parameters of Stein kernel, we automatically learn their optimal values by using the training samples of a classification task. Our work brings forth two advantages: i) Although not restoring the unbiased eigenvalue estimates\footnote{Note that the proposed method is not designed for (and is not even interested in) restoring the unbiased estimates of eigenvalues. Instead, it aims to achieve better classification when the above two issues are present.}, DSK mitigates the negative impact of the biasness to class discrimination; ii) By adaptively learning the adjustment parameters from training data, it makes Stein kernel better align with specific classification tasks. Both advantages help to boost the classification performance of Stein kernel in practical applications. Three kernel learning criteria, including kernel alignment~\cite{os2002}, class separability~\cite{wang2008feature}, and the radius margin bound~\cite{chapelle2002choosing}, are employed to optimize the adjustment parameters, respectively. The proposed DSK is experimentally compared with the original Stein kernel on a variety of classification tasks. As demonstrated, it not only leads to consistent improvement when the samples for SPD matrix estimation are relatively limited, but also outperforms the original Stein kernel when there are enough samples. 

\section{Related Work}\label{Sec:related-work}
The set of SPD matrices with the size of  $d\times{d}$ can be defined as $\mathrm{Sym}_{d}^{+} = \{{\bm A} |{\bm A}= {\bm A}^{\top}, \forall {\bm x} \in {\mathbb R}^{d}, {\bm x} \neq {\bm 0}, {\bm x}^{\top}{\bm A}{\bm x} > 0\}$. SPD matrices arise in various pattern analysis and computer vision tasks. Geometrically, SPD matrices form a convex half-cone in the vector space of matrices and the cone constitutes a  Riemannian manifold. A Riemannian manifold is a real smooth manifold that is differentiable and equipped with a smoothly varying inner product for each tangent space. The family of the inner products is referred to as a Riemannian metric. The special manifold structure of SPD matrices is of great importance in analysis and optimization~\cite{pd2011}. 

A commonly encountered example of SPD matrices is the covariance descriptor~\cite{rc2006} in computer vision.  Given a set of feature vectors $\{\bm{x}_i; \bm{x}_i \in  \mathbb{R}^d\}_{i = 1}^{n}$ extracted from an image region, this region can be represented by a $d \times d$ covariance matrix $\bm X$, where $\bm X$ is defined as $\frac{1}{n-1} \sum_{i = 1}^{n}(\bm{x}_i - \boldsymbol\mu)(\bm{x}_i - \boldsymbol \mu)^\top$ and $\boldsymbol\mu$ is defined as $\frac{1}{n} \sum_{i = 1}^{n}\bm{x}_i$. Using a covariance matrix as a region descriptor gives several advantages. For instance, the covariance matrix can fuse all kinds of features in a natural way. Also, the size of the covariance matrix is independent of the size of the region and the number of features extracted, inducing a certain scale and rotation invariance over regions.

Let ${\bm X}$ and ${\bm Y}$ be two SPD matrices. How to measure the similarity between ${\bm X}$ and ${\bm Y}$ is a fundamental issue in SPD data processing and analysis. Recent years have seen extensive work on this issue. Respecting the Riemannian manifold, one widely used Riemannian metric is the affine-invariant Riemannian metric (AIRM)~\cite{pennec2006riemannian}, which is defined as 
\begin{equation}
d(\bm{X}, \bm{Y}) = \| \log(\bm{X}^{-\frac{1}{2}} \cdot \bm{Y} \cdot \bm{X}^{-\frac{1}{2}}) \|_{F}
\end{equation}
where $\log (\cdot)$ represents the matrix logarithm and $\parallel \cdot \parallel_{F}$ is the Frobenius norm.
The computational cost of AIRM could be high due to the use of matrix inverse and square rooting. Some other methods directly map SPD matrices into Euclidean spaces to utilize linear algorithms~\cite{ar2010,gd2011}. However, they fail to take full advantage of the geometry structure of Riemannian manifold.

To address these drawbacks, kernel based methods have been generalized to handle SPD data residing on a manifold. A point $\bm{X}$ on a manifold  $\mathcal{M}$  is mapped to a feature vector $\phi({\bm{X}})$ in some feature space $\mathcal{F}$. The mapping is usually implicitly induced by a kernel function $k: (\mathcal{M},\mathcal{M}) \rightarrow \mathbb{R}$, which defines the inner product in $\mathcal{F}$, i.e. $k({\bm{X}_i},{\bm{X}_j}) = \langle \phi({\bm{X}_i}), \phi({\bm{X}_j}) \rangle$. Besides allowing linear algorithms to be used in $\mathcal{F}$, kernel functions are often efficient to compute. 
A family of Riemannian metrics and the corresponding kernels are listed in Table \ref{tb:spdmetric}. 
\begin{table*}[!ht]
\caption{Definitions and properties of the metrics on $\mathrm{Sym}_{d}^{+}$.}
\label{tb:spdmetric} \centering
\begin{tabular}{ p{80pt}p{170pt}p{90pt}p{50pt}p{60pt}}
\toprule
 Metric name &{Formula (denoted as $d$)}&Does $k = \exp(- \theta \cdot d^2)$ define a valid kernel?&{Kernel abbr. in the paper}&Time complexity\\
\midrule
AIRM~\cite{pennec2006riemannian}&$\| \log(\bm{X}_1^{-\frac{1}{2}} \cdot \bm{X}_2 \cdot \bm{X}_1^{-\frac{1}{2}}) \|_{F}$&No &N.A.&$\mathcal O(d^3)$\\
\midrule
Cholesky~\cite{ns2009}&$\| \chol(\bm{X}_1) -  \chol(\bm{X}_2)\|_{F}$&Yes&CHK&$\mathcal O(d^3)$\\
\midrule
Euclidean~\cite{ns2009}&$\| \bm{X}_1 -  \bm{X}_2\|_{F}$&Yes &EUK&$\mathcal O(d^2)$\\
\midrule
Log-Euclidean~\cite{lm2006}&$\| \log(\bm{X}_1) -  \log(\bm{X}_2)\|_{F}$&Yes &LEK&$\mathcal O(d^3)$\\
\midrule
Power-Euclidean~\cite{ns2009}&$\frac{1}{\zeta}\|\bm{X}_1^\zeta - \bm{X}_2^\zeta \|_{F}$&Yes &PEK&$\mathcal O(d^3)$\\
\midrule
S-Divergence root~\cite{pd2011}&$\left[\log\left(\det\left(\frac{{\bm X_1} + {\bm X_2}}{2}\right)\right) - \frac{1}{2} \log\left(\det({\bm X_1}{\bm X_2})\right)\right]^\frac{1}{2}$&Yes ($\theta\in{\boldsymbol\Theta}$) &SK&$\mathcal O(d^{2.373})$\\
\bottomrule
\end{tabular}
\end{table*}
 A recently proposed Stein kernel~\cite{pd2011} has demonstrated notable improvement on discriminative power in a variety of applications~\cite{sc2012,salehianrecursive,alavi2013relational,harandi2014bregman}. It is expressed as
\begin{equation}
k({\bm X},{\bm Y}) = \exp\left(-\theta{\cdot}S\left({\bm X},{\bm Y}\right)\right) \end{equation} where $\theta$ is a tunable positive scalar. $S({\bm X},{\bm Y})$ is called S-Divergence and it is defined as
\begin{equation}  
 S({\bm X},{\bm Y}) = \log\left(\det\left(\frac{{\bm X} + {\bm Y}}{2}\right)\right) - \frac{1}{2} \log\left(\det({\bm X}{\bm Y})\right),
\label{sd}
\end{equation}where $\det(\cdot)$ denotes the determinant of a square matrix. The S-Divergence has several desirable  properties.  For example, i) It is invariant to affine transformations, which is important for computer vision algorithms~\cite{harandi2014bregman}; ii) The square-root of S-Divergence is proven to be a metric on $\mathrm{Sym}_{d}^{+}$~\cite{pd2011}; iii) Stein kernel is guaranteed to be a Mercer kernel when $\theta$ varies within the range of ${\boldsymbol\Theta} = \{\frac{1}{2},\frac{2}{2},\frac{3}{2},\cdots,\frac{(d-1)}{2}\}\cup(\frac{(d-1)}{2},+\infty)$. Readers are referred to \cite{pd2011} for more details. In general, 
S-Divergence enjoys higher computational efficiency than AIRM while well maintaining its measurement performance. When compared to other SPD metrics, such as Cholesky distance~\cite{ns2009}, Euclidean distance~\cite{ns2009}, Log-Euclidean distance~\cite{lm2006,vemulapalli2013kernel},  and Power Euclidean distance~\cite{ns2009}, S-Divergence usually helps to achieve better classification performance~\cite{sc2012}. All the metrics in Table \ref{tb:spdmetric} will be compared in the experimental study.

\section{The proposed method}
\subsection{Issues with Stein Kernel}\label{Limitations}\label{subsec:issues}
Let $\lambda_{i}({\bm X})$ denote the $i^{th}$ eigenvalue of a SPD matrix ${\bm X}$, where $\lambda_{i}({\bm X})$ is always positive due to the SPD property. Throughout this paper, we assume that the $d$ eigenvalues have been sorted in descending order. Noting that the determinant of ${\bm X}$ equals $\prod_{i=1}^{d}\lambda_{i}({\bm X})$, the S-Divergence in Eq.~(\ref{sd}) can be rewritten as
\begin{eqnarray} \label{eq:sd2}
\begin{split}
S(\bm{X},\bm{Y}) =&\sum_{i = 1}^d \log\lambda_i\left(\frac{\bm{X} + \bm{Y}}{2}\right)\\
& - \frac{1}{2}\sum_{i = 1}^d \left(\log\lambda_i(\bm{X})+ \log\lambda_i(\bm{Y})\right).
\end{split}
\end{eqnarray}
We can easily see the important role of eigenvalues in computing $S(\bm{X},\bm{Y})$. Inappropriate eigenvalues will affect the precision of S-Divergence and in turn the Stein kernel. 


\textit{On Issue-I.} It has been well realized in the literature that the eigenvalues of sample-based covariance matrix are biased estimates of true eigenvalues \cite{ea1986}, especially when the number of samples is small. Usually, the smaller eigenvalues tend to be underestimated while the larger eigenvalues tend to be overestimated. To better show this case, we conduct a toy experiment. A set of $n$ data is sampled from a $40$-dimensional normal distribution $\mathcal{N}({\bm 0} ,{\boldsymbol\Sigma})$, where the covariance matrix ${\boldsymbol\Sigma}$ is defined as $\mathrm{diag}(1,2,\cdots,40)$ and the true eigenvalues of ${\boldsymbol\Sigma}$ are just the diagonal entries. The $n$ data are used to estimate ${\boldsymbol\Sigma}$ and calculate the eigenvalues. When $n$ is $100$, the largest eigenvalue is obtained as $67$ while the smallest one is $0.4$, which are poor estimates. When $n$ increases to $1000$, the largest eigenvalue is still overestimated as $46$. From our observation, tens of thousands of samples are required to achieve sufficiently good eigenvalue estimates. Note that the dimensions of $40$ are common in practice. A covariance descriptor of $43$-dimensional features is used in~\cite{sc2012} for face recognition, and it is also used in our experimental study.  

\textit{On Issue-II.} As previously mentioned, even if true eigenvalues could be obtained, a more important issue exists when the goal is to classify different sets of SPD matrices. In specific, a SPD matrix can be expressed as
\begin{displaymath}
\bm{X} = \lambda_1 \bm{u}_1  \bm{u}_1^\top + \lambda_2 \bm{u}_2  \bm{u}_2^\top + \cdots + \lambda_d \bm{u}_d  \bm{u}_d^\top,
\end{displaymath}where $\lambda_i$ and $\bm{u}_i$ denote the $i^{th}$ eigenvalue and the corresponding eigenvector. The magnitude of $\lambda_i$ only reflects the property of this specific SPD matrix, for example, the data variance along the direction of $\bm{u}_i$. It does not characterize this matrix from the perspective of discriminating different sets of SPD matrices. We know that, by fixing the $d$ eigenvectors, varying the eigenvalues changes the matrix ${\bm X}$. Geometrically, a SPD matrix corresponds to a hyper-ellipsoid in a $d$-dimensional Euclidean space. This change is analogous to varying the lengths of the axes of the hyper-ellipsoid while maintaining their directions. A question then arises: to make the Stein kernel better prepared for class discrimination, \textit{can we adjust the eigenvalues to make the SPD matrices in the same class similar to each other, as much as possible, while maintaining the SPD matrices across classes to be sufficiently different?} The ``similar'' and ``different'' are defined in the sense of Stein kernel. This idea can also be understood in the other way. An ideal similarity measure shall be more sensitive to inter-class difference and less affected by intra-class variation. Without exception, this shall apply to Stein kernel too. 
\subsection{Proposed Discriminative Stein Kernel (DSK)} \label{secDSK}
Let $\boldsymbol{\alpha} = [\alpha_1, \alpha_2, \cdots, \alpha_d]^{\top}$ be a vector of adjustment parameters. Let $\bm{X} = \bm{U}\bm{\Lambda} \bm{U^\top}$ denote the eigen-decomposition of a SPD matrix, where the columns of ${\bm U}$ correspond to the eigenvectors and $\bm{\Lambda}=\mathrm{diag}(\lambda_1,\cdots,\lambda_d)$. We use $\boldsymbol{\alpha}$ in two ways for eigenvalue adjustment and define the adjusted $\bm{X}$, respectively, as:
\begin{equation} \label{eq:mmatrixp}
\begin{split}
\tilde{\bm{X}}_{p}&=\bm{U}
\left( \begin{array}{cccc}
\lambda_1^{\alpha_1}  & & &\\
  & \lambda_2^{\alpha_2}& &\\
  & &\ddots &\\
  & & &\lambda_d^{\alpha_d}\\
\end{array} \right)
\bm{U^\top}
\end{split}
\end{equation} 
\begin{equation} \label{eq:mmatrixc}
\begin{split}
~\mathrm{and}~~
\tilde{\bm{X}}_{c}& = \bm{U}
\left( \begin{array}{cccc}
{\alpha_1}\lambda_1  & & &\\
& {\alpha_2} \lambda_2& &\\
  & &\ddots &\\
  & & &{\alpha_d}\lambda_d\\
\end{array} \right)
\bm{U^\top}.
\end{split}
\end{equation}In the first way, $\boldsymbol{\alpha}$ is used as the \textit{power} of eigenvalues. It can naturally maintain the SPD property because $\lambda_{i}^{\alpha_{i}}$ is always positive. In the second way, $\boldsymbol{\alpha}$ is used as the \textit{coefficient} of eigenvalues. It is mathematically simpler but needs to impose the constraint $\alpha_{i}>0~(i=1,\cdots,d)$ to maintain the SPD property. The two adjusted matrices are denoted by $\tilde{\bm{X}}_{p}$ and $\tilde{\bm{X}}_{c}$, where $p$ and $c$ are short for ``power'' and ``coefficient''. Both ways will be investigated in this paper. 

Given two SPD matrices ${\bm X}$ and ${\bm Y}$, we define the ${\boldsymbol{\mathbb{\alpha}}}$-adjusted S-Divergence as 
\begin{equation}
S_{\boldsymbol{{\alpha}}}({\bm{X}},{\bm{Y}})\triangleq{S(\tilde{\bm{X}},\tilde{\bm{Y}})}.
\end{equation} For the two ways of using ${\boldsymbol{\mathbb{\alpha}}}$, the term ${S(\tilde{\bm{X}},\tilde{\bm{Y}})}$ can be expressed as
\begin{eqnarray*} \label{eq:msd-p}
\begin{split}
S(\tilde{\bm{X}}_p,\tilde{\bm{Y}}_p)
&=\sum_{i = 1}^d \log\lambda_i\left( \frac{\tilde{\bm{X}}_p + \tilde{\bm{Y}}_p}{2}\right)\\
& - \frac{1}{2}\sum_{i = 1}^d \alpha_{i}\left(\log\lambda_i(\bm{X})+ \log\lambda_i(\bm{Y})\right)
\end{split}
\end{eqnarray*}
\begin{eqnarray*} \label{eq:msd-c}
\begin{split}
~\text{and}~~S(\tilde{\bm{X}}_c,\tilde{\bm{Y}}_c)
&=\sum_{i = 1}^d \log\lambda_i\left( \frac{\tilde{\bm{X}}_c + \tilde{\bm{Y}}_c}{2}\right) \\
&- \frac{1}{2}\sum_{i = 1}^d \left(2\log\alpha_{i}+\log\lambda_i(\bm{X})+ \log\lambda_i(\bm{Y})\right).
\end{split}
\end{eqnarray*}Based on the above definition, the discriminative Stein kernel (DSK) is proposed as
\begin{eqnarray} \label{eq:msk}
k_{\boldsymbol{{\alpha}}}({\bm{X}},{\bm{Y}}) =  \exp\left(-\theta\cdot S_{\boldsymbol{{\alpha}}}\left({\bm{X}},{\bm{Y}}\right)\right). 
\end{eqnarray}Note that the DSK will remain a Mercer kernel as long as $\theta$ varies in the range of ${\boldsymbol\Theta}$ defined in Section~\ref{Sec:related-work}, because $k_{\boldsymbol{{\alpha}}}({\bm{X}},{\bm{Y}})$ can always be viewed as $k(\tilde{\bm{X}},\tilde{\bm{Y}})$, the original Stein kernel applied to two adjusted SPD matrices $\tilde{\bm{X}}$ and $\tilde{\bm{Y}}$.

Treating $\boldsymbol{{\alpha}}$ as the kernel parameter of $k_{\boldsymbol{{\alpha}}}({\bm{X}},{\bm{Y}})$, we resort to kernel learning techniques to find its optimal value.
Kernel learning methods have received much attention in the past decade. Many learning criteria such as kernel alignment~\cite{os2002}, kernel class separability \cite{wang2008feature}, and radius margin bound \cite{chapelle2002choosing} have been proposed. In this work, to investigate the generality of the proposed DSK, we employ all the three criteria, respectively, to solve the kernel parameters $\boldsymbol{{\alpha}}$. 

Let $\boldsymbol\Omega = \{\left(\bm{X}_i, t_{i}\right)\}_{i=1}^{n}$ be a set of $n$ training SPD matrices, each of which represents a sample, e.g., an image to be classified. $t_{i}$ denotes the class label of the $i^{th}$ sample, where $t_{i}\in\{1,\cdots,M\}$ with $M$ denoting the number of classes. $\bm{K}$ denotes the kernel matrix computed with DSK on $\boldsymbol\Omega$, with $\bm{K}_{ij} = k_{\boldsymbol{{\alpha}}}({\bm{X}_i},{\bm{X}_j})$. In the following part, three frameworks are developed to learn the optimal value of  $\boldsymbol{{\alpha}}$. 

\subsubsection{Kernel Alignment based Framework}
Kernel alignment measures the similarity of two kernel functions and can be used to quantify  the degree of agreement between a kernel and a given classification task~\cite{os2002}. Kernel alignment possesses several desirable properties, including conceptual simplicity, computational efficiency, concentration of empirical estimate, and theoretical guarantee for generalization performance~\cite{os2002,aw2012}. Furthermore, kernel alignment is a general-purpose criterion that does not depend on a specific classifier and often leads to simple optimization. Also, it can uniformly handle binary and multi-class classification. Due to these merits, the kernel alignment criterion has been widely used in kernel-related learning tasks~\cite{aw2012}, including kernel parameter tuning~\cite{zhong2013optimizing}, multiple kernel learning~\cite{gonen2011multiple},  spectral kernel learning~\cite{mao2012parameter} and feature selection~\cite{ramona2012multiclass}. 

 With the kernel alignment, the optimal $\boldsymbol{\alpha}$ can be obtained through the following optimization problem:
\begin{equation} \label{eq:ka}
\boldsymbol{{\alpha}}^*=\arg\max_{{\boldsymbol{\alpha}}\in{\mathcal A}} J(\bm{K}, {\bm T}) - \lambda\|\boldsymbol{\alpha} - \boldsymbol{{\alpha}}_{0}\|_2^2,
\end{equation}
where ${\bm T}$ is an $n\times{n}$ matrix with ${\bm T}_{ij}=1$ if ${\bm X}_i$ and ${\bm X}_j$ are from the same class and ${\bm T}_{ij}=-1$ otherwise. Note that this definition of ${\bm T}$ naturally handles multi-class classification. $J(\bm{K}, {\bm T})$ is defined as the kernel alignment criterion:
\begin{equation}\label{eq:kab}
J(\bm{K}, {\bm T}) = \frac{\langle{\bm T}, \bm{K}\rangle_{F}}{\sqrt{\langle {\bm T}, {\bm T}\rangle_{F} \langle \bm{K}, \bm{K}\rangle}_{F}} 
\end{equation}
where $\langle\cdot, \cdot\rangle_F$ denotes the Frobenius inner product between two matrices. $J(\bm{K}, {\bm T})$ measures the degree of agreement between ${\bm K}$ and ${\bm T}$, where ${\bm T}$ is regarded as the ideal kernel of a learning task. 
The $\boldsymbol{{\alpha}}_{0}$ is a priori estimate of $\boldsymbol{{\alpha}}$, and $\|\boldsymbol{\alpha} - \boldsymbol{{\alpha}}_{0}\|_2^2$ is the regularizer which constrains $\boldsymbol{\alpha}$ to be around $\boldsymbol{\alpha}_0$ to avoid overfitting. We can simply set $\boldsymbol{{\alpha}}_{0}=[1,\cdots,1]^{\top}$, which corresponds to the original Stein kernel. $\lambda$ is the regularization parameter to be selected via cross-validation. ${\mathcal A}$ denotes the domain of $\boldsymbol{\alpha}$: when $\boldsymbol{\alpha}$ is used as a power, ${\mathcal A}$ denotes a Euclidean space ${\mathbb R}^{d}$; when $\boldsymbol{\alpha}$ is used as a coefficient, ${\mathcal A}$ is constrained to ${\mathbb R}^{d}_{+}$.

 $J(\bm{K}, {\bm T})$ is differentiable with respect to $\bm{K}$ and $\boldsymbol{{\alpha}}$:
\begin{equation}\label{eq:kabd}
\begin{aligned}
\frac{\partial J(\bm{K}, {\bm T})}{\partial \alpha_z} = \frac{\langle{\bm T}, \frac{\partial\bm{K}}{\partial \alpha_z}\rangle_{F}}{\sqrt{\langle {\bm T}, {\bm T}\rangle_{F} \langle \bm{K}, \bm{K}\rangle}_{F}}
 - \frac{\langle{\bm T}, \bm{K}\rangle_{F}\langle{\bm K}, \frac{\partial\bm{K}}{\partial \alpha_z}\rangle_{F}}{\sqrt{\langle {\bm T}, {\bm T}\rangle_{F}} \langle \bm{K}, \bm{K}\rangle_{F}^{3/2}}
\end{aligned}
\end{equation}
where ${\alpha_z}$ denotes the $z^{th}$ parameter of $\boldsymbol{{\alpha}}$ and the entry of {\small{$\frac{\partial\bm{K}}{\partial \alpha_z}$ is $\frac{\partial{k_{\boldsymbol{{\alpha}}}({\bm{X}},{\bm{Y}})}}{\partial \alpha_z}$}}. Based on Eq.~(\ref{eq:msk}), it can be calculated  as\footnote{The detailed derivation can be found in the supplementary material.}

{\small{$\frac{\partial{k_{\boldsymbol{{\alpha}}}({\bm{X}},{\bm{Y}})}}{\partial \alpha_z} =\frac{\theta k_{\boldsymbol{{\alpha}}}({\bm{X}},{\bm{Y}})}{2}  \tr\Big[ {{\tilde{\bm X}}}^{-1} \left( \frac{\partial{\tilde{\bm X}}}{\partial \alpha_z}\right) +  {{\tilde{\bm Y}}}^{-1} \left( \frac{\partial{\tilde{\bm Y}}}{\partial \alpha_z}\right) - \left( \frac{{\tilde{\bm X}} + {\tilde{\bm Y}}}{2} \right)^{-1} \left( \frac{\partial{\tilde{\bm X}}}{\partial \alpha_z} + \frac{\partial{\tilde{\bm Y}}}{\partial \alpha_z}\right)\Big]$}}, where $\tr(\cdot)$ denotes the trace of a matrix.
Therefore, any gradient-based optimization technique can be applied to solve the optimization problem in Eq.~(\ref{eq:ka}).
\subsubsection*{On choosing $\theta$} As seen in Eq.~(\ref{eq:msk}), there is a kernel parameter $\theta$ inherited from the original Stein kernel. Note that $\theta$ and ${\boldsymbol\alpha}$ play different roles in the proposed kernel and cannot be replaced with each other. The value of $\theta$ needs to be appropriately chosen because it impacts the kernel value and in turn the optimization of $\boldsymbol{\mathbb{\alpha}}$. A commonly used way to tune $\theta$ is $k$-fold cross-validation. In this paper, to better align with the kernel alignment criterion, we also tune $\theta$ by maximizing the kernel alignment and do this before adjusting $\boldsymbol{\alpha}$,
\begin{equation} \label{eq:ka-theta}
\theta^{*} = \arg\max_{\theta\in{\boldsymbol\Theta}} J(\bm{K}|_{\boldsymbol{\alpha} = \bm{1}}, {\bm T}). 
\end{equation}
where $\bm{1}$ is a $d$-dimensional vector with all entries equal to $1$ and $\bm{K}|_{\boldsymbol{\alpha} = \bm{1}}$ denotes the kernel matrix computed by the original Stein kernel without $\boldsymbol\alpha$-adjustment. Through this optimization, we find a reasonably good $\theta$ and then optimize $\boldsymbol\alpha$ on top of it. The maximization problem in Eq.~(\ref{eq:ka-theta}) can be conveniently solved by choosing $\theta$ in the range of $\boldsymbol\Theta=\left\{ \frac{1}{2},\frac{2}{2},\frac{3}{2},\cdots , \frac{d - 1}{2} \right\}   \cup  \left(\frac{d-1}{2},+\infty \right)$. $\theta$ is not optimized jointly with $\boldsymbol \alpha$ since the noncontinuous range of $\theta$ could complicate the gradient-based optimization. As will be shown in the experimental study, optimizing $\theta$ and $\boldsymbol \alpha$ sequentially can already lead to promising results.

After obtaining $\theta^*$ and $\boldsymbol \alpha^*$, the proposed DSK will be applied to both training and test data for classification, with certain classifiers such as $k$-nearest neighbor ($k$-NN) or SVM. Note that for a given classification task, the optimization of $\theta$ and $\boldsymbol \alpha$ only needs to be conducted \textit{once} with training data. After that, they are used as fixed parameters to compute the Stein kernel for each pair of SPD matrices. The DSK with kernel alignment criterion is outlined in Algorithm \ref{alg:1}.
\begin{algorithm}[htb] %
\renewcommand{\algorithmicrequire}{\textbf{Input:}}
\renewcommand\algorithmicensure {\textbf{Output:} }
\caption{Proposed discriminative Stein kernel learning with the kernel alignment criterion} %
\label{alg:1} %
\begin{algorithmic}[1] %
\REQUIRE 
A training sample set $\boldsymbol\Omega = \{(\bm{X}_i,t_{i})\}_{i=1}^{n}$, $\boldsymbol{{\alpha}}_{0}$ and $\lambda$.

\ENSURE 
$\theta^*$,~~$\boldsymbol{\alpha}^*$~~;\\
\STATE Find $\theta^* = \arg\max_{\theta\in{\boldsymbol\Theta}} J(\bm{K}|_{\boldsymbol{\alpha} = \bm{1}}, {\bm T})$ first to obtain $\theta^*$;
\STATE Learn
$\boldsymbol{{\alpha}}^*=\arg\max_{{\boldsymbol{{\alpha}}}\in{\mathcal A}} J(\bm{K}|_{\theta = \theta^*}, {\bm T}) - \lambda\|\boldsymbol{\alpha} - \boldsymbol{{\alpha}}_{0}\|_2^2$;

\RETURN $\theta^*$, $\boldsymbol{\alpha}^*$; %
\end{algorithmic}
\end{algorithm}
\subsubsection{Class Separability based Framework}\label{csframe}
Class separability is another commonly used criterion for model and feature selection \cite{wang2008two, wang2008feature, xiong2005optimizing}. Recall that the training sample set is defined as $\boldsymbol\Omega = \{\left(\bm{X}_i, t_{i}\right)\}_{i=1}^{n}$, where $t_{i}\in\{1,\cdots,M\}$. Let $\boldsymbol\Omega_i$ be the set of training samples from the $i^{th}$ class, with $n_i$ denoting the size of $\boldsymbol\Omega_i$.  $\bm{K}_{\boldsymbol\Omega',\boldsymbol\Omega''}$ denotes a kernel matrix computed over two training subsets $\boldsymbol\Omega'$ and $\boldsymbol\Omega''$, where $\{\bm{K}_{\boldsymbol\Omega',\boldsymbol\Omega''}\}_{ij} = k({\bm{X}_i},{\bm{X}_j}) = \langle \phi({\bm{X}_i}), \phi({\bm{X}_j}) \rangle$ with ${\bm{X}_i} \in \boldsymbol\Omega'$ and ${\bm{X}_j} \in \boldsymbol\Omega''$.
The class separability in the feature space $\mathcal{F}$ induced by a kernel $k$ can be defined as
\begin{equation}\label{eq:cs}
J = \frac{\tr(\bm{S}_B)}{\tr(\bm{S}_W)},
\end{equation}
where $\tr(\cdot)$ is the trace of a matrix, and  $\bm{S}_B$ and $\bm{S}_W$ are the \textit{between-class scatter matrix} and the \textit{within-class scatter matrix}, respectively. 
Let $\bm{m}$ and $\bm{m}_i$ denote the total sample mean and the $i^{th}$ class mean.  They can be expressed as $\bm{m} = \frac{1}{n}\sum_{{\bm{X}_i} \in \boldsymbol\Omega}\phi(\bm{X}_i)$ and 
$\bm{m}_i = \frac{1}{n_i}\sum_{{\bm{X}_j} \in \boldsymbol\Omega_i}\phi(\bm{X}_j)$.

 $\tr(\bm{S}_B)$ and $\tr(\bm{S}_W)$ can be expressed  as:
\begin{equation}\label{trsb}
\begin{aligned}
\tr(\bm{S}_B) &= \tr \left[ \sum_{i = 1}^M n_i\left( \bm{m}_i - \bm{m}\right)\left( \bm{m}_i - \bm{m}\right)^\top\right]\\
&= \sum_{i = 1}^M \frac{\bm{1}^\top\bm{K}_{\boldsymbol\Omega_i,\boldsymbol\Omega_i}\bm{1}}{n_i} - \frac{\bm{1}^\top\bm{K}_{\boldsymbol\Omega,\boldsymbol\Omega}\bm{1}}{n},
\end{aligned}
\end{equation}\text{and}~
\begin{equation}\label{trsw}
\begin{aligned}
 \tr(\bm{S}_W) &= \tr \left[ \sum_{i = 1}^M \sum_{j = 1}^{n_i} \Big( \phi\left({\bm{X}_{ij}}\right) - \bm{m}_i\Big)\Big( \phi\left({\bm{X}_{ij}}\right) - \bm{m}_i\Big)^\top\right]\\
&= \tr\left(\bm{K}_{\boldsymbol\Omega,\boldsymbol\Omega}\right) - \sum_{i = 1}^M \frac{\bm{1}^\top\bm{K}_{\boldsymbol\Omega_i,\boldsymbol\Omega_i}\bm{1}}{n_i}
\end{aligned}
\end{equation}
where $\bm{1} = [1, 1, \cdots, 1]^\top$. 
The derivatives of $\tr(\bm{S}_B)$ and $\tr(\bm{S}_W)$ with respect to $\alpha_z$ can be shown as:
\begin{equation}
\begin{aligned}
\frac{\partial\tr(\bm{S}_B)}{\partial\alpha_z} = \sum_{i = 1}^M \frac{\bm{1}^\top\frac{\partial \bm{K}_{\boldsymbol\Omega_i,\boldsymbol\Omega_i}}{\partial\alpha_z}\bm{1}}{n_i} - \frac{\bm{1}^\top\frac{\partial \bm{K}_{\boldsymbol\Omega,\boldsymbol\Omega}}{\partial\alpha_z}\bm{1}}{n},
\end{aligned}
\end{equation}
\begin{equation}
\begin{aligned}
 \text{and}~~
\frac{\partial\tr(\bm{S}_W)}{\partial\alpha_z}= \tr\left(\frac{\partial\bm{K}_{\boldsymbol\Omega,\boldsymbol\Omega}}{\partial\alpha_z}\right) - \sum_{i = 1}^M \frac{\bm{1}^\top\frac{\partial \bm{K}_{\boldsymbol\Omega_i,\boldsymbol\Omega_i}}{\partial\alpha_z}\bm{1}}{n_i}
\end{aligned}
\end{equation}
The class separability can reflect the goodness of a kernel function with respect to a given task. 
The DSK learning procedure outlined in Algorithm \ref{alg:1} can be fully taken advantage to optimize the parameter $\boldsymbol\alpha$ when class separability measure is used. The only modification is to replace the definition of $J$ with Eq. (\ref{eq:cs}).


\subsubsection{Radius Margin Bound based Framework}
Radius margin bound is an upper bound on the number of classification errors in a leave-one-out (LOO) procedure of a hard margin binary SVM~\cite{chapelle2002choosing,keerthi2002efficient}. This bound can be extended to $L_2$-norm soft margin SVM with a slightly modified kernel. It has been widely used for parameter tuning~\cite{chapelle2002choosing} and model selection~\cite{wang2008two}.
We first consider a binary classification task and then extend the result to the multi-class case.
Let $\boldsymbol\Omega' \cup \boldsymbol\Omega''$ be a training set of $l$ samples, and without loss of generality, the samples are labeled by $t \in\{-1, 1\}$.
With a given kernel function $k$, 
the optimization problem of SVM with $L_2$-norm soft margin can be expressed as
\begin{equation}\label{eq:svm}
\begin{aligned}
&\frac{1}{2} \| \bm{w}\|^2 = &&\max_{\boldsymbol{\eta} \in \mathbb{R}^l}  \Big [ \sum_{i = 1}^{l} \eta_i - \frac{1}{2} \sum_{i,j = 1}^{l}\eta_i \eta_j t_{i} t_{j} \tilde{k}({\bm{X}_i},{\bm{X}_j}) \Big]\\
& \text{subject to:} &&  \sum_{i = 1}^{l} \eta_i t_{i} = 0;~ \eta_i\geq 0 ~(i = 1, 2, \dots, l)
\end{aligned}
\end{equation}
where $\tilde{k}({\bm{X}_i},{\bm{X}_j})$ $= {k}({\bm{X}_i},{\bm{X}_j}) + \frac{1}{C}\delta_{ij}$;  
 $C$ is the regularization parameter; $\delta_{ij} = 1$ if $i = j$, and $0$ otherwise; and $\bm{w}$ is the normal vector of the optimal separating hyperplane of SVM.
Tuning the parameters in $\tilde{k}$ can be achieved by minimizing an estimate of the LOO  errors. It is shown in \cite{vapnik1998statistical}  that the following \textit{radius margin bound} holds:
\begin{equation}\label{eloo}
E(\text{LOO}) \leq 4 \cdot \frac{{R}^2}{\gamma^2} =  4 {R}^2\|\bm{w}\|^2,
\end{equation}
where $E(\text{LOO})$ denotes the number of LOO errors performed on the $l$ training samples in ${\boldsymbol\Omega' \cup \boldsymbol\Omega''}$; $R$ is the radius of the smallest sphere enclosing all the $l$ training samples; and $\gamma$ denotes the margin with respect to the optimal separating hyperplane and equals $1/\|\bm{w}\|$. ${R}^2$ can be obtained by the following optimizing problem,
\begin{equation}\label{eq:r2}
\begin{aligned}
{R}^2 &= \max_{\beta \in \mathbb{R}^l} \Big [ \sum_{i = 1}^{l} \beta_i \tilde{k}({\bm{X}_i},{\bm{X}_i}) -  \sum_{i,j = 1}^{l}\beta_i \beta_j \tilde{k}({\bm{X}_i},{\bm{X}_j}) \Big]\\
& \text{subject to:}   ~\sum_{i = 1}^{l} \beta_i = 1;~~ \beta_i\geq 0 ~~(i = 1, 2, \dots, l).
\end{aligned}
\end{equation}
Note that both $R$ and $\bm{w}$ are the function of the kernel $\tilde{k}$. 
We set the kernel function $k$ as $k_{\boldsymbol{{\alpha}}}$ defined in Eq.~(\ref{eq:msk}). The model parameters in $\tilde{k}$, i.e. $\{\theta, \boldsymbol{\alpha},  {C}\}$, can be optimized by minimizing $R^2\|\bm{w}\|^2$ on the training set. As previous, we can first choose a reasonably good $\theta^*$ by  optimizing Eq. (\ref{eq:rmbtheta}) with respect to $\theta$ and $C$ while fixing $\boldsymbol{\alpha}$ as $\bm{1}$.
\begin{equation}\label{eq:rmbtheta}
\begin{aligned}
 \{\theta^*, C^*\} = \arg \min_{\boldsymbol\theta \in \boldsymbol\Theta, {C} >0| \boldsymbol{\alpha} = \bm{1}}~~~  R^2\|\bm{w}\|^2.
\end{aligned}
\end{equation}
Once $\theta^*$ is obtained, $\{\boldsymbol{\alpha},  {C}\}$, denoted by $\boldsymbol \upsilon$, can then be  jointly optimized  as follows:
\begin{equation}\label{eq:rmb}
\begin{aligned}
\boldsymbol \upsilon^* = \arg \min_{\boldsymbol\upsilon \in \boldsymbol\Upsilon | \theta = \theta^*}~~~  R^2\|\bm{w}\|^2,
\end{aligned}
\end{equation}
where $\boldsymbol\Upsilon = \{\boldsymbol{\alpha},  C| \boldsymbol\alpha \in {\mathcal A}; C > 0 \}$. Let $\upsilon_z$ be the $z^{th}$ parameter of $\boldsymbol\upsilon$. The derivative of $ R^2\|\bm{w}\|^2$ with respect to $\upsilon_z$  can be shown as:
\begin{equation}\label{eq:deri}
\frac{\partial(R^2\|\bm{w}\|^2)}{\partial \upsilon_z} = \|\bm{w}\|^2\frac{\partial R^2}{\partial \upsilon_z} + R^2\frac{\partial\|\bm{w}\|^2}{\partial \upsilon_z},
\end{equation}
\begin{equation}\label{eq:dw2}
\text{where}~~~~~~~~~~
\frac{\partial \|\bm{w}\|^2}{\partial \upsilon_z} =  - \sum_{i,j = 1}^{l}\eta_i^* \eta_j^* t_{i} t_{j} \frac{\partial \tilde{k}({\bm{X}_i},{\bm{X}_j})}{\partial \upsilon_z}
\end{equation}
\begin{equation}\label{eq:dr2}
\text{and}~
\frac{\partial R^2}{\partial \upsilon_z} = \sum_{i = 1}^{l} \beta_i^* \frac{\partial \tilde{k}({\bm{X}_i},{\bm{X}_i})}{\partial \upsilon_z} -  \sum_{i,j = 1}^{l}\beta_i^* \beta_j^* \frac{\partial\tilde{k}({\bm{X}_i},{\bm{X}_j})}{\partial \upsilon_z}
\end{equation}
where $\eta_i^*$ and $\beta_i^*$ denote the optimal solutions of Eq. (\ref{eq:svm}) and (\ref{eq:r2}), respectively.

For multi-class classification tasks, we optimize $\boldsymbol \upsilon$ by a pairwise combination of the radius margin bounds of binary SVM classifiers~\cite{wang2008two}. Specifically, an $M$-class classification task can be split into $M(M-1)/2$ pairwise classification tasks by using the one-vs-one strategy.  For any class pair $(i, j)$ with $1\leq i < j \leq M$, a binary SVM classifier, denoted by $\text{SVM}_{ij}$, can be trained using the samples from classes $i$ and $j$. The corresponding radius margin bound, denoted by $R_{ij}^2\|\bm{w}_{ij}\|^2$, can be calculated by Eq. (\ref{eq:svm}) and (\ref{eq:r2}).
As shown in~\cite{wang2008two},  the LOO error of an $M$-class SVM classifier can be upper bounded by the combination of the radius margin bounds of $M(M-1)/2$ pairwise binary SVM classifiers. The combination is defined as:
\begin{equation}\label{eq:svmcomb}
 J = \sum_{1\leq i < j \leq M}R_{ij}^2\|\bm{w}_{ij}\|^2.
\end{equation}
As previous, a reasonably good $\theta^*$ can be firstly chosen by
\begin{equation}\label{eq:mrmbtheta}
\begin{aligned}
 \{\theta^*, C^*\} = \arg \min_{\boldsymbol\theta \in \boldsymbol\Theta,  {C} > 0| \boldsymbol{\alpha} = \bm{1}}~~~  J~~.
\end{aligned}
\end{equation}
The optimal kernel parameter $\boldsymbol \upsilon^*$ for an $M$-class SVM classifier can then be obtained by
\begin{equation}\label{eq:mrmb}
\begin{aligned}
\boldsymbol \upsilon^* = \arg \min_{\boldsymbol \upsilon \in \boldsymbol\Upsilon | \theta = \theta^*}~~~  J~~.
\end{aligned}
\end{equation}
The derivative of $J$ with respect to  $\upsilon_z$ is given by
\begin{equation}\label{eq:mrmbd}
\begin{aligned}
\frac{\partial J}{\partial \upsilon_z} =
\sum_{1\leq i < j \leq M}\left( \|\bm{w}_{ij}\|^2\frac{\partial R_{ij}^2}{\partial \upsilon_z} + R_{ij}^2\frac{\partial\|\bm{w}_{ij}\|^2}{\partial \upsilon_z}\right),
\end{aligned}
\end{equation}
where $\frac{\partial R_{ij}^2}{\partial \upsilon_z}$ and $\frac{\partial\|\bm{w}_{ij}\|^2}{\partial \upsilon_z}$  for $\text{SVM}_{ij}$ can be obtained by following Eq. (\ref{eq:dr2}) and (\ref{eq:dw2}). $J$ can be optimized by using gradient-based methods.

In classification, the label of a test sample $\bm{X}$ can be assigned  using the max-wins classification rule by
\begin{equation}\label{eq:svmdec}
\begin{aligned}
t(\bm{X}) = \arg \max_{i = 1, \dots, M} \left( \sum_{j = 1, j \neq i}^M \sign \left( s_{ij}\right)\right),
\end{aligned}
\end{equation}
where $s_{ij}$ is the decision score of the binary  classifier $\text{SVM}_{ij}$, and it is computed as
\begin{displaymath}
s_{ij} = \sum_{z = 1}^{n_i + n_j} {\eta^*_z}t_z k(\bm{X}, \bm{X_z}) + b^*_{ij},~~~\bm{X_z} \in \boldsymbol\Omega_i \cup \boldsymbol\Omega_j~~.
\end{displaymath}

\begin{algorithm}[!tb] %
\renewcommand{\algorithmicrequire}{\textbf{Input:}}
\renewcommand\algorithmicensure {\textbf{Output:} }
\caption{Proposed discriminative Stein kernel learning with the radius margin bound or trace margin criterion} %
\label{alg:2} %
\begin{algorithmic}[1] %
\REQUIRE 
A training set $\boldsymbol\Omega = \{(\bm{X}_i,t_{i})\}_{i=1}^{n}$, stopping criteria: i) The total number of iterations $T$; ii) A small positive value $\tau$.
\ENSURE 
$\theta^*$; $\boldsymbol \upsilon^* = \{\boldsymbol{\alpha}^*, C^*\}$.~~\\
\STATE Find $\{\theta^*, C^*\}$  by  solving Eq. (\ref{eq:mrmbtheta});

\FOR { $t =  1:T$}
    \STATE Solve $\|\bm{w}_{ij}\|^2$ in Eq. (\ref{eq:svmcomb}) according to Eq. (\ref{eq:svm});
    \STATE Solve $R_{ij}^2$ in Eq. (\ref{eq:svmcomb}) according to Eq. (\ref{eq:r2}) or approximate it with $\tr(\bm{S}_T)$;
    \STATE Update $\boldsymbol \upsilon$ by a gradient-based method via Eq. (\ref{eq:mrmbd});
    \IF{$|J_{t+1} - J_{t}| \leq \tau J_{t}$~~($J$ is defined in Eq. (\ref{eq:svmcomb}))}
     \STATE Break;
     \ENDIF
\ENDFOR
\RETURN $\theta^*, \boldsymbol \upsilon^*$; %
\end{algorithmic}
\end{algorithm}
We also consider a variant of the radius margin bound by replacing $R_{ij}^2$ with $\tr(\bm{S}_T)$,  where $\tr(\bm{S}_T) = \tr(\bm{S}_B) + \tr(\bm{S}_W)$. It can be calculated by using Eq.~(\ref{trsb}) and (\ref{trsw}) on $\boldsymbol\Omega_i \cup \boldsymbol\Omega_j$. As revealed in~\cite{wang2008feature}, $R_{ij}^2$ is closely related to $\tr(\bm{S}_T)$ and both of them measure the scattering of samples in a feature space $\mathcal{F}$.
Replacing $R_{ij}^2$ with $\tr(\bm{S}_T)$ can often result in more stable optimization~\cite{liu2013efficient}, and solving the quadratic programming (QP) problem in Eq. (\ref{eq:r2}) can also be avoided. 
In the experimental study, both methods, named \textit{radius margin bound} and \textit{trace margin criterion}, are implemented to investigate the performance of DSK. The overall procedure is outlined in Algorithm \ref{alg:2}.

\section{Computational issue}
As will be shown in the experiments, the optimization problems of the proposed DSK can be efficiently solved and often converge in a few iterations. 
Two stopping criteria are used: i) Optimization will be terminated when the difference of the objective values at two successive iterations is below a predefined threshold $\tau$; ii) The optimization will also be stopped when the number of iterations exceeds a predefined threshold $T$. We set $\tau = 10^{-5}$ and $T = 100$  in our experiment. 

The kernel alignment and class separability criteria (defined in Eq. (\ref{eq:kab}) and (\ref{eq:cs}), respectively) can be  quickly computed with a given kernel matrix.  The major computational bottleneck is at the computation of  the kernel matrix $\bm{K}$, which is repeatedly evaluated for various $\boldsymbol\alpha$ values. For a training set of $n$ SPD matrices, the time complexity of optimizing $\boldsymbol\alpha$ using the kernel alignment or the class separability criterion is $\mathcal O(mn^2d^{3})$, where $m$ is the total number of objective function evaluations and $\mathcal O(d^{3})$ is the complexity of eigen-decomposition of a $d \times d$ SPD matrix. Note that the complexity $\mathcal O(n^2)$ is common for all kernel parameter learning algorithms.  
Once $\boldsymbol\alpha^{*}$ is obtained, computing the proposed DSK on a pair of SPD matrices is $\mathcal O(d^{3})$. It is comparable to the original Stein kernel, which has the complexity of $\mathcal O(d^{2.373})$ via computing the determinant instead of conducting an eigen-decomposition. 

In the framework using the radius margin bound, $\boldsymbol \upsilon$~(defined before Eq.~(\ref{eq:rmb})) is optimized by a combination of $M(M-1)/2$ binary SVM classifiers. Once $\boldsymbol \upsilon$ is updated, 
$\mathcal O \left(\left(n_i + n_j\right)^2d^{3}\right)$ is required to calculate the kernel matrix for $\text{SVM}_{ij}$, and a QP problem involving $n_i + n_j$ samples needs to be solved to update $R_{ij}^2$ or $\|\bm w_{ij}\|^2$. 
Let $\mathcal O(\text{QP}(n))$ denote the computational complexity to solve a QP problem of $n$ samples.
The overall complexity to optimize  $\boldsymbol\alpha$ in the framework of  radius margin bound will be $\mathcal O  \left( m \sum_{1 \leq i < j \leq M}\left[\left(n_i + n_j\right)^2d^{3} +  2\text{QP}(n_i + n_j)\right]\right)$. In addition, one QP optimization can be avoided when the trace margin criterion is used, leading to a reduced complexity of $\mathcal O  \left( m \sum_{1 \leq i < j \leq M}\left[\left(n_i + n_j\right)^2d^{3} +  \text{QP}(n_i + n_j)\right]\right)$. 
At last, the computational complexity of classifying a test sample by Eq. (\ref{eq:svmdec}) is $\mathcal O \left(\sum_{1 \leq i < j \leq M}\left[\left(n_i + n_j\right)d^{3}\right]\right)$.

\section{Experimental Result}
In this experiment, we compare the proposed discriminative Stein kernel ({DSK}) with the original Stein kernel ({SK}) on various image classification tasks. The other metrics listed in Table \ref{tb:spdmetric} (in Section \ref{Sec:related-work}) will also be compared. Source code implementing the proposed method is publicly available\footnote{\url{https://github.com/seuzjj/DSK.git}}.


Four data sets are used. Two of them are the Brodatz data set \cite{ff1999} for texture classification and the FERET data set \cite{tf2000} for face recognition, which have been used in the literature \cite{sc2012}. The third one is the ETH-80~\cite{leibe2003analyzing} data set widely used for visual object categorization~\cite{km2013,vemulapalli2013kernel}.
The last one is a resting-state functional Magnetic Resonance Imaging (rs-fMRI) data set from ADNI benchmark database (http://adni.loni.usc.edu). 
In Brodatz, FERET and ETH-$80$ data sets, images are represented by covariance descriptors. In the rs-fMRI data set, correlation matrix is extracted from each image to represent the corresponding subject. The details of these datasets will be introduced in the following subsections.

We employ both $k$-NN  and SVM as the classifiers. For the kernel alignment and class separability frameworks, $k$-NN is used with the DSK as the similarity measure, since it does not involve any other (except $k$) algorithmic parameter. This allows the comparison to directly reflect the change from SK to DSK. For the radius margin bound framework, SVM classifier is used since it is inherently related to this bound. 

In this experiment, the {DSK} obtained by the kernel alignment and class separability are called DSK-KA and DSK-CS. Also, DSK-RM indicates the {DSK} obtained by the radius margin bound, while DSK-TM denotes the {DSK} obtained by trace margin criterion. Subscripts $p$ or $c$ is used to indicate whether $\boldsymbol{\alpha}$ acts as the power or the coefficient of eigenvalues. All the names are summarized in Table \ref{DSKdenote}.

\begin{table}[!ht]
\caption{The name of DSK under different learning criteria}
\label{DSKdenote} \centering
\begin{tabular}{p{42pt}p{35pt}p{33pt}p{40pt}p{43pt}}
\toprule
$\boldsymbol{\alpha}$  as&Kernel alignment&Class separability& Radius margin bound& Trace margin criterion\\
\hline
  power&DSK-KA$_p$&DSK-CS$_p$&DSK-RM$_p$&DSK-TM$_p$\\
\hline
 coefficient&DSK-KA$_c$&DSK-CS$_c$&DSK-RM$_c$&DSK-TM$_c$\\
\bottomrule
\end{tabular}
\end{table}
All parameters, including the $k$ of $k$-NN, the regularization parameter of SVM, $\lambda$ in Eq.~(\ref{eq:ka}), $\theta$ in all the kernels in Table \ref{tb:spdmetric}, and the power order $\zeta$ in the Power-Euclidean metric  are chosen via multi-fold cross-validation on the training set. 

In the experiments, we perform binary classification on the Brodatz and rs-fMRI data sets and  multi-class classification on the Brodatz, FERET and ETH-$80$ data sets. For each experiment on the Brodatz, FERET and ETH-$80$ data sets, the data are randomly split into two equal-sized subsets for training and test. The procedure is repeated $20$ times for each task to obtain stable statistics. LOO strategy is used for the  rs-fMRI data set because the size of this data set is small. Besides classification accuracy, the $p$-value obtained by  paired Student's t-test between DSK and SK will be used to evaluate the significance of improvement ($p$-value $\leq 0.05$ is used).

\subsection{Results on the Brodatz texture data set (binary and multi-class cases)}
The Brodatz data set contains $112$ images, each of which represents one class of texture. Following the literature~\cite{sc2012}, a set of sub-regions are cropped from each image  as the samples of the corresponding texture class. The covariance descriptor~\cite{rc2006} is used to describe a texture sample (sub-region) as follows.

\renewcommand{\labelenumi}{(\theenumi)}

\begin{enumerate}
\item Each original texture image is scaled to a uniform size of $256 \times 256$;
\item Each image is then split into $64$ non-overlapping sub-regions of size $32 \times{32}$. Each image is considered as a texture class and its sub-regions are used as the samples of this class;
\item A five-dimensional feature vector $\phi(x,y) = [ I(x,y), |\frac{\partial I}{\partial x}|, |\frac{\partial I}{\partial y}|, |\frac{\partial^2 I}{\partial x^2}|, |\frac{\partial^2 I}{\partial y^2}| ]$  is extracted at pixel $(x,y)$ in each sub-region, where $I(x,y)$ denotes the intensity value at that pixel;
\item Each sub-region is represented by a $5 \times 5$ covariance matrix estimated by using all ($1024=32\times{32}$) the features vectors obtained from that sub-region.
\end{enumerate}
For the experiment of binary classification, we first run pairwise classification between the $112$ classes by using SK with the $k$-NN classifier. The obtained classification accuracies are sorted in ascending order. The top $15$ pairs with the lowest accuracies, which represent the most difficult classification tasks, are selected. The rest of these pairs are not included because SK has been able to obtain almost $100\%$ accuracy on them. 
 As we observed in the experiment, DSK achieves equally excellent performance as SK on these pairs. The selected $15$ pairs are shown in the supplementary material with image IDs. The texture images in each pair are visually similar to each other and it is challenging to classify them. In short, we obtain $15$ pairs of classes. Each class consists of $64$ samples, and each sample is represented by a $5 \times 5$ covariance descriptor.
\begin{table}[ht]
\caption{Comparison of Classification accuracy (in percentage) on each of the $15$ most difficult pairs from Brodatz texture data set.}
\label{tab:brodatzb15} \centering
\begin{tabular}{|p{35 pt}|p{\smalltablecolumnwidth pt}|p{\smalltablecolumnwidth pt}|p{\smalltablecolumnwidth pt}|p{\smalltablecolumnwidth pt}|p{\smalltablecolumnwidth pt}|p{\smalltablecolumnwidth pt}|p{\smalltablecolumnwidth pt}|p{20 pt}|}
\hline
Index&1&2&3&4&5&6&7&8\\
\hline
SK&62.50&67.19&68.75&75.00&75.78&75.79&76.56&77.34\\
\hline
DSK-KA$_p$&\textbf{70.31}&\textbf{73.44}&\textbf{75.00}&\textbf{81.25}&\textbf{76.56}&\textbf{79.69}&\textbf{82.81}&\textbf{79.69}\\
\hline
\hline
Index&9&10&11&12&13&14&15&\textbf{Avg.}\\
\hline
SK&78.13&79.69&80.47&81.25&82.04&83.59&85.94&76.67\\
\hline
DSK-KA$_p$&\textbf{84.37}&\textbf{84.39}&\textbf{84.38}&\textbf{84.38}&\textbf{84.35}&\textbf{84.42}&\textbf{87.50}&\textbf{80.85}\\
\hline
\end{tabular}
\end{table}
\begin{table*}[!ht]
\caption{Comparison of  classification accuracy (in percentage)  averaged on 15 most difficult pairs from Brodatz texture data set.}
\label{tab:brodatzb} \centering
\begin{tabular}{|p{\tablecolumnwidth pt}|p{\tablecolumnwidth pt}|p{\tablecolumnwidth pt}||p{\tablecolumnwidth pt}|p{\tablecolumnwidth pt}||p{\tablecolumnwidth pt}|p{\tablecolumnwidth pt}|p{\tablecolumnwidth pt}||p{\tablecolumnwidth pt}|p{\tablecolumnwidth pt}|}
\hline
  \multicolumn{5}{|c||}{$k$-NN}&\multicolumn{5}{c|}{SVM}\\
\hline
\multicolumn{3}{|c||}{Competing methods}&\multicolumn{2}{|c||}{DSK (proposed)}&\multicolumn{3}{|c||}{Competing methods}&\multicolumn{2}{|c|}{DSK (proposed)}\\
\hline
AIRM&CHK&EUK&DSK-KA$_p$&DSK-KA$_c$&AIRM&CHK&EUK&DSK-RM$_p$&DSK-RM$_c$\\
\hline
74.67&73.78&73.59&\textbf{80.85}&78.33&N.A.&77.87&76.04&\textbf{80.57}&79.16\\
$\pm$ 4.48&$\pm$ 4.68&$\pm$ 6.31&$\pm$ 4.96&$\pm$ 4.79&N.A.&$\pm$ 5.89&$\pm$ 5.93&$\pm$ 5.95&$\pm$ 6.34\\
\hline
SK&LEK&PEK&DSK-CS$_p$&DSK-CS$_c$&SK&LEK&PEK&DSK-TM$_p$&DSK-TM$_c$\\
\hline
76.67&74.67&74.70&79.69&77.29&78.38&78.89&78.06&80.47&79.08\\
$\pm$ 5.30&$\pm$ 4.17&$\pm$ 4.16&$\pm$ 3.74&$\pm$ 4.25&$\pm$ 6.21&$\pm$ 4.69&$\pm$ 5.90&$\pm$ 6.55&$\pm$ 6.93\\
\hline
\end{tabular}
\end{table*}

\begin{table*}[!ht]
\caption{Comparison of Classification accuracy (in percentage)  averaged on 112-class classification on Brodatz texture data set.}
\label{tab:brodatzall} \centering
\begin{tabular}{|p{\tablecolumnwidth pt}|p{\tablecolumnwidth pt}|p{\tablecolumnwidth pt}||p{\tablecolumnwidth pt}|p{\tablecolumnwidth pt}||p{\tablecolumnwidth pt}|p{\tablecolumnwidth pt}|p{\tablecolumnwidth pt}||p{\tablecolumnwidth pt}|p{\tablecolumnwidth pt}|}
\hline
  \multicolumn{5}{|c||}{$k$-NN}&\multicolumn{5}{c|}{SVM}\\
\hline
\multicolumn{3}{|c||}{Competing methods}&\multicolumn{2}{|c||}{DSK (proposed)}&\multicolumn{3}{|c||}{Competing methods}&\multicolumn{2}{|c|}{DSK (proposed)}\\
\hline
AIRM&CHK&EUK&DSK-KA$_p$&DSK-KA$_c$&AIRM&CHK&EUK&DSK-RM$_p$&DSK-RM$_c$\\
\hline
74.93&72.19&64.72&78.12&77.50&N.A.&77.39&74.27&\textbf{83.40}&82.94\\
$\pm$ 0.61&$\pm$ 0.55&$\pm$ 0.78&$\pm$ 0.75&$\pm$ 0.69&N.A.&$\pm$ 0.82&$\pm$ 1.26&$\pm$ 0.58&$\pm$ 0.71\\
\hline
SK&LEK&PEK&DSK-CS$_p$&DSK-CS$_c$&SK&LEK&PEK&DSK-TM$_p$&DSK-TM$_c$\\
\hline
76.80&74.38&72.02&\textbf{78.43}&77.80&78.01&78.22&76.88&80.41&80.10\\
$\pm$ 0.84&$\pm$ 0.62&$\pm$ 0.65&$\pm$ 0.59&$\pm$ 0.81&$\pm$ 0.43&$\pm$ 1.00&$\pm$ 0.84&$\pm$ 0.47&$\pm$ 0.53\\
\hline
\end{tabular}
\end{table*}
The average classification accuracies on the $15$ binary classification tasks are compared in Table~\ref{tab:brodatzb}. The left half of the table shows the results when the $k$-NN is the classifier, while the right half is for the SVM classifier. As seen from the left half, SK achieves the best performance (76.67\%) under the column of ``Competing methods''. At the same time, the proposed {DSK-KA} and {DSK-CS} consistently achieve better performance than {SK}, when the parameter $\boldsymbol \alpha$ is used as the power or coefficient. Especially, DSK-KA$_p$ achieves the best performance (80.85\%), obtaining an improvement of above $4$ percentage points over SK and more than $6$ percentage points over the other methods. The relatively large standard deviation in Table~\ref{tab:brodatzb} is mainly due to the variation of accuracy rates of the  $15$ tasks. Actually, the improvement of DSK over SK is statistically significant because the $p$-value  between DSK-KA$_p$ and SK is as small as $5.4 \times 10^{-6}$. To better show the difference between DSK-KA$_p$ and SK, their performance on each of the $15$ pairs is reported in Table \ref{tab:brodatzb15}. As seen, DSK-KA$_p$ consistently outperforms SK on each task. The right half of Table~\ref{tab:brodatzb} compares the DSK obtained by the radius margin bound with the other kernel methods, by using SVM as the classifier. As seen, the four variants of DSK in this case, DSK-RM$_p$, DSK-RM$_c$, DSK-TM$_p$ and DSK-TM$_c$, outperform the other kernel methods, including SK. Note that AIRM does not admit a valid kernel~\cite{km2013} and is not included in the comparison.

We also test DSK on multi-class classification involving all the $112$ classes of Brodatz  data. As seen from Table \ref{tab:brodatzall}, all the DSK methods outperform SK and other methods in comparison. Specifically, as indicated in the left part of Table \ref{tab:brodatzall}, SK has the highest classification accuracy ($76.80\%$) among all these existing methods when $k$-NN is used as the classifier. Meanwhile, compared with SK, DSK-CS$_p$ achieves a further improvement of $1.6$ percentage points with $p$-value of $0.0018$. When SVM is used in the right part of Table \ref{tab:brodatzall}, DSK-RM$_p$ boosts the performance of SK from $78.01\%$ to $83.40\%$, obtaining an improvement of $5.39$ percentage points. 

\begin{figure}[!ht]
\begin{center}
\begin{tabular}{cc}
{\includegraphics[width = \twosmallfigwidth mm]{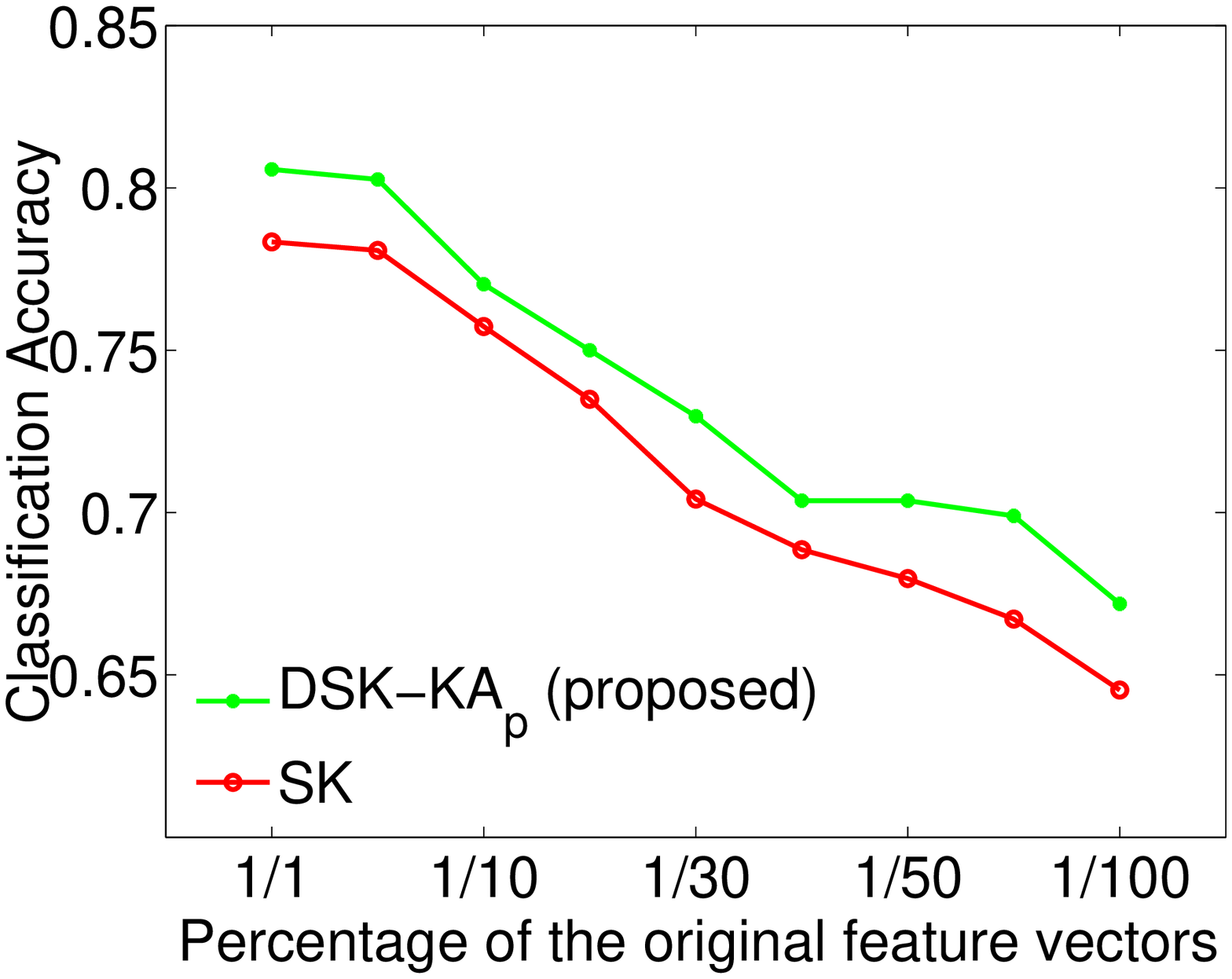}}&
{\includegraphics[width = \twosmallfigwidth mm]{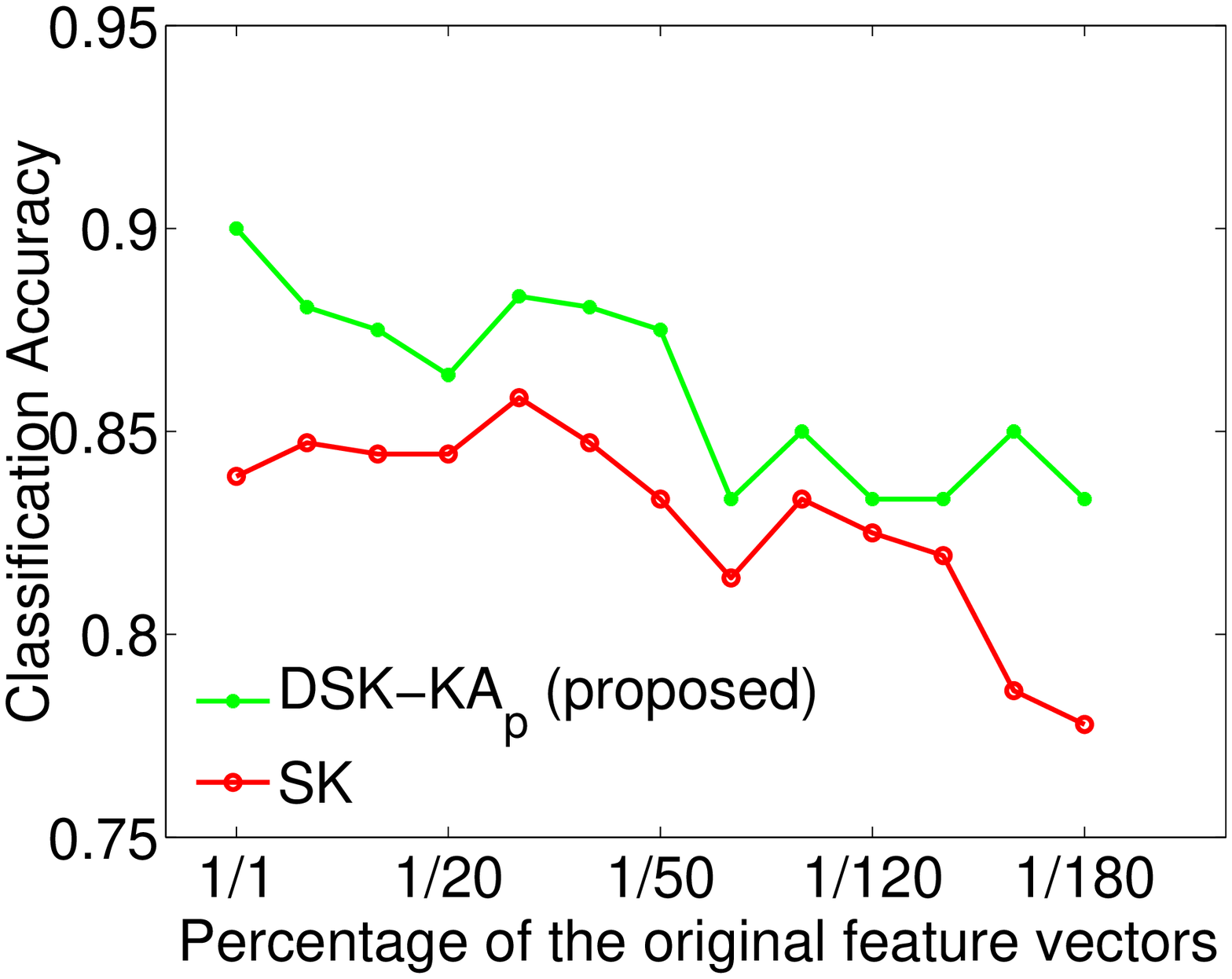}}\\ 
(a) On Brodatz dataset & (b) On FERET dataset\\
\end{tabular}
\end{center}
\caption{Comparison of {DSK-KA}$_p$ and {SK} when different percentage of the $1024$ feature vectors are used to estimate the covariance descriptor for each image on the Brodatz and FERET data sets.} 
\label{fig:texturedown}
\end{figure}

In addition, we investigate the performance of {DSK} with respect to the number of samples used to estimate the covariance descriptors. Recall that a $5 \times 5$ covariance descriptor is estimated from $1024$ feature vectors $\phi(x,y)$ to represent an image region. This number is sufficiently large compared with the dimensions of the covariance descriptor, which are only five. The improvement in the above results demonstrates the effectiveness of {DSK} over {SK} when there are sufficient samples to estimate the covariance descriptor.  As discussed in Section \ref{Limitations}, covariance matrix estimation is significantly affected by the number of samples. This motivates us to investigate how the performance of   {DSK} and {SK} will change, if the number of feature vectors used to estimate the covariance descriptor is reduced. We take {DSK-KA}$_{p}$ as an example. In Figure \ref{fig:texturedown}(a),  we plot the classification accuracy  of {DSK-KA}$_{p}$ and  {SK} averaged over the $15$ binary classification tasks, when different percentage of the $1024$ feature vectors are used. As shown, {DSK-KA}$_{p}$ consistently achieves better performance than {SK}, although both of them degrade with the decreasing number of feature vectors. This result shows that: i) When the samples available for estimation are inadequate, Stein kernel will become less effective; ii)  {DSK} can effectively improve the performance of {SK} in this case. Figure \ref{fig:texturedown}(b) plots similar result obtained  on the FERET face data set with the same experimental setting. That is, pairwise classification is performed by using SK with the $k$-NN classifier and the top $15$ pairs with the lowest accuracies are selected. The classification accuracies  of {DSK-KA}$_{p}$ and  {SK} averaged over the $15$ selected binary classification tasks are shown in Figure \ref{fig:texturedown}(b). 

\begin{table*}[!ht]
\caption{Comparison of Classification accuracy (in percentage) averaged on 198-class classification on FERET data set.}
\label{tab:feret} \centering
\begin{tabular}{|p{\tablecolumnwidth pt}|p{\tablecolumnwidth pt}|p{\tablecolumnwidth pt}||p{\tablecolumnwidth pt}|p{\tablecolumnwidth pt}||p{\tablecolumnwidth pt}|p{\tablecolumnwidth pt}|p{\tablecolumnwidth pt}||p{\tablecolumnwidth pt}|p{\tablecolumnwidth pt}|}
\hline
  \multicolumn{5}{|c||}{$k$-NN}&\multicolumn{5}{c|}{SVM}\\
\hline
\multicolumn{3}{|c||}{Competing methods}&\multicolumn{2}{|c||}{DSK (proposed)}&\multicolumn{3}{|c||}{Competing methods}&\multicolumn{2}{|c|}{DSK (proposed)}\\
\hline
AIRM&CHK&EUK&DSK-KA$_p$&DSK-KA$_c$&AIRM&CHK&EUK&DSK-RM$_p$&DSK-RM$_c$\\
\hline
84.45&80.43&52.14&\textbf{84.98}&84.83&N.A.&78.52&68.37&81.80&\textbf{84.60}\\
$\pm$ 3.23&$\pm$ 3.54&$\pm$ 4.10&$\pm$ 3.37&$\pm$ 3.38&N.A.&$\pm$ 2.23&$\pm$ 4.04&$\pm$ 2.67&$\pm$ 1.71\\
\hline
SK&LEK&PEK&DSK-CS$_p$&DSK-CS$_c$&SK&LEK&PEK&DSK-TM$_p$&DSK-TM$_c$\\
\hline
83.37&83.02&73.07&84.03&83.95&79.7&78.16&75.22&80.70&83.20\\
$\pm$ 3.33&$\pm$ 3.27&$\pm$ 3.66&$\pm$ 3.55&$\pm$ 3.45&$\pm$ 3.10&$\pm$ 1.73&$\pm$ 3.97&$\pm$ 2.30&$\pm$ 2.44\\
\hline
\end{tabular}
\end{table*}

\begin{figure}[!htb]
\begin{center}
\hspace{5 mm}{\includegraphics[width = 82 mm]{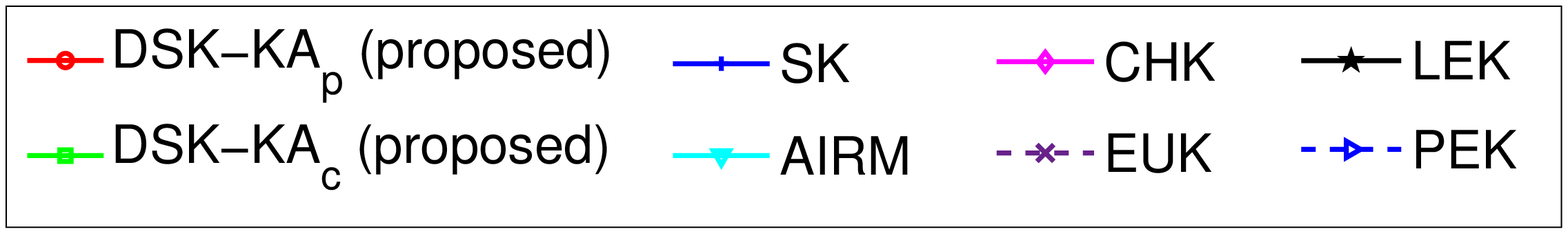}} \\
\begin{tabular}{cc}
{\includegraphics[width = \twosmallfigwidth mm]{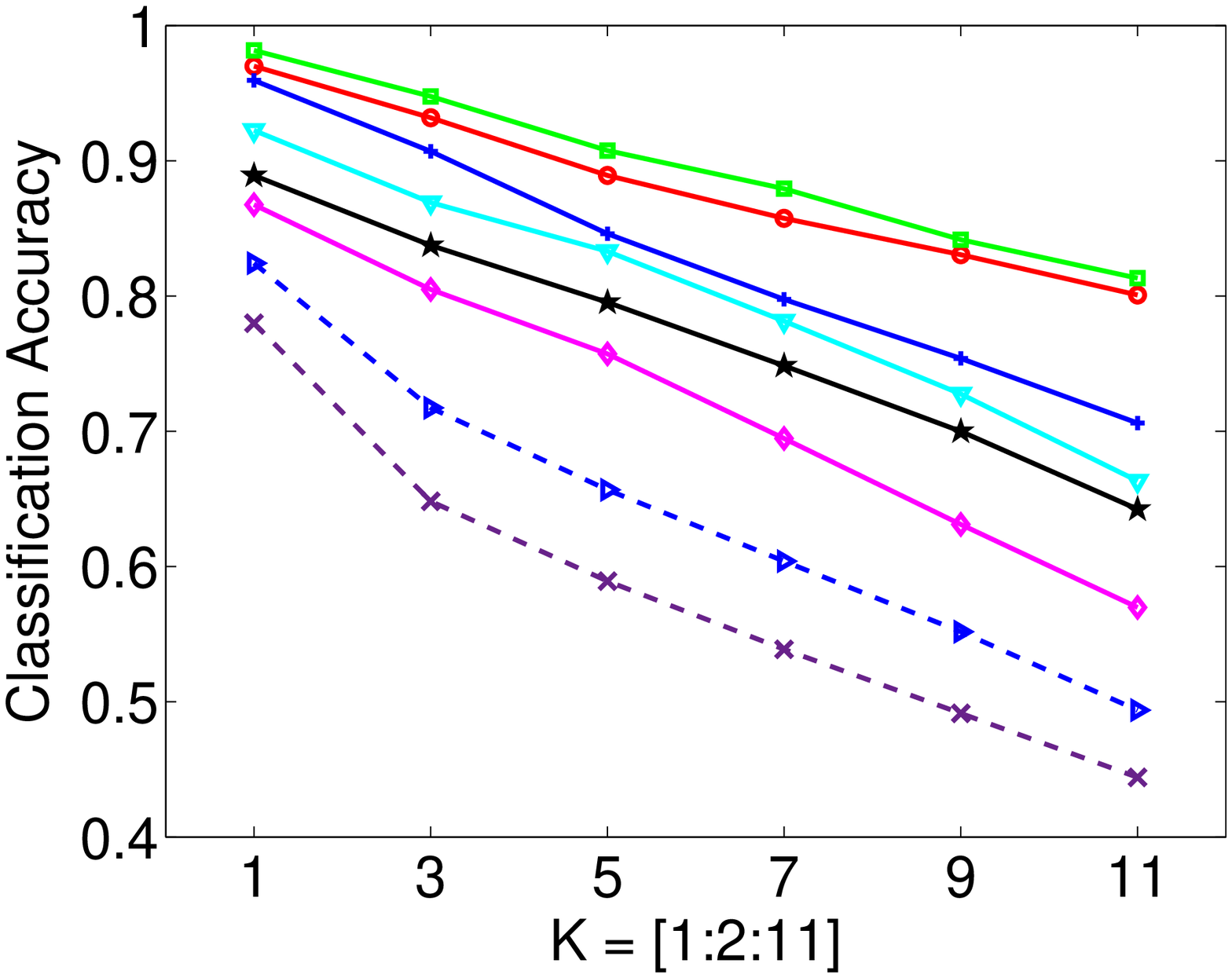}} &
{\includegraphics[width = \twosmallfigwidth mm]{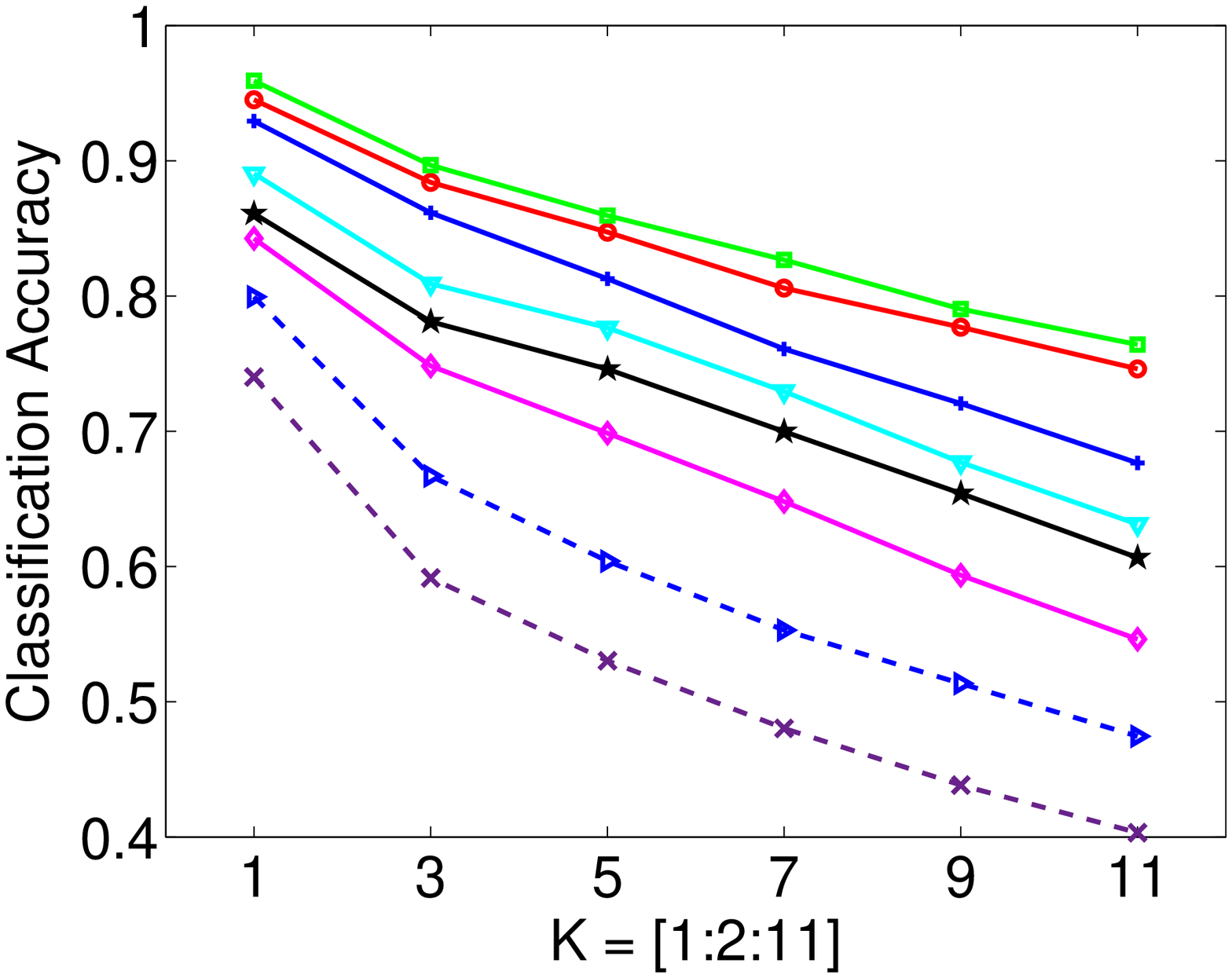}}\\
(a) 10 classes & (b) 20 classes\\
{\includegraphics[width = \twosmallfigwidth mm]{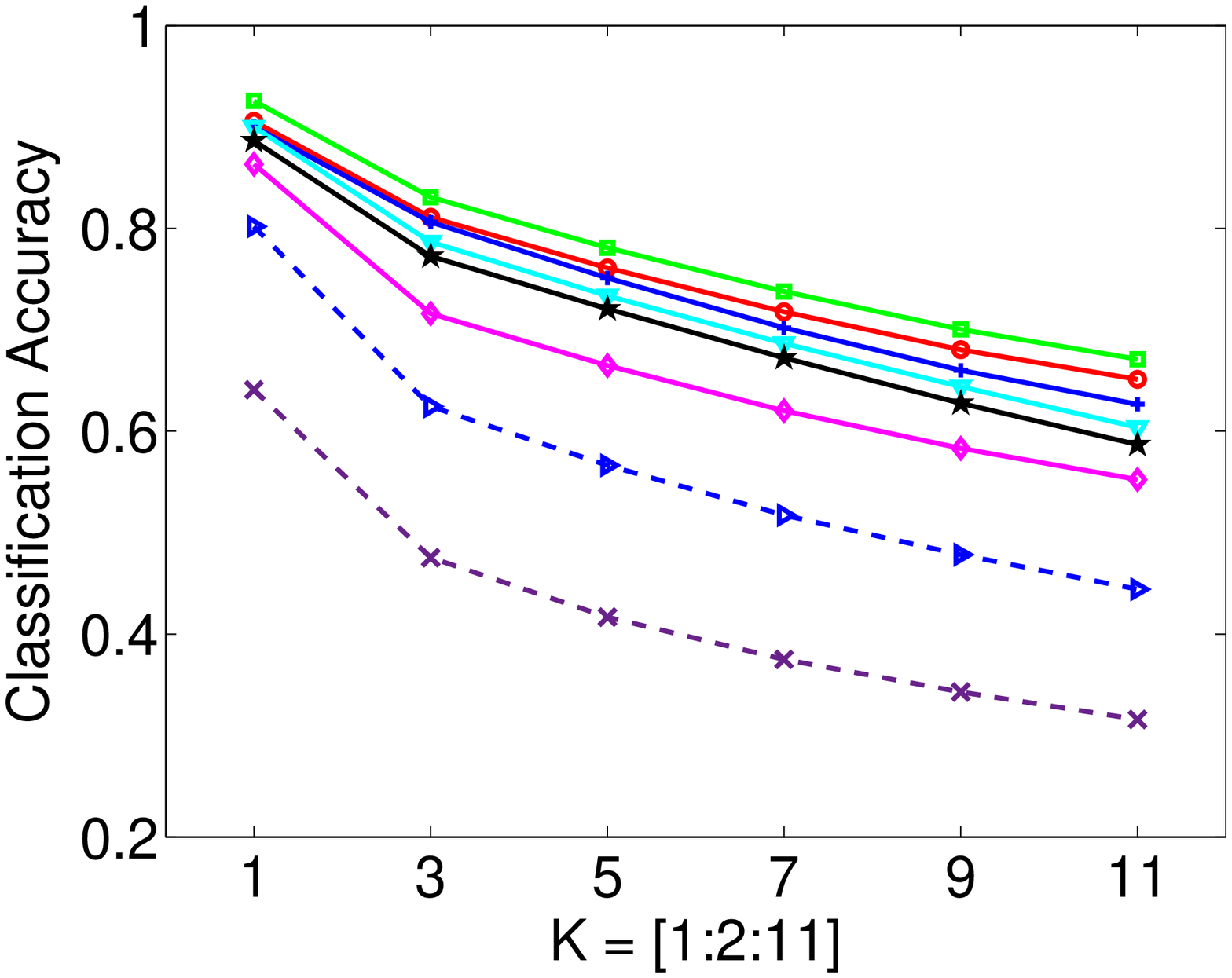}} &
{\includegraphics[width = \twosmallfigwidth mm]{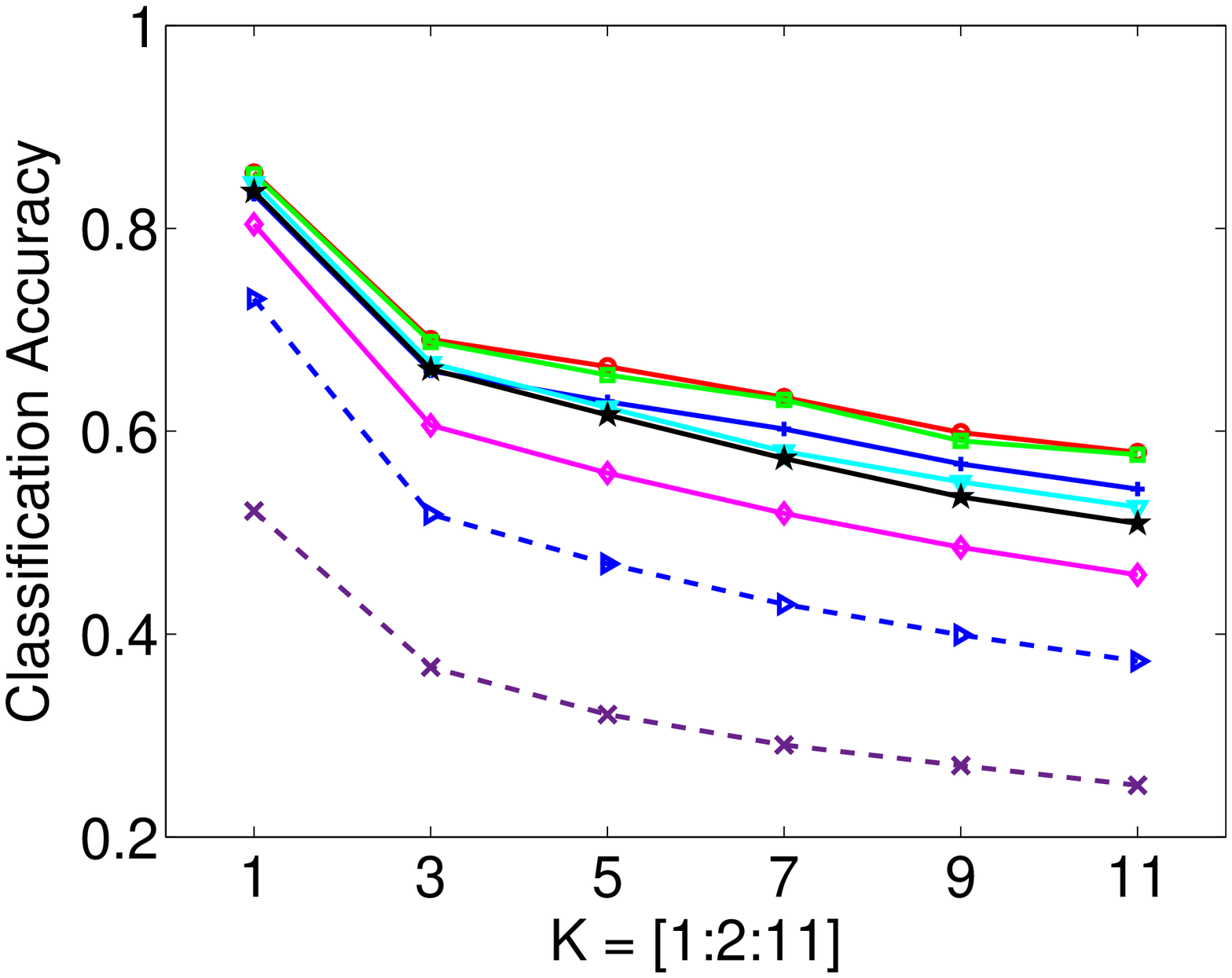}}\\
(c) 40 classes& (d) 198 classes
\end{tabular}
\end{center}
\caption{Comparison of {DSK-KA}, {SK}, and other methods on $10$, $20$, $40$ and all $198$ classes from FERET data set with various $k$ values of $k$-NN.}
\label{fig:fremulti}
\end{figure}

In this experiment, we observe that DSK can often be solved efficiently. The optimization in all the three frameworks only require a few iterations to converge. An example of the evolution of objective function on the Brodatz data set is shown in the supplementary material. In that example, DSK needs at most $15$ iterations to converge.  

\subsection{Results on the FERET face data set (multi-class case)}
We evaluate the proposed {DSK} for face recognition on FERET \cite{tf2000} face data set. We use the `b' subset, which consists of $198$ subjects and each subject has $10$ images  with various  poses and illumination conditions. 
Following the literature \cite{sc2012}, to represent an image, a covariance descriptor is estimated for a $43$-dimensional feature vector extracted at each pixel:
\begin{displaymath} 
\begin{split} 
\phi(x,y) = [I(x,y), x, y, |G_{0,0}(x,y)|, \cdots, |G_{0,7}(x,y)|,\\
 |G_{1,0}(x,y)|, \cdots, |G_{4,7}(x,y)|];
\end{split} 
\end{displaymath} 
where $I(x,y)$ is the intensity value, and $|G_{u,v}(x,y)|$ is the image feature of 2D Gabor wavelets~\cite{il1996}. 

Table \ref{tab:feret} compares the classification accuracy on all the $198$ classes of FERET data. As seen, DSK-KA$_p$ obtains an improvement of 1.6 percentage points over SK. The $p$-value between DSK-KA$_p$ and SK is $0.0026$, which indicates the statistical significance of the improvement. 
When the radius margin bound framework is used, DSK also consistently performs better than SK. Especially, DSK-RM$_c$  achieves an improvement as high as 4.9 percentage points over SK, with $p$-value of $ 1.8 \times $$10^{-5}$.

To test how the {DSK} performs with the number of classes and the $k$ value of $k$-NN, we evaluate  {DSK-KA}$_p$ and {DSK-KA}$_c$ by using  $10$, $20$, $40$, and all $198$ classes, respectively, with $k = [1:2:11]$. The experiment is conducted as follows.


1) For the classification tasks with $10$, $20$ and $40$ classes, these classes are randomly selected from the $198$ classes. Five facial images are randomly chosen from a class for training, and the remaining ones are used for test. Both the selection of classes and samples are repeated $10$ times, respectively.

2) For the classification tasks with all the $198$ classes,  five facial images are randomly chosen from each class for training, and the remaining ones are used for test. The selection of samples is repeated $20$ times.

The averaged classification accuracy of $10$, $20$, $40$ and $198$ classes with various $k$ values are plotted in Fig.~\ref{fig:fremulti}(a)-(d). As seen, both {DSK-KA}$_p$ and {DSK-KA}$_c$ consistently outperform {SK} and other methods. This confirms in further that the proposed DSK can increase class discrimination by adjusting the eigenvalues of the SPD matrices, making Stein kernel better align with specific classification tasks. An example of the learned adjustment parameters  in various classification tasks is shown in the supplementary material.

\subsection{Results on ETH-$80$ data set (multi-class case)}
ETH-$80$ contains eight categories with ten objects per category and $41$ images for each object. 
The features same as those used on the Brodatz texture data set  are extracted from each image and a $5 \times 5$ covariance descriptor is constructed as a representation of the image.
The ten objects in the same category are labeled as the same class. We perform an eight-class classification task using DSK. As previously mentioned, data are randomly split into $20$ pairs of training/test subsets ($50\%:50\%$) to obtain stable statistics. 
Table \ref{tab:eth80} reports the performance of various methods. As seen, DSK still demonstrates the best performance. Specifically, DSK-CS$_p$ achieves $2.3$ percentage points improvement over SK with $k$-NN as the classifier, while DSK-TM$_p$ achieves an improvement of $2.4$ percentage points over SK, when SVM is used as the classifier. 

The  above experiments  demonstrate the advantage of DSK in various important image recognition tasks. Also, this advantage is consistently observed when DSK is learned with three different criteria. This verifies the generality of DSK.

\begin{table*}[!ht]
\caption{Comparison of Classification accuracy (in percentage) averaged on ETH-80 data set.}
\label{tab:eth80} \centering
\begin{tabular}{|p{\tablecolumnwidth pt}|p{\tablecolumnwidth pt}|p{\tablecolumnwidth pt}||p{\tablecolumnwidth pt}|p{\tablecolumnwidth pt}||p{\tablecolumnwidth pt}|p{\tablecolumnwidth pt}|p{\tablecolumnwidth pt}||p{\tablecolumnwidth pt}|p{\tablecolumnwidth pt}|}
\hline
  \multicolumn{5}{|c||}{$k$-NN}&\multicolumn{5}{c|}{SVM}\\
  \hline
\multicolumn{3}{|c||}{Competing methods}&\multicolumn{2}{|c||}{DSK (proposed)}&\multicolumn{3}{|c||}{Competing methods}&\multicolumn{2}{|c|}{DSK (proposed)}\\
\hline
AIRM&CHK&EUK&DSK-KA$_p$&DSK-KA$_c$&AIRM&CHK&EUK&DSK-RM$_p$&DSK-RM$_c$\\
\hline
79.39&80.14&78.33&80.92&80.71&N.A.&80.76&79.27&81.30&80.67\\
$\pm$ 0.78&$\pm$ 0.47&$\pm$ 1.19&$\pm$ 0.87&$\pm$ 0.85&N.A.&$\pm$ 1.10&$\pm$ 1.98&$\pm$ 0.81&$\pm$ 0.93\\
\hline
SK&LEK&PEK&DSK-CS$_p$&DSK-CS$_c$&SK&LEK&PEK&DSK-TM$_p$&DSK-TM$_c$\\
\hline
79.71&79.29&79.65&\textbf{82.04}&80.11&80.30&80.21&80.16&\textbf{82.70}&81.55\\
$\pm$ 0.82&$\pm$ 0.75&$\pm$ 0.65&$\pm$ 0.91&$\pm$ 0.88&$\pm$ 0.79&$\pm$ 1.16&$\pm$ 1.04&$\pm$ 1.05&$\pm$ 0.84\\
\hline
\end{tabular}
\end{table*}

\subsection{Brain imaging classification}
We now test DSK on brain imaging analysis using correlation matrix. Correlation matrix is a SPD matrix in which each off-diagonal element denotes the
correlation coefficient between a pair of variables. It is commonly used in neuroimaging analysis to model functional brain networks for discriminating patients with Alzheimer's Disease (AD) from healthy controls~\cite{wee2012resting}. In this task, a correlation matrix is extracted from each rs-fMRI image and used to represent the brain network of the corresponding subject. This is also a classification task involving SPD matrices. 

The rs-fMRI data set from ADNI consists of $44$ mild cognitive impairment ({MCI}, an early warning stage of Alzheimer's disease) patients and $38$ healthy controls. The rs-fMRI images of these subjects are pre-processed by a standard pipeline using SPM8 (http://www.fil.ion.ucl.ac.uk/spm) for rs-fMRI. All the images are spatially normalized into a common space and parcellated into $116$ regions of interest (ROI) based on a predefined brain atlas. $42$ ROIs that are known to be related to AD are selected~\cite{huang2009learning} in our experiment and the mean rs-fMRI signal within each ROI is extracted as the features. We then construct a $42\times42$ correlation matrix for each subject~\cite{wee2012resting}. 
The rs-fMRI images and the correlation matrices are illustrated in the supplementary material.

In this experiment, we compare {DSK} with the other methods to classify the correlation matrices. The classification is conducted in the LOO manner due to the limited number of samples. Specifically,  one sample is left out as the test set with the remaining samples as the training set. This process is repeated for each of the samples. As seen in Table~\ref{tab:fmri}, DSK again achieves the best classification performance. DSK-CS$_c$ increases the classification accuracy of SK from $56.1\%$ to $62.2\%$ with $k$-NN as the classifier, obtaining an improvement of $6.1$ percentage points. Similarly, DSK-RM$_p$ obtains an improvement of $4.9$ percentage points over SK when SVM is used as the classifier. This experimental  result indicates that DSK holds promise for handling various types of SPD matrices in broad applications.


\begin{table*}[!ht]
\caption{Comparison of Classification accuracy (in percentage) on {fMRI} data.}
\label{tab:fmri} \centering
\begin{tabular}{|p{\tablecolumnwidth pt}|p{\tablecolumnwidth pt}|p{\tablecolumnwidth pt}||p{\tablecolumnwidth pt}|p{\tablecolumnwidth pt}||p{\tablecolumnwidth pt}|p{\tablecolumnwidth pt}|p{\tablecolumnwidth pt}||p{\tablecolumnwidth pt}|p{\tablecolumnwidth pt}|}
\hline
  \multicolumn{5}{|c||}{$k$-NN}&\multicolumn{5}{c|}{SVM}\\
  \hline
\multicolumn{3}{|c||}{Competing methods}&\multicolumn{2}{|c||}{DSK (proposed)}&\multicolumn{3}{|c||}{Competing methods}&\multicolumn{2}{|c|}{DSK (proposed)}\\
\hline
AIRM&CHK&EUK&DSK-KA$_p$&DSK-KA$_c$&AIRM&CHK&EUK&DSK-RM$_p$&DSK-RM$_c$\\
\hline
56.10&52.44&50.00&60.98&59.76&N.A.&53.66&54.88&\textbf{59.76}&54.88\\
\hline
SK&LEK&PEK&DSK-CS$_p$&DSK-CS$_c$&SK&LEK&PEK&DSK-TM$_p$&DSK-TM$_c$\\
\hline
 56.10&54.88&51.22&60.98&\textbf{62.20}&54.88&53.66&57.32&\textbf{59.76}&53.66\\
\hline
\end{tabular}
\end{table*}

\section{Discussion}
\subsection{On using $\boldsymbol\alpha$ as power or coefficient}


The two ways of using $\boldsymbol\alpha$ can be theoretically related. This is because for any $\lambda^{\alpha_p}$($\lambda >0$), we can always find a coefficient $\alpha_c$ that satisfies ${\alpha_c}\lambda = \lambda^{\alpha_p}$ by setting $\alpha_c = \frac{\lambda^{\alpha_p}}{\lambda}$. In practice, these two ways could lead to different solutions, because the corresponding objective functions are different and the resulting optimizations are not convex. Comparatively, the power method is recommended due to: i) using $\boldsymbol\alpha$ as a power can automatically maintain the SPD property since $\lambda^{\alpha_p}$($\lambda >0$) is always positive, while using it as a coefficient requires an additional constraint of ${\alpha_c} > 0$; ii) we empirically find that the power method often converges faster and achieves better performance than  the coefficient method.
\subsection{On the computational efficiency of DSK}
\begin{figure}[!ht]
\begin{center}
\begin{tabular}{cc}
{\includegraphics[height = 58mm, width = 85 mm]{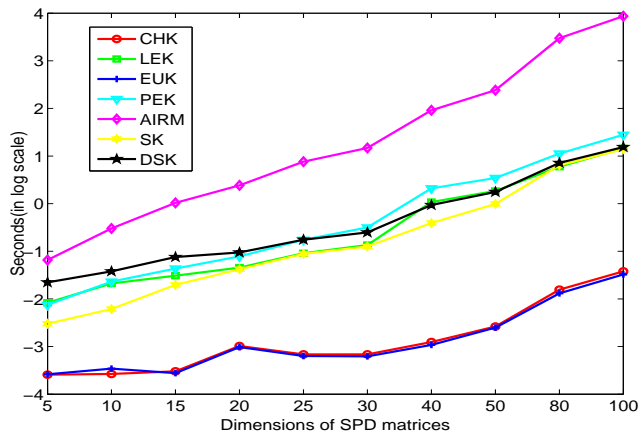}}
\end{tabular}
\end{center}
\caption{Comparison of timing results of the measures with respect to the dimensions of SPD matrices}
\label{fig:timing}
\end{figure}

Once the adjustment parameter $\boldsymbol\alpha$ is obtained, DSK can be computed for a set of SPD matrices. 
Fig.~\ref{fig:timing} compares the timing result of the methods in Table \ref{tb:spdmetric} for computing a similarity matrix of $100$ SPD matrices. The dimensions of these SPD matrices are gradually increased from $5$ to $100$. The experiment is conducted on a desktop computer with $3.6$ GHz $\mathrm{Core}^{\mathrm{TM}}$ i$7-3820$ CPU and $32$GB memory. As seen, DSK can be computed as efficiently as SK, PEK and LEK, and all of them are significantly faster than AIRM. For example, DSK only needs $3.3$ seconds to compute the similarity matrix of $100$-dimensional SPD matrices, while AIRM needs as many as $51$ seconds. AIRM will become  less efficient when the dimensions are high or the number of SPD matrices is large. The kernel methods, such as CHK, LEK, EUK, PEK, SK and DSK, can usually handle the situation more efficiently.

In addition, the computational cost of learning $\boldsymbol\alpha$ could be reduced by taking advantage of the facts that i) kernel matrix computation can be run in a parallel manner by evaluating every entry separately; ii) the most time-consuming step, eigen-decomposition, could be speeded up by using approximate techniques, such as the Nystr$\ddot{\text{o}}$m method~\cite{Williams01usingthe}. 
These improvements will be fully explored in the future work.

\begin{figure*}[!ht]
\begin{center}
\begin{tabular}{cccc}
{\includegraphics[width = \fourfigwidth mm]{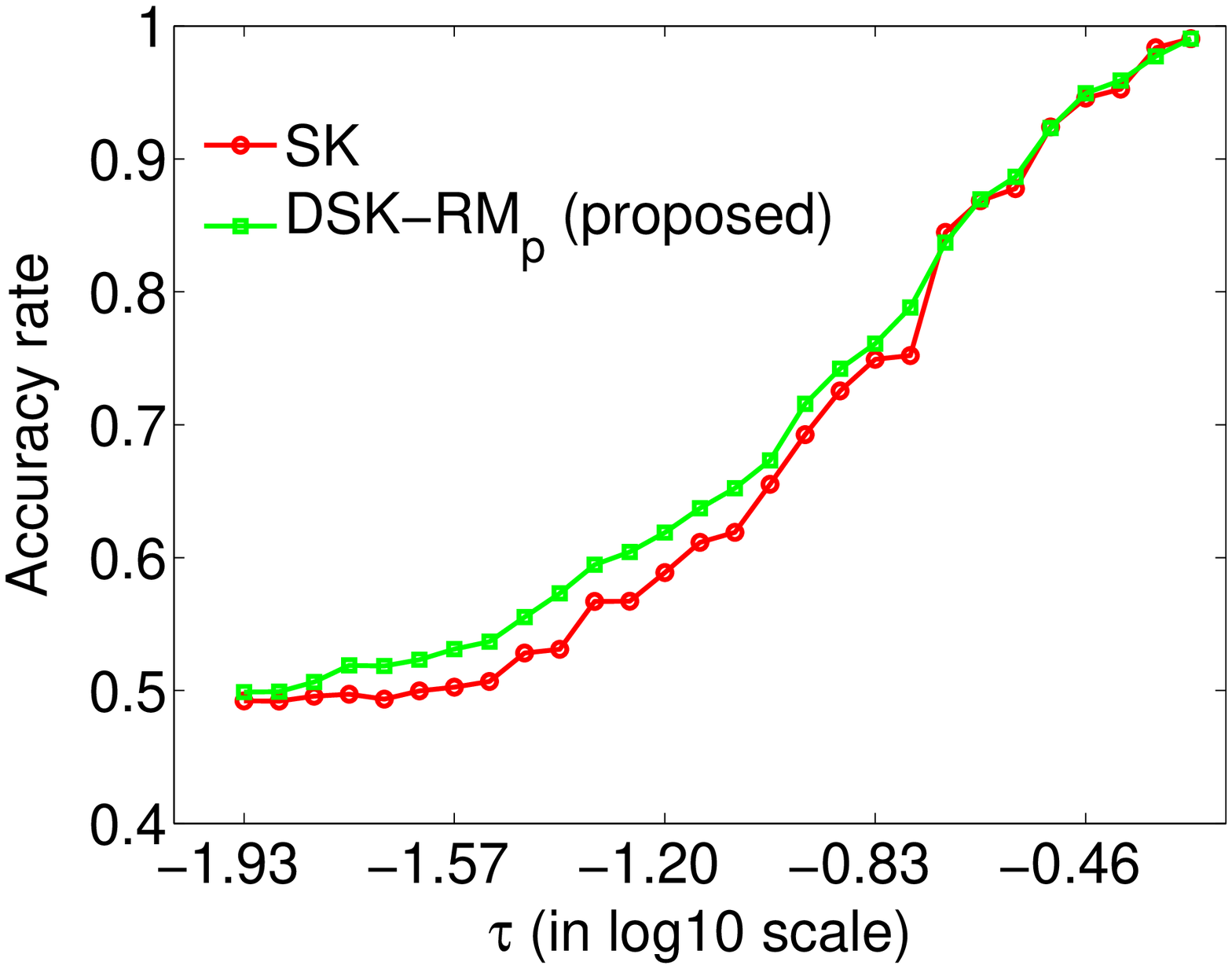}} &
{\includegraphics[width = \fourfigwidth mm]{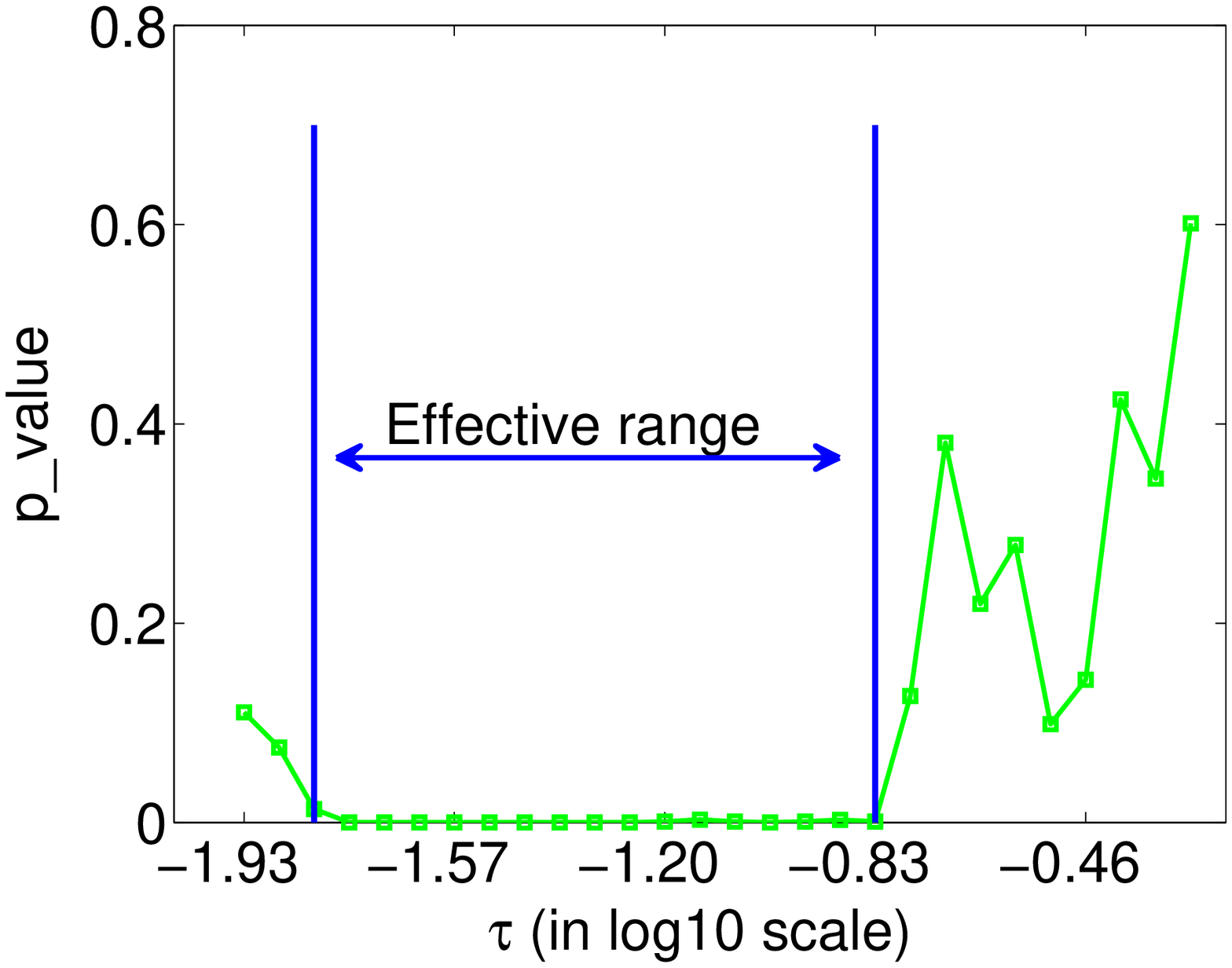}} &
{\includegraphics[width = \fourfigwidth mm]{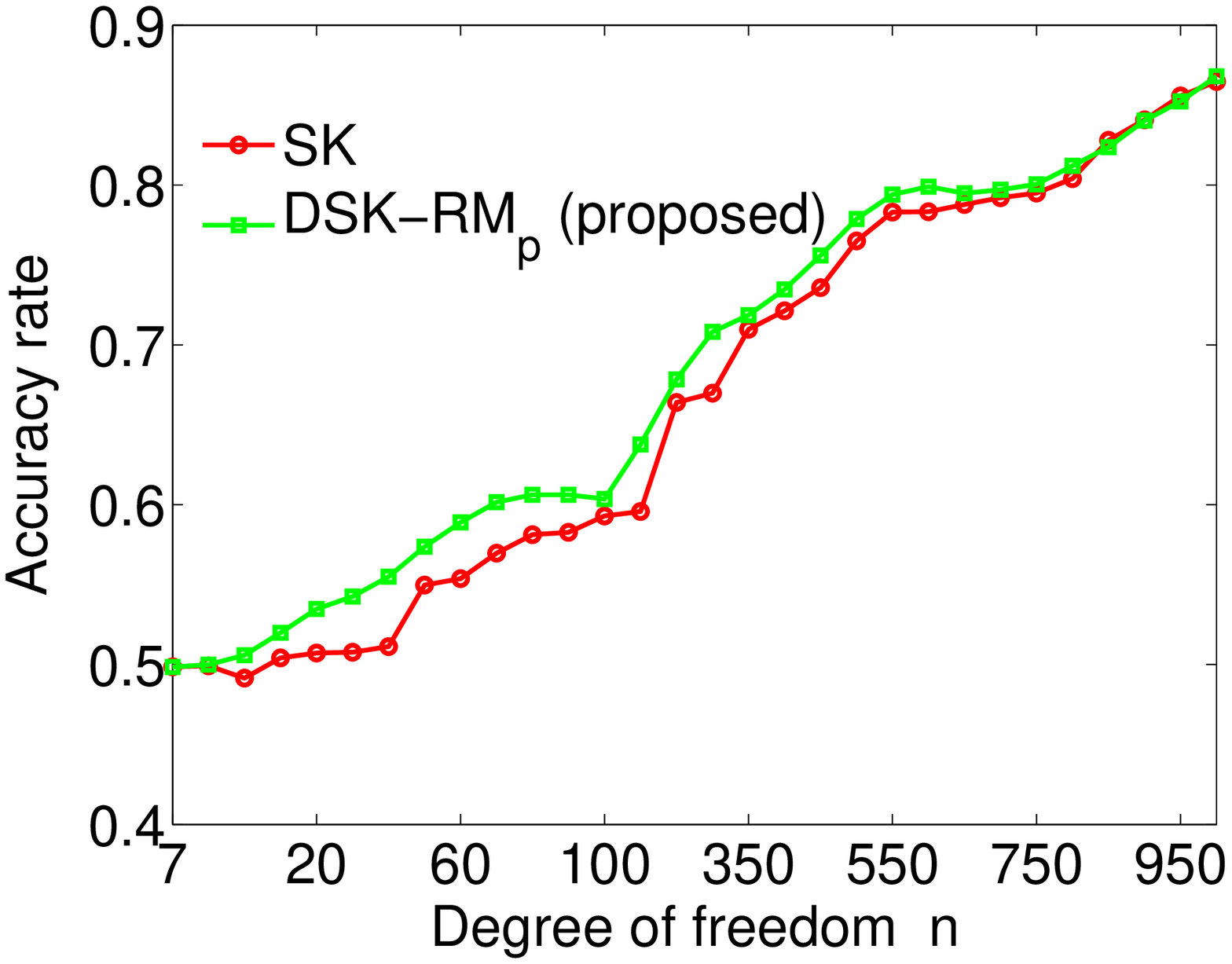}}&
{\includegraphics[width = \fourfigwidth mm]{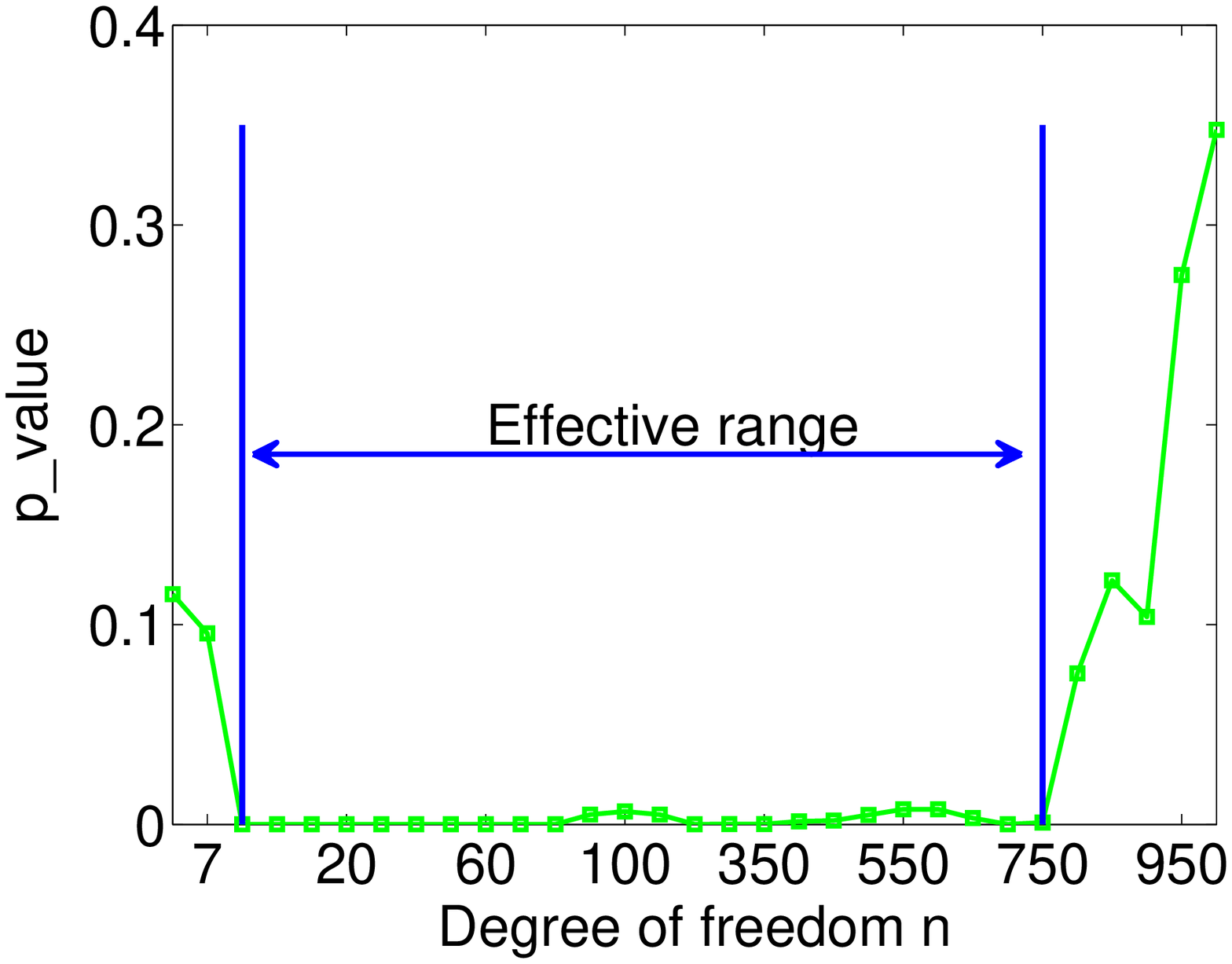}}\\
(a) $n = 200$, varying $\tau$ & (b) $p$-value &(c) $\tau = 10^{-1}$, varying $n$ & (d) $p$-value
\end{tabular}
\end{center}

\caption{Comparison of {DSK} and {SK} on the synthetic data. (a) Performance of DSK and SK with fixed $n = 200$, varying $\tau$. (b) The corresponding $p$-value obtained by the paired Student's $t$-test between DSK and SK with fixed $n = 200$, varying $\tau$. (c) Performance of DSK and SK with fixed $\tau = 10^{-1}$, varying $n$. (d)  The corresponding $p$-value obtained by the paired Student's $t$-test between DSK and SK with fixed $\tau = 10^{-1}$, varying $n$.}
\label{fig:toywhi}
\end{figure*}
\subsection{More insight on when DSK works}
By adjusting the eigenvalues of SPD matrices, DSK can increase the similarity of the SPD matrices within the same class and decrease the similarity of those different classes. We want to gain more insight on in what case DSK can work effectively.
Let $\mathcal{D} = \{\bm{x}_1, \bm{x}_2, \cdots , \bm{x}_n, \bm{x}_{n+1}\}$ denote a set of $d$-dimensional vectors randomly sampled from a normal distribution $\mathcal{N}_d(\boldsymbol{\mu}, \boldsymbol\Sigma)$. It is known that the scatter matrix $\bm S$ follows the Wishart distribution~\cite{tokuda2011visualizing}: $\bm S   \sim \mathcal W_d(\boldsymbol\Sigma, n)$, 
where $\bm S$ is defined as $\sum_{i = 1}^{n + 1}(\bm{x}_i - \bm m) ({\bm{x}_i} - \bm m)^\top$; $\bm{m}$ is defined as $\frac{1}{n + 1}\sum_{i = 1}^{n + 1}\bm{x}_i$; and  $n$ is called the degree of freedom. Increasing the degree of freedom results in a smaller overall variance of $\bm S$~\cite{nydick2012wishart}. 
Note that the features extracted from an image region or an entire image can be considered as a random sample set  $\mathcal{D}$  and the covariance descriptors used in the above experiments can be considered as the samples from certain Wishart distributions. In light of this connection, we use a set of synthetic SPD matrices sampled from various Wishart distributions to investigate the effectiveness of DSK. Specifically, in our experiment, two classes of SPD matrices are obtained by sampling Wishart distributions $W_d(\boldsymbol\Sigma_1, n)$ and $W_d(\boldsymbol\Sigma_2, n)$, where $\boldsymbol\Sigma_1$ is set as a $5 \times 5$ identity matrix $\bm I_{5 \times 5}$ and $\boldsymbol\Sigma_2$ is set as $(1 + \tau)\bm I_{5 \times 5}$. $\tau$ is a small positive scalar and its magnitude controls the difference between $\boldsymbol\Sigma_1$  and $\boldsymbol\Sigma_2$. By varying $\tau$ and the degree of freedom $n$, we can generate a set of binary classification tasks with different levels of classification difficulty to evaluate DSK. First, we set $n$ as $200$ and vary $\tau$ to generate two classes of SPD matrices, with $1000$ samples in each class. Larger $\tau$ will make the classification task easier since it leads to more different distributions. For each classification task, we randomly halve the samples to create $20$ pairs of training/test subsets. Fig.~\ref{fig:toywhi}(a) shows the performance of DSK (DSK-RM$_p$ is used as an example) and SK with respect to $\tau$, while Fig.~\ref{fig:toywhi}(b) reports the corresponding $p$-values between DSK and SK. As seen, DSK can consistently achieve statistically significant ($p$-value $<0.05$) improvement over SK when $10^{- 1.8} \le \tau \le 10^{- 0.83}$. 
When $\tau$ is out of this range, the classification task becomes too difficult or too easy.  In this case, DSK cannot improve the performance of SK.  
Similar results are obtained in Fig.~\ref{fig:toywhi}(c) and Fig.~\ref{fig:toywhi}(d), where we fix $\tau$ as $10^{- 1}$ and change the degree of freedom $n$. As seen, DSK outperforms SK when $8 \le n \le 750$ and has a similar performance as SK otherwise. 

This experiment reveals that DSK can effectively improve the performance of SK for a wide range of classification tasks unless the task is too difficult or too easy. 

\subsection{Comparison between DSK and the methods of improving eigenvalue estimation}
Reshaping the eigenvalues of a sample covariance matrix has been extensively studied to improve the eigenvalue estimation and recover the true covariance matrix, especially when the number of samples is small~\cite{mestreimproved, efronmultivariate, beneigenvalue}. We highlight the differences of the proposed DSK from these methods as follows. i) Handling the biasness of the eigenvalue estimation is only one of our motivations to propose DSK. The other more important motivation is to increase the discrimination of different sets of SPD matrices through eigenvalue adjustment; ii) DSK does not aim at (and is not even interested in) restoring the unbiased estimates of eigenvalues. Instead, DSK adaptively adjusts the eigenvalues of sample covariance matrices in a supervised manner to increase the discriminative power of Stein kernel. 

Although DSK and the methods of improving eigenvalue estimation have different goals, it is still desirable to make a comparison between them in terms of the classification performance. We perform the comparison by using three kinds of SPD matrices: i) the sample covariance matrix; ii) the covariance matrix improved by the methods in \cite{mestreimproved,efronmultivariate,beneigenvalue};  and iii) the covariance matrix obtained by the proposed eigenvalue adjustment in DSK. Stein kernel is used in the classification with SVM as the classifier.
As seen in Table~\ref{tab:cmpdskeig}, the three methods in \cite{mestreimproved,efronmultivariate,beneigenvalue} are comparable to the method using the sample covariance matrices on Brodatz, FERET and ETH80 data sets, while they obtain slightly better performance on fMRI data set. We believe that this is because the improved estimation of eigenvalues may boost the classification performance, when the ratio of sample size to the dimensions of covariance matrix (denoted by $n$/$Dim$ in the table) is small. For example, the ratio $n$/$Dim$ is less than two for the fMRI data set. However, when the ratio becomes larger, the sample covariance matrices will gradually approach to the ground truth, and therefore the methods of improving eigenvalue estimation become less helpful. Since DSK aims to classify different sets of SPD matrices, the eigenvalues are adjusted towards better discriminability. This is why DSK achieves better performance than the methods in \cite{mestreimproved,efronmultivariate,beneigenvalue} on the four data sets. For example, the improvement of DSK is as high as 5.4 and 4.9 percentage points over the method in \cite{efronmultivariate} on Brodatz and FERET data sets.



\begin{table}[hbt]
\caption{Comparison of average classification accuracy (in percentage)  between DSK and the methods of improving eigenvalue estimation.}
\label{tab:cmpdskeig} \centering
\begin{tabular}{|p{18 pt}|p{28 pt}|p{\mediumtablecolumnwidth pt}|p{\mediumtablecolumnwidth pt}|p{\mediumtablecolumnwidth pt}|p{25 pt}|p{\mediumtablecolumnwidth pt}|}
\hline
Data&$n$/$Dim$&sample cov.&\cite{mestreimproved}
&\cite{efronmultivariate}
&\cite{beneigenvalue}
&DSK\\
\hline
Brodatz&1,024/5 $\approx$ 205&78.01 $\pm$ 0.43& 77.50 $\pm$ 0.41&78.00 $\pm$ 0.43&78.00 $\pm$ 0.48&\textbf{83.40} $\pm$ 0.58\\
\hline
FERET&98,304/43 $\approx$ 2286&{79.70} $\pm$ 3.10&78.10 $\pm$ 2.98&{79.70} $\pm$ 3.10&{79.68} $\pm$ 3.10&\textbf{84.60} $\pm$ 1.71\\
\hline
ETH80&16,384/5 $\approx$ 3276&80.30 $\pm$ 0.79&78.80 $\pm$ 0.89&80.30 $\pm$ 0.82&80.31 $\pm$ 0.59&\textbf{82.70} $\pm$ 1.05\\
\hline
fMRI&130/90 $\approx$ 1.44&54.88&54.88&56.10&56.10&\textbf{59.76}\\
\hline
\end{tabular}
\end{table}

\subsection{On the discovery of better SPD kernels}
At last, we discuss what aspects may benefit the discovery of better kernels for SPD matrices. From our point of view, the following two aspects play an important role. \\
\noindent\textit{Distance measure.} A good distance measure should effectively take the underlying structure of SPD matrices into account. In this paper, we utilize the recently developed Stein kernel to meet this requirement. As a specially designed distance measure, the (square-rooted) S-Divergence well respects the Riemannian manifold where SPD matrices reside. The other distance measures, such as Cholesky, Log Euclidean, and Power Euclidean, listed in Table~\ref{tb:spdmetric} could be investigated within our framework in the future work. Also, all the existing distance measures for SPD matrices (except the simplest Euclidean distance) involve matrix decomposition or inverse operation. This results in significant computational cost, especially when a kernel matrix needs to be computed over a large sample set. In the course of discovering better SPD kernels, a computationally efficient distance measure will be highly desirable. \\
\noindent\textit{Class information.} The class information should be effectively integrated into SPD kernels to improve its quality in further. In this work, we achieve this by utilizing the class information to adjust the eigenvalues to make Stein kernel better align with specific classification tasks. For Stein kernel, adjusting the eigenvalues only (rather than including the matrix of eigenvectors) may have been sufficient, because this kernel is invariant to affine transformations\footnote{It means that the S-Divergence is invariant to any nonsingular congruence transformation on the SPD matrices, i,e., $S({\bm X},{\bm Y}) = S({{\bm W}^{\top} \bm X \bm W},{{\bm W}^{\top} \bm Y \bm W})$, where $\bm W$ is an invertible matrix.  The proof is included in the supplementary material.}. There could be different but effective adjustment ways for other types of kernels and this is worth exploring in further. Besides, we focus on improving a SPD kernel in the supervised learning case in this work. Nevertheless, the proposed approach shall be extendable to the unsupervised case, which usually has a wider range of applications. In that situation, how to incorporate cluster information to improve SPD kernels will also be an interesting topic to explore.

In addition, as shown in this work, discovering better SPD kernels may need to optimize certain properties of SPD matrices. In this case, how to design and solve the resulting optimization problem becomes a critical issue. 
In particular, having a convex objective function and a computationally efficient optimization algorithm will be of great importance. This could be possibly achieved by appropriately convexifying and approximating the employed non-convex objective functions. 
In this work, we focus on validating the effectiveness of the proposed approach and demonstrating its advantages, and employ the commonly used gradient-based techniques to solve the involved optimization problems. In our future work, more advanced optimization techniques and algorithms will be developed to improve the proposed approach in further.\\

\section{Conclusion}
In this paper, we analyzed two potential issues of the recently proposed Stein kernel for classification tasks and proposed a novel method called discriminative Stein kernel. It automatically adjusts the eigenvalues of the input SPD matrices to help Stein kernel to achieve greater discrimination. This problem is formulated as a kernel parameter learning process and solved in three frameworks. The proposed kernel is evaluated on both synthetic and real data sets for a variety of applications in pattern analysis and computer vision. The results show that it consistently achieves better performance than the original Stein kernel and other methods for SPD matrices. We also provided more insights on when and how DSK works, discussed the aspects that could contribute to discovering better SPD kernels, and pointed out the future work to enhance the proposed approach.
\bibliographystyle{IEEEtran}
\bibliography{tnn}
\ifCLASSINFOpdf
\else
\fi

\onecolumn
%
\section*{Supplementary Material for:\\ Learning Discriminative Stein Kernel for SPD Matrices and Its Applications}

\section{How to compute the derivative $\frac{\partial\mathbf{K}}{\partial \alpha_z}$}

\begin{equation} \label{eq:kz}
\frac{\partial\bm{K}}{\partial \alpha_z} = \left[\frac{\partial{k({\bm{X}},{\bm{Y}})}}{\partial \alpha_z}\right]_{i,j}
\end{equation}
where 

\begin{equation} \label{eq:kij}
k({\bm{X}},{\bm{Y}}) =  \exp\left(-\theta\cdot S\left({\tilde{\bm X}},{\tilde{\bm Y}}\right)\right). 
\end{equation}
and 
\begin{equation}  \label{eq:sxy}
\begin{split}
 S({\tilde{\mathbf X}},{\tilde{\mathbf Y}}) &= \log\left(\det\left(\frac{{\tilde{\mathbf X}} + {\tilde{\mathbf Y}}}{2}\right)\right) - \frac{1}{2} \log\left(\det({\tilde{\mathbf X}}{\tilde{\mathbf Y}})\right)\\
 &=  \log\left(\det\left(\frac{{\tilde{\mathbf X}} + {\tilde{\mathbf Y}}}{2}\right)\right) - \frac{1}{2} \left[\log\left(\det({\tilde{\mathbf X}})\right) + \log\left(\det({\tilde{\mathbf Y}})\right)\right]
\end{split}
\end{equation}

\begin{equation}  \label{eq:kijz}
\frac{\partial{k({\bm{X}},{\bm{Y}})}}{\partial \alpha_z} = k({\bm{X}},{\bm{Y}}) \cdot -\theta \cdot \frac{\partial{S\left({\tilde{\mathbf X}},{\tilde{\mathbf Y}}\right)}}{\partial \alpha_z}
\end{equation}
where
\begin{equation} \label{eq:sz}
\begin{split}
\frac{\partial{S\left({\tilde{\mathbf X}},{\tilde{\mathbf Y}}\right)}}{\partial \alpha_z}&= \frac{\partial{\log\left(\det\left(\frac{{\tilde{\mathbf X}} + {\tilde{\mathbf Y}}}{2}\right)\right)}}{\partial \alpha_z} - \frac{1}{2} \left[ \frac{\partial{\log\left(\det({\tilde{\mathbf X}})\right)}}{\partial \alpha_z} + \frac{\partial{\log\left(\det({\tilde{\mathbf Y}})\right)}}{\partial \alpha_z}\right]
\end{split}
\end{equation}
and by applying the rule\footnote{Petersen, Kaare Brandt, and Michael Syskind Pedersen. "The matrix cookbook." Technical University of Denmark (2008): 7-15.}:
\begin{equation}
\boxed{
\frac{\partial{\det\left(\mathbf X \right)}}{\partial \alpha} = \det\left( \mathbf X \right)  \tr\left[ \mathbf{X}^{- 1}  \frac{\partial{\mathbf X}}{\partial \alpha}\right]
}
\end{equation}
\begin{equation}\label{eq:detxy}
\begin{split}
\frac{\partial{\log\left(\det\left(\frac{{\tilde{\mathbf X}} + {\tilde{\mathbf Y}}}{2}\right)\right)}}{\partial \alpha_z} &= \frac{1}{\det\left(\frac{{\tilde{\mathbf X}} + {\tilde{\mathbf Y}}}{2}\right)}\det\left(\frac{{\tilde{\mathbf X}} + {\tilde{\mathbf Y}}}{2}\right) \tr\left[ \left( \frac{{\tilde{\mathbf X}} + {\tilde{\mathbf Y}}}{2} \right)^{-1}\cdot\frac{1}{2}\cdot \left( \frac{\partial{\tilde{\mathbf X}}}{\partial \alpha_z} + \frac{\partial{\tilde{\mathbf Y}}}{\partial \alpha_z}\right)\right]\\
&= \frac{1}{2}\cdot\tr\left[ \left( \frac{{\tilde{\mathbf X}} + {\tilde{\mathbf Y}}}{2} \right)^{-1}\cdot \left( \frac{\partial{\tilde{\mathbf X}}}{\partial \alpha_z} + \frac{\partial{\tilde{\mathbf Y}}}{\partial \alpha_z}\right)\right]
\end{split}
\end{equation}
\begin{equation}\label{eq:detx}
\begin{split}
\frac{\partial{\log\left(\det\left({{\tilde{\mathbf X}}}\right)\right)}}{\partial \alpha_z} = \frac{1}{\det\left({{\tilde{\mathbf X}}}\right)}\det\left({{\tilde{\mathbf X}}}\right) \tr\left[ {{\tilde{\mathbf X}}}^{-1}\cdot \left( \frac{\partial{\tilde{\mathbf X}}}{\partial \alpha_z}\right)\right] =\tr\left[ {{\tilde{\mathbf X}}}^{-1}\cdot \left( \frac{\partial{\tilde{\mathbf X}}}{\partial \alpha_z}\right)\right]
\end{split}
\end{equation}
\begin{equation}\label{eq:dety}
\begin{split}
\frac{\partial{\log\left(\det\left({{\tilde{\mathbf Y}}}\right)\right)}}{\partial \alpha_z} = \frac{1}{\det\left({{\tilde{\mathbf Y}}}\right)}\det\left({{\tilde{\mathbf Y}}}\right) \tr\left[ {{\tilde{\mathbf Y}}}^{-1}\cdot \left( \frac{\partial{\tilde{\mathbf Y}}}{\partial \alpha_z}\right)\right] =\tr\left[ {{\tilde{\mathbf Y}}}^{-1}\cdot \left( \frac{\partial{\tilde{\mathbf Y}}}{\partial \alpha_z}\right)\right]
\end{split}
\end{equation}
Substituting Eq.~(\ref{eq:detxy}), Eq.~(\ref{eq:detx}) and Eq.~(\ref{eq:dety}) in Eq.~(\ref{eq:sz}), $\frac{\partial{k({\bm{X}},{\bm{Y}})}}{\partial \alpha_z}$ can be expressed as:
\begin{equation}  \label{eq:final}
\begin{split}
\frac{\partial{k({\bm{X}},{\bm{Y}})}}{\partial \alpha_z} &= k({\bm{X}},{\bm{Y}}) \cdot -\theta \cdot \frac{\partial{S\left({\tilde{\mathbf X}},{\tilde{\mathbf Y}}\right)}}{\partial \alpha_z}\\
&= \frac{\theta}{2} \cdot k({\bm{X}},{\bm{Y}}) \cdot \left\{ \tr\left[ {{\tilde{\mathbf X}}}^{-1}\cdot \left( \frac{\partial{\tilde{\mathbf X}}}{\partial \alpha_z}\right)\right] + \tr\left[ {{\tilde{\mathbf Y}}}^{-1}\cdot \left( \frac{\partial{\tilde{\mathbf Y}}}{\partial \alpha_z}\right)\right] -\tr\left[ \left( \frac{{\tilde{\mathbf X}} + {\tilde{\mathbf Y}}}{2} \right)^{-1}\cdot \left( \frac{\partial{\tilde{\mathbf X}}}{\partial \alpha_z} + \frac{\partial{\tilde{\mathbf Y}}}{\partial \alpha_z}\right)\right]\right\}
\end{split}
\end{equation}

In the case of power:
\begin{displaymath} \label{eq:mmatrixp}
\begin{split}
\tilde{\bm{X}}_{p}&=\bm{U}
\left( \begin{array}{ccccc}
\lambda_1^{\alpha_1}  & & & &\\
& \ddots & & &\\
  & & \lambda_z^{\alpha_z} & &\\
  & & &\ddots &\\
  & & & &\lambda_d^{\alpha_d}\\
\end{array} \right)
\bm{U^\top}
\end{split},
\hspace{2em}
\begin{split}
\frac{\partial{\tilde{\bm X}}_{p}}{\partial \alpha_z} &=\bm{U}
\left( \begin{array}{ccccc}
0  & & & &\\
& \ddots & & &\\
  & & \ln(\lambda_z)\lambda_z^{\alpha_z} & &\\
  & & &\ddots &\\
  & & & &0\\
\end{array} \right)
\bm{U^\top}
\end{split}
\end{displaymath} 

In the case of coefficient:
\begin{displaymath} \label{eq:mmatrixc}
\begin{split}
\tilde{\bm{X}}_{c}&=\bm{U}
\left( \begin{array}{ccccc}
{\alpha_1}\lambda_1  & & & &\\
& \ddots & & &\\
  & & {\alpha_z}\lambda_z & &\\
  & & &\ddots &\\
  & & & &{\alpha_d}\lambda_d\\
\end{array} \right)
\bm{U^\top}
\end{split},
\hspace{2em}
\begin{split}
\frac{\partial{\tilde{\bm X}}_{c}}{\partial \alpha_z} &=\bm{U}
\left( \begin{array}{ccccc}
0  & & & &\\
& \ddots & & &\\
  & & \lambda_z & &\\
  & & &\ddots &\\
  & & & &0\\
\end{array} \right)
\bm{U^\top}
\end{split}
\end{displaymath} 

\newpage

\section{The most difficult $15$ pairs of texture classes from Brodatz texture data set}
\begin{figure}[!ht]
\begin{center}
\renewcommand{\arraystretch}{0.5}
\begin{tabular}{ccccc}
{\includegraphics[width = \figwidths mm,height = \figheights mm]{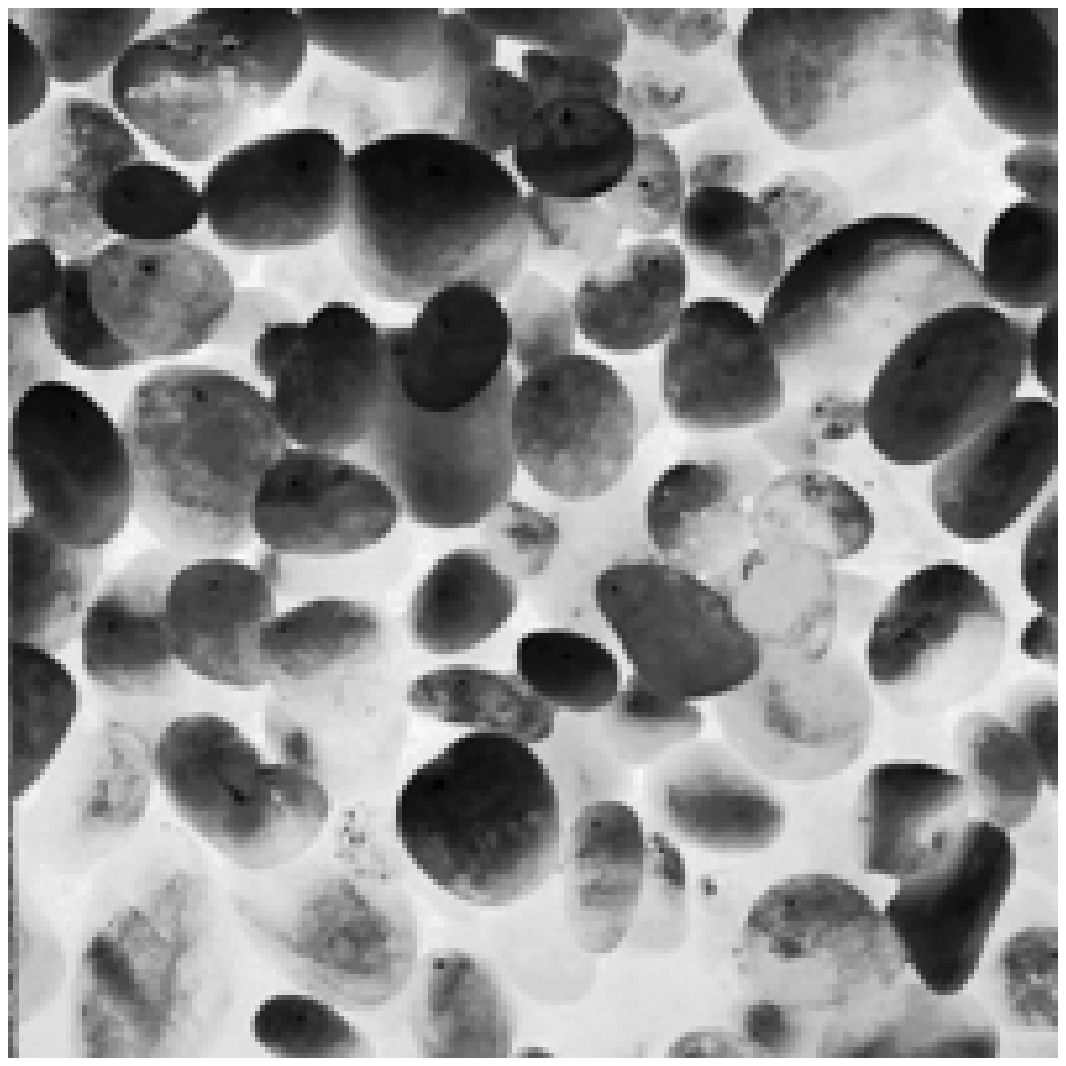}}&
{\includegraphics[width = \figwidths mm,height = \figheights mm]{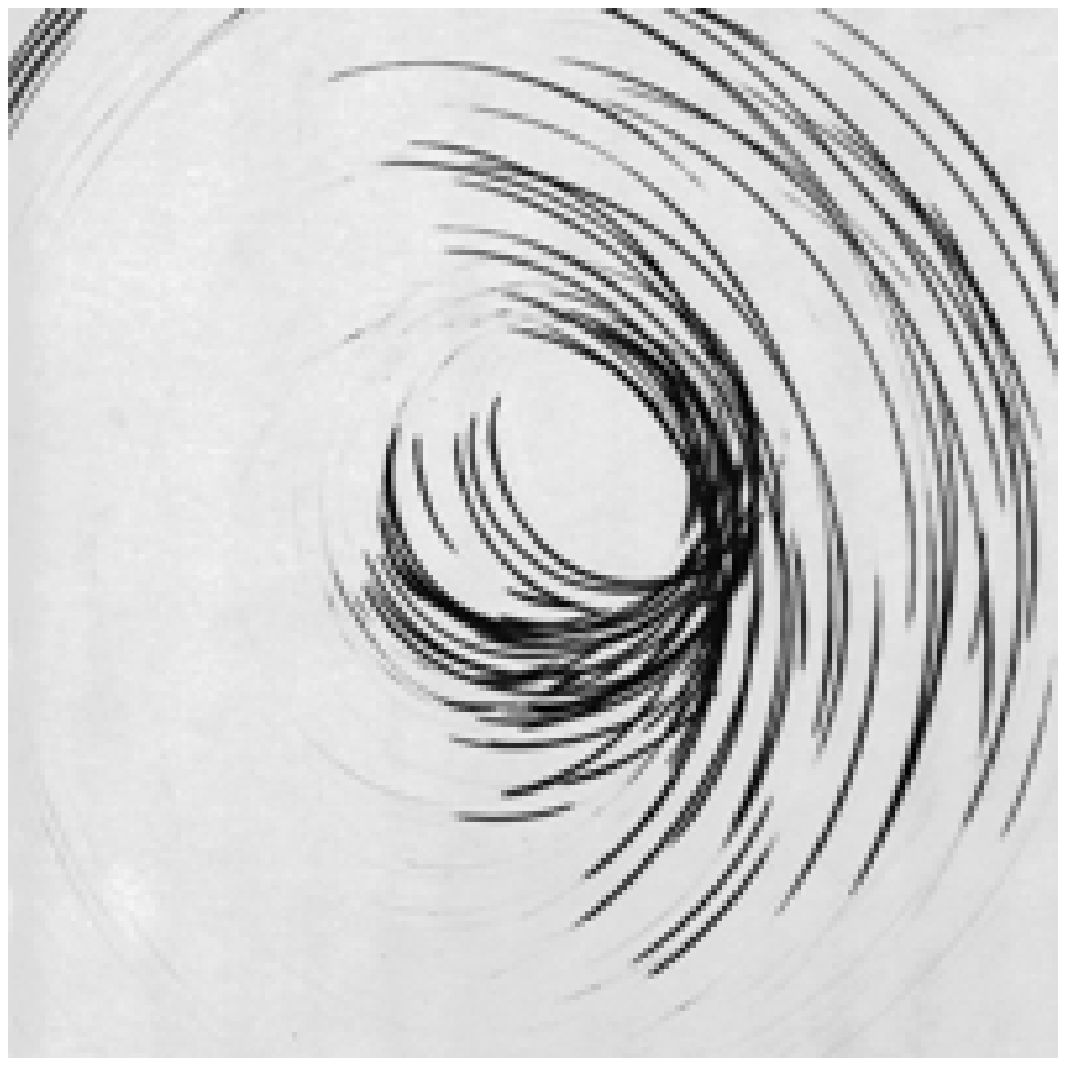}}&
{\includegraphics[width = \figwidths mm,height = \figheights mm]{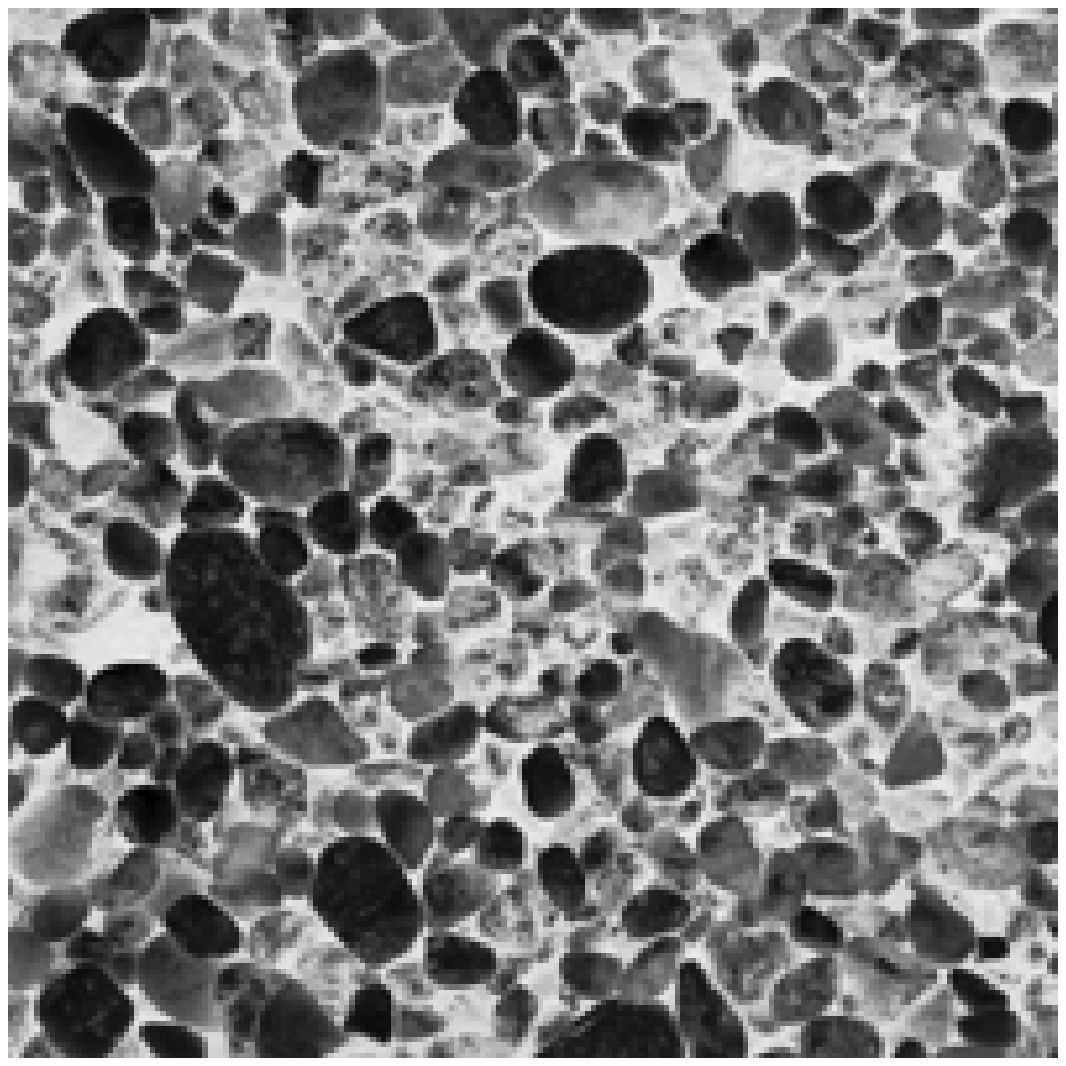}}&
{\includegraphics[width = \figwidths mm,height = \figheights mm]{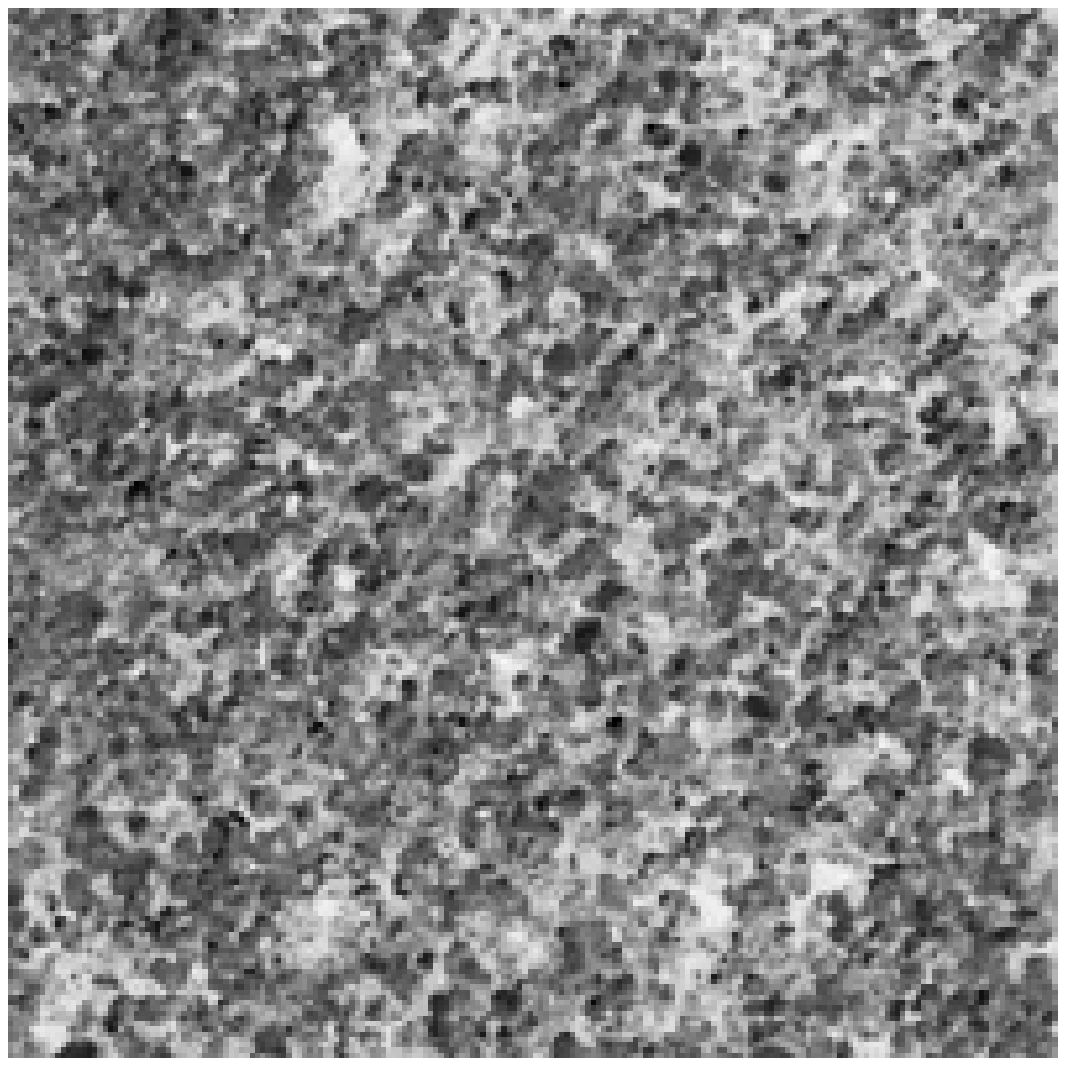}}&
{\includegraphics[width = \figwidths mm,height = \figheights mm]{D30}}\\ 
{\includegraphics[width = \figwidths mm,height = \figheights mm]{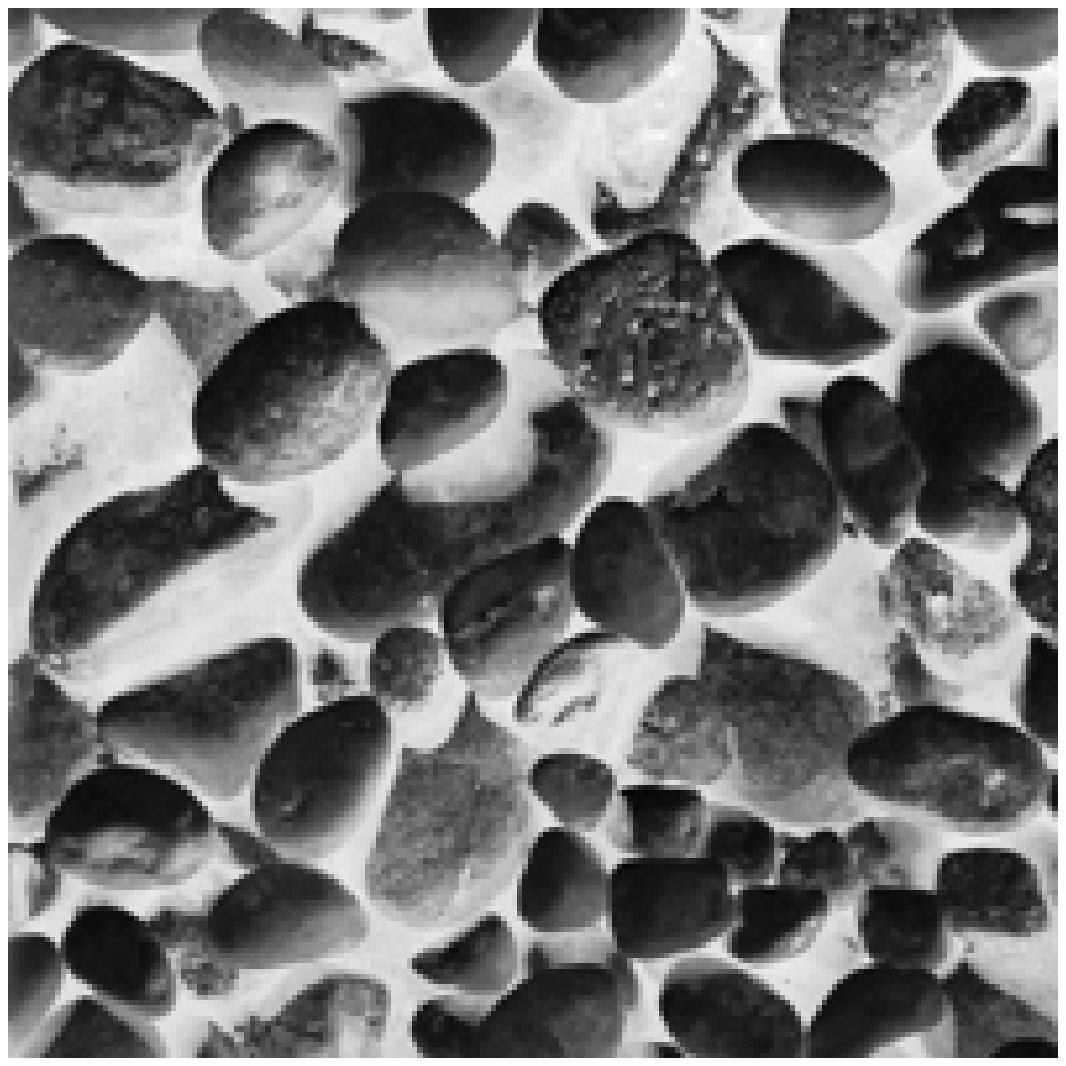}}&
{\includegraphics[width = \figwidths mm,height = \figheights mm]{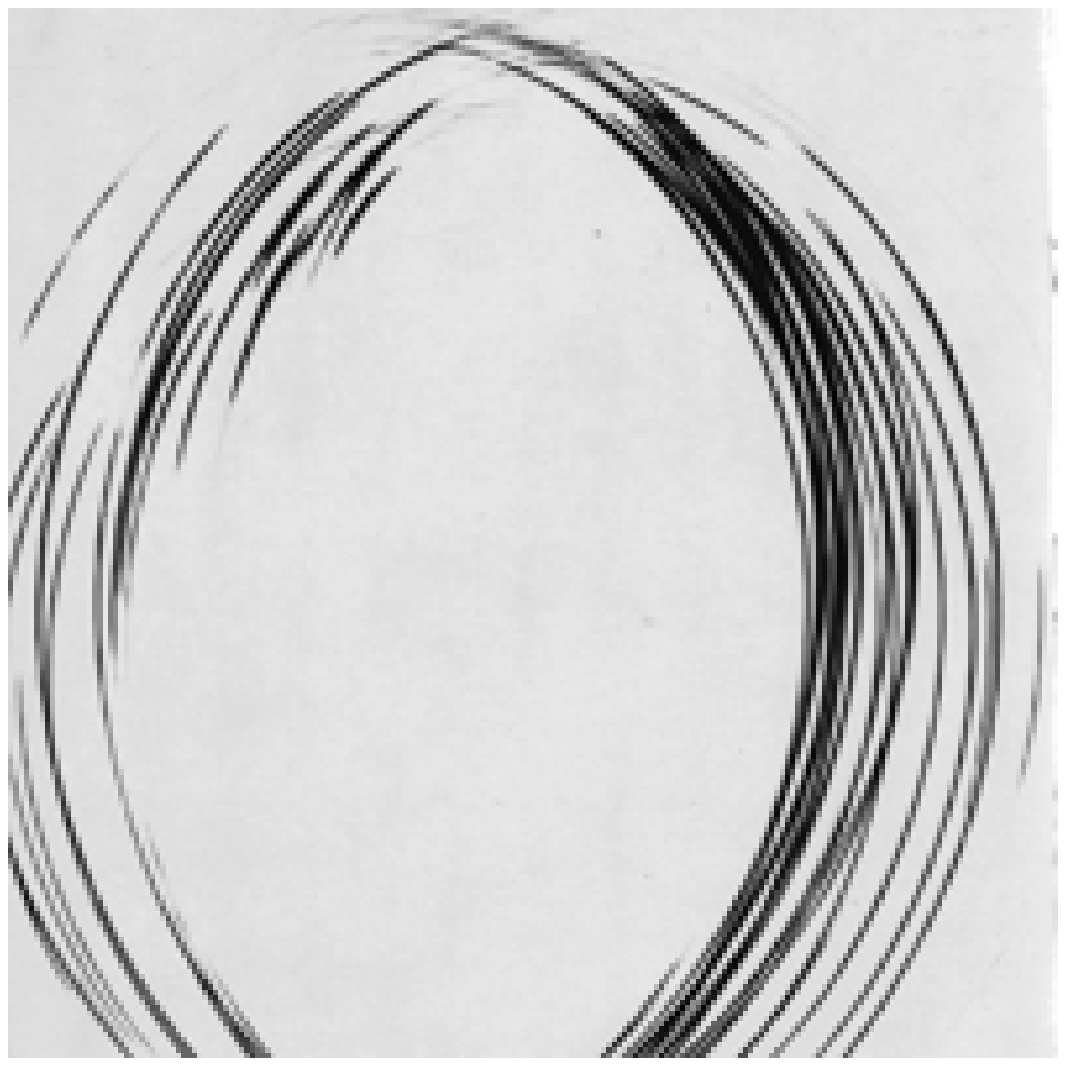}}&
{\includegraphics[width = \figwidths mm,height = \figheights mm]{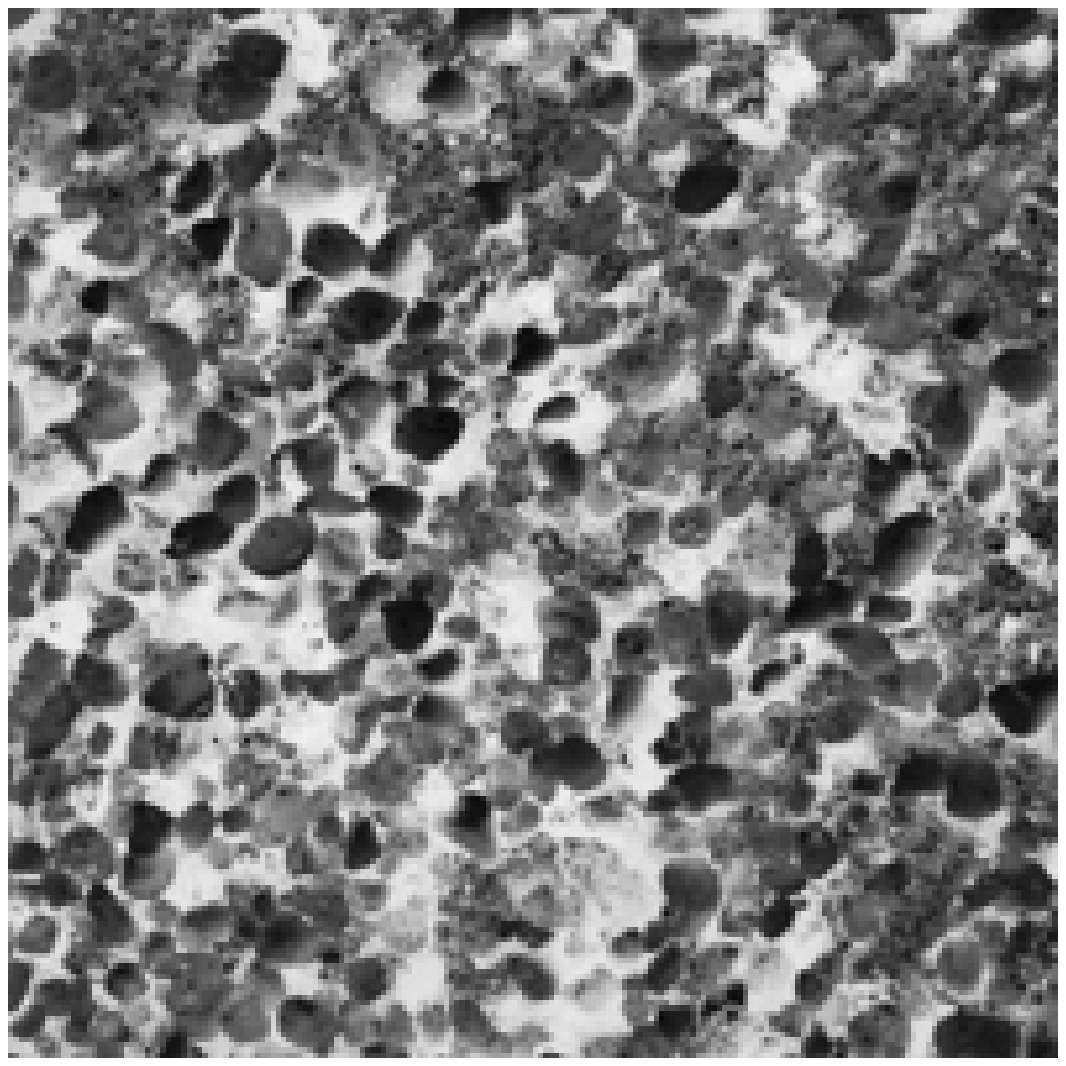}}&
{\includegraphics[width = \figwidths mm,height = \figheights mm]{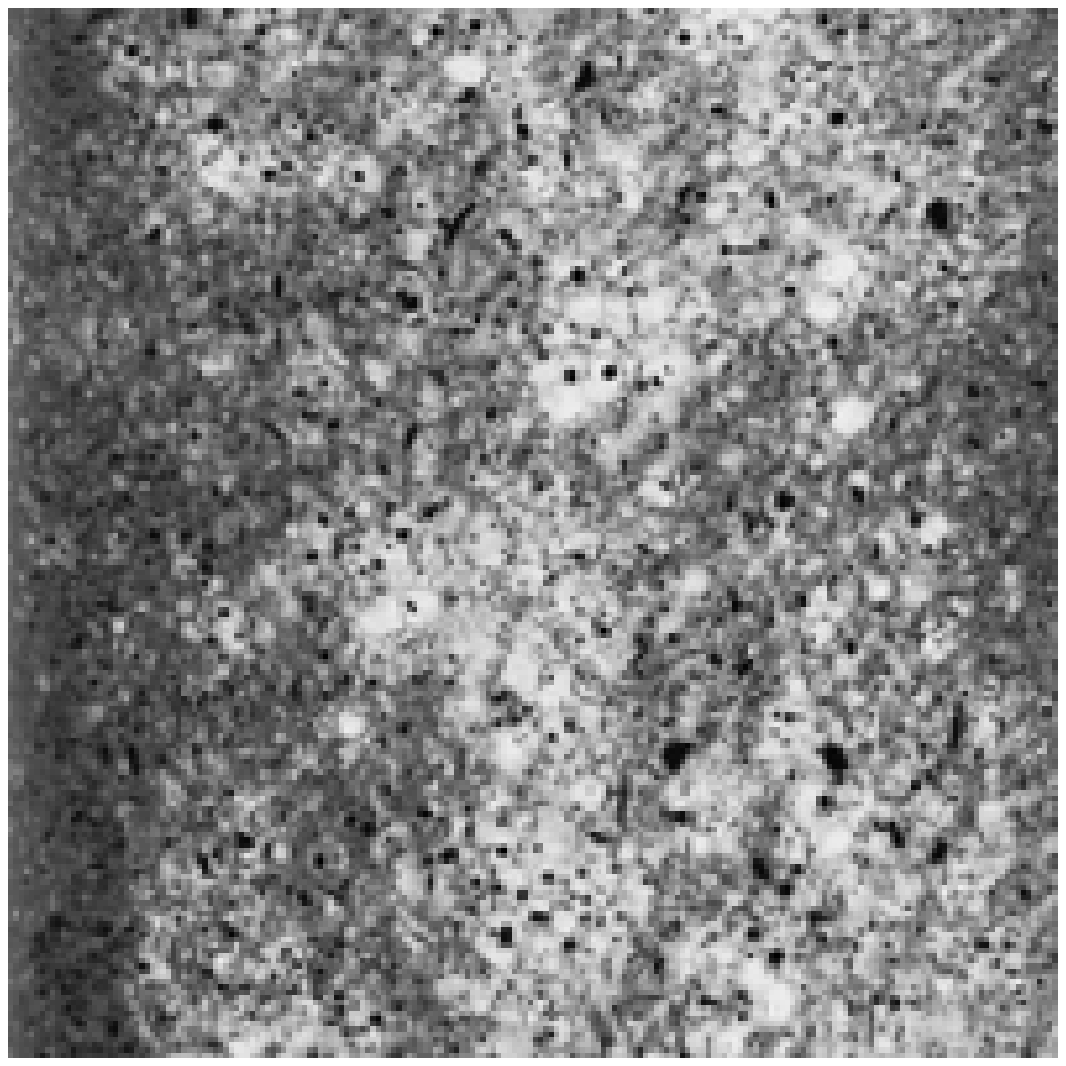}}&
{\includegraphics[width = \figwidths mm,height = \figheights mm]{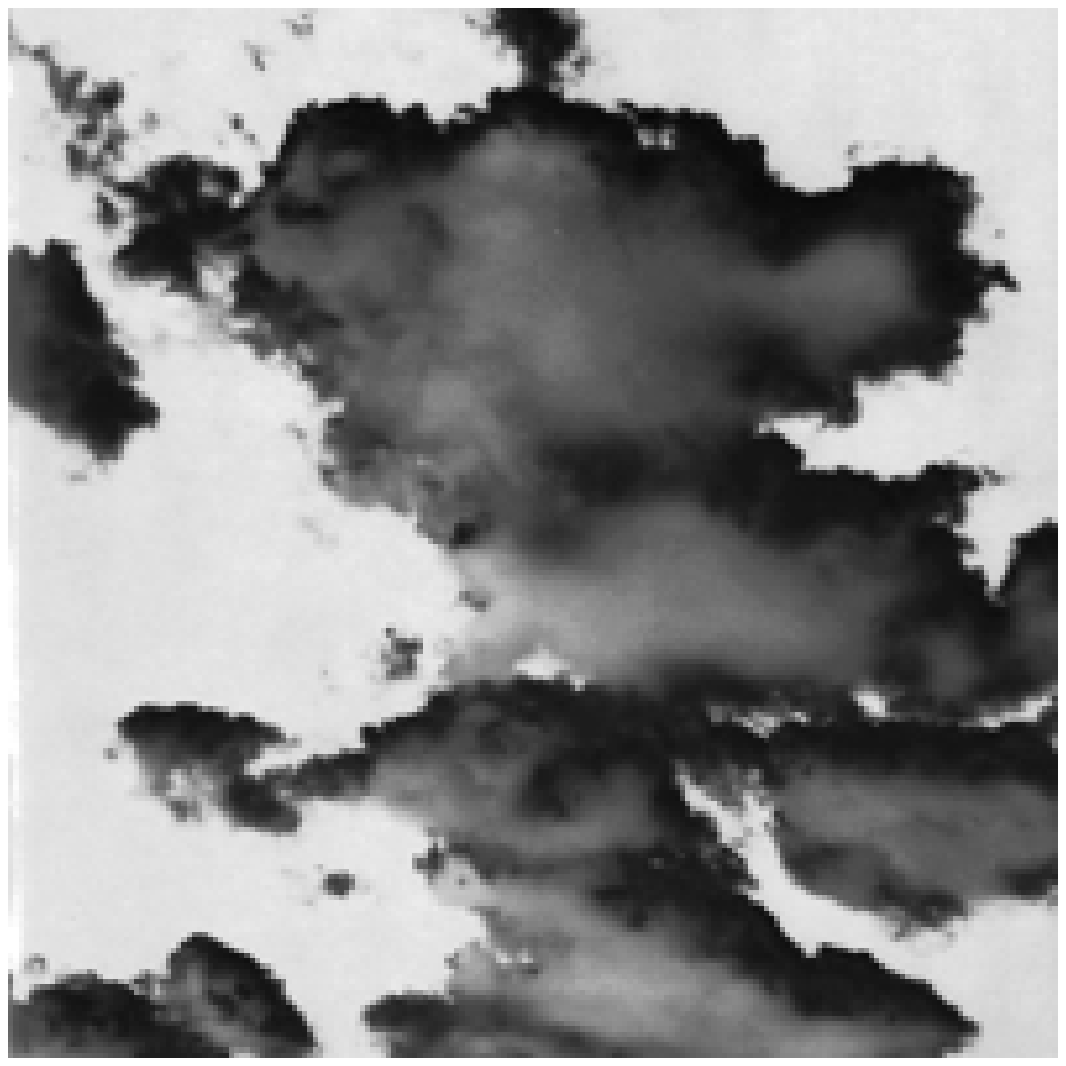}}\\ 
{(a) 30 vs. 31} & {(b) 43 vs. 44}& {(c) 23 vs. 27}& {(d) 28 vs. 73}& {e) 30 vs. 91}\\
{\includegraphics[width = \figwidths mm,height = \figheights mm]{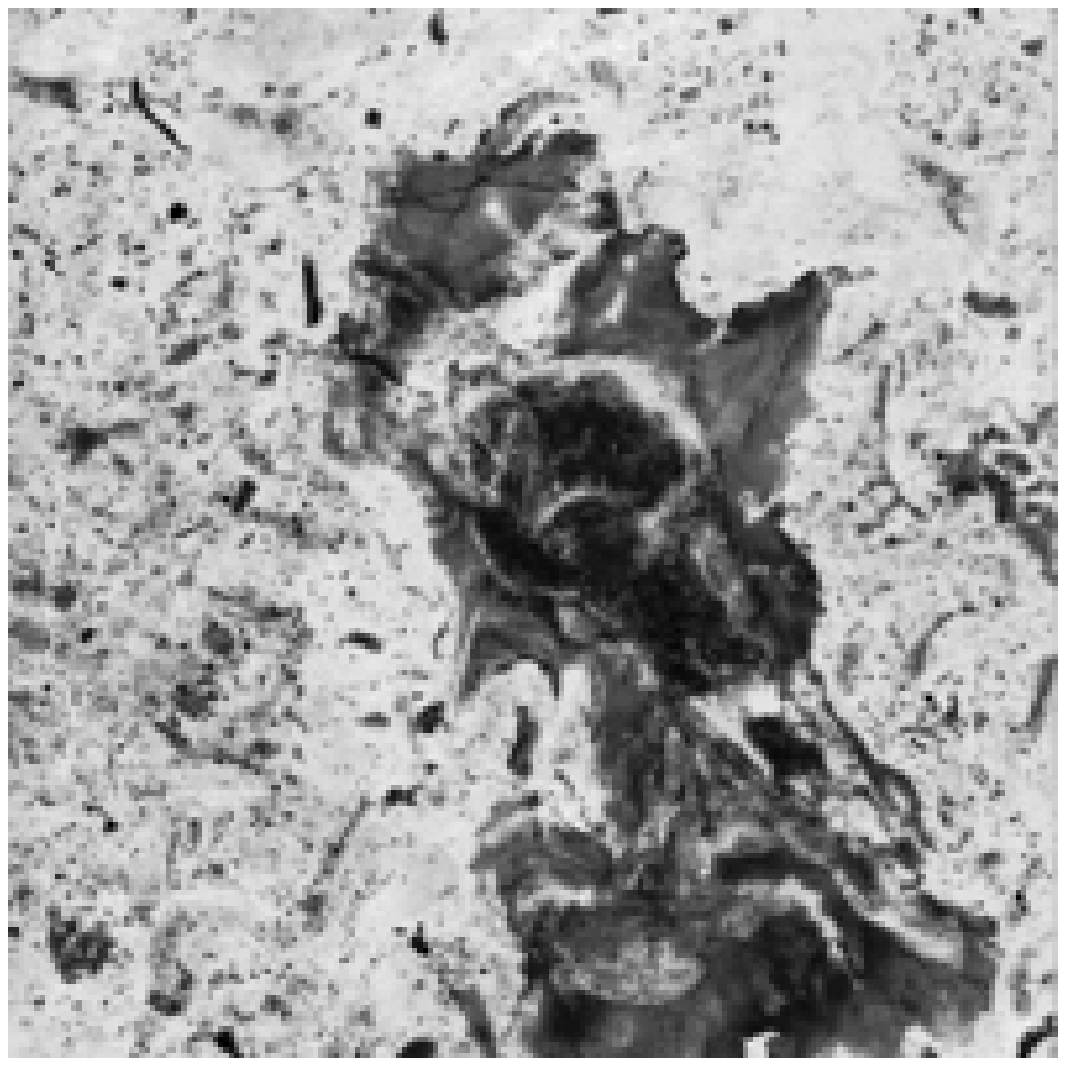}}&
{\includegraphics[width = \figwidths mm,height = \figheights mm]{D30}}&
{\includegraphics[width = \figwidths mm,height = \figheights mm]{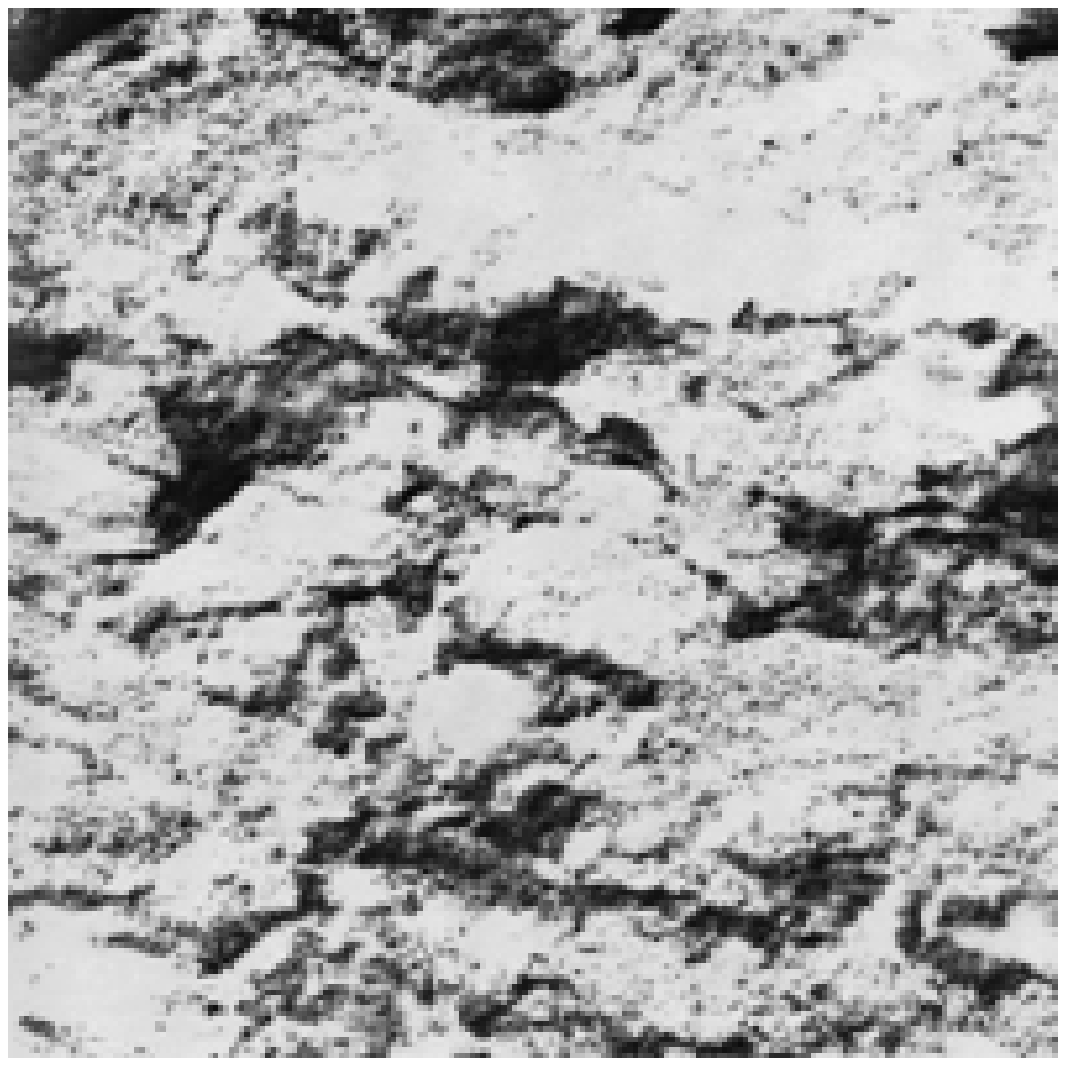}}&
{\includegraphics[width = \figwidths mm,height = \figheights mm]{D31}}&
{\includegraphics[width = \figwidths mm,height = \figheights mm]{D31}}\\ 
{\includegraphics[width = \figwidths mm,height = \figheights mm]{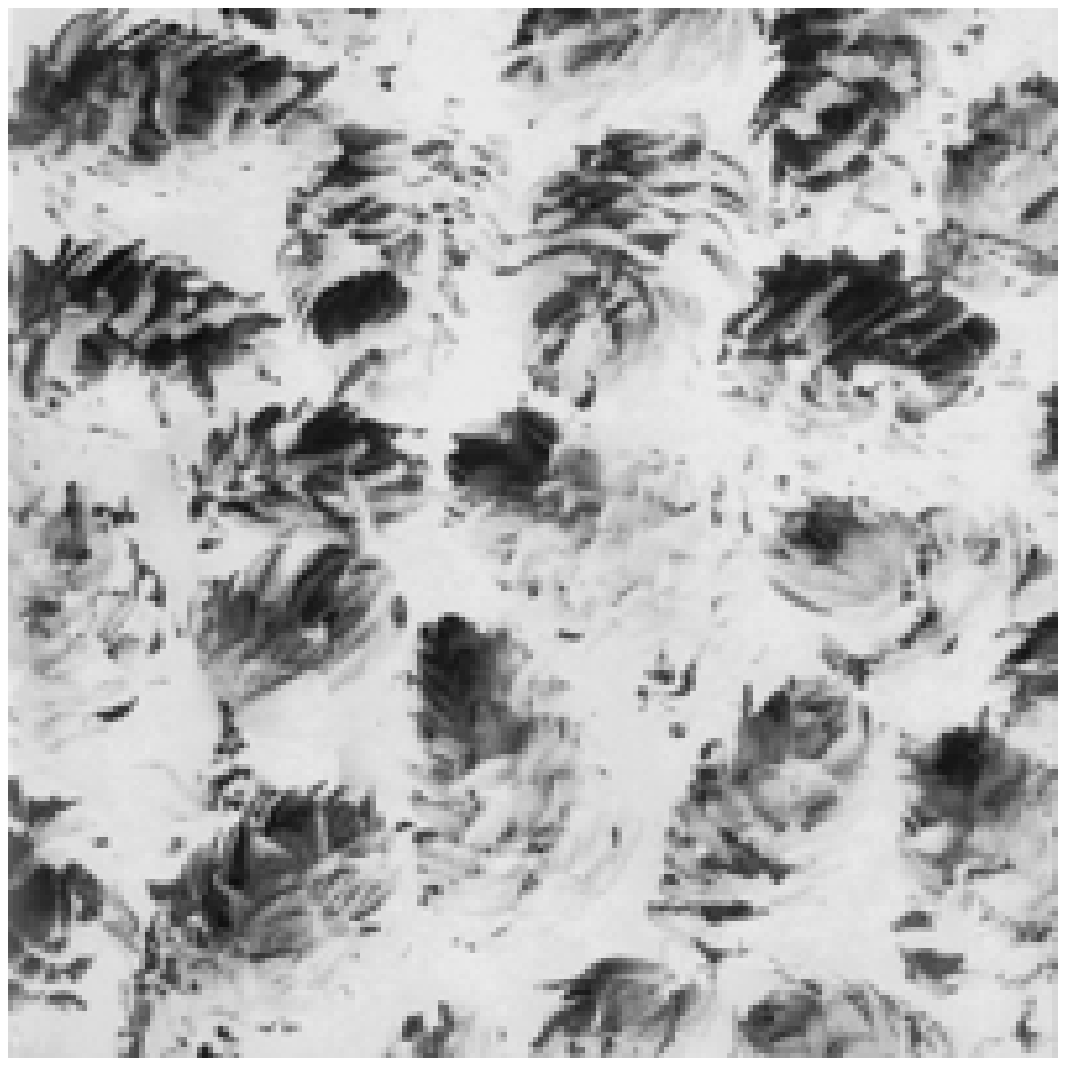}}&
{\includegraphics[width = \figwidths mm,height = \figheights mm]{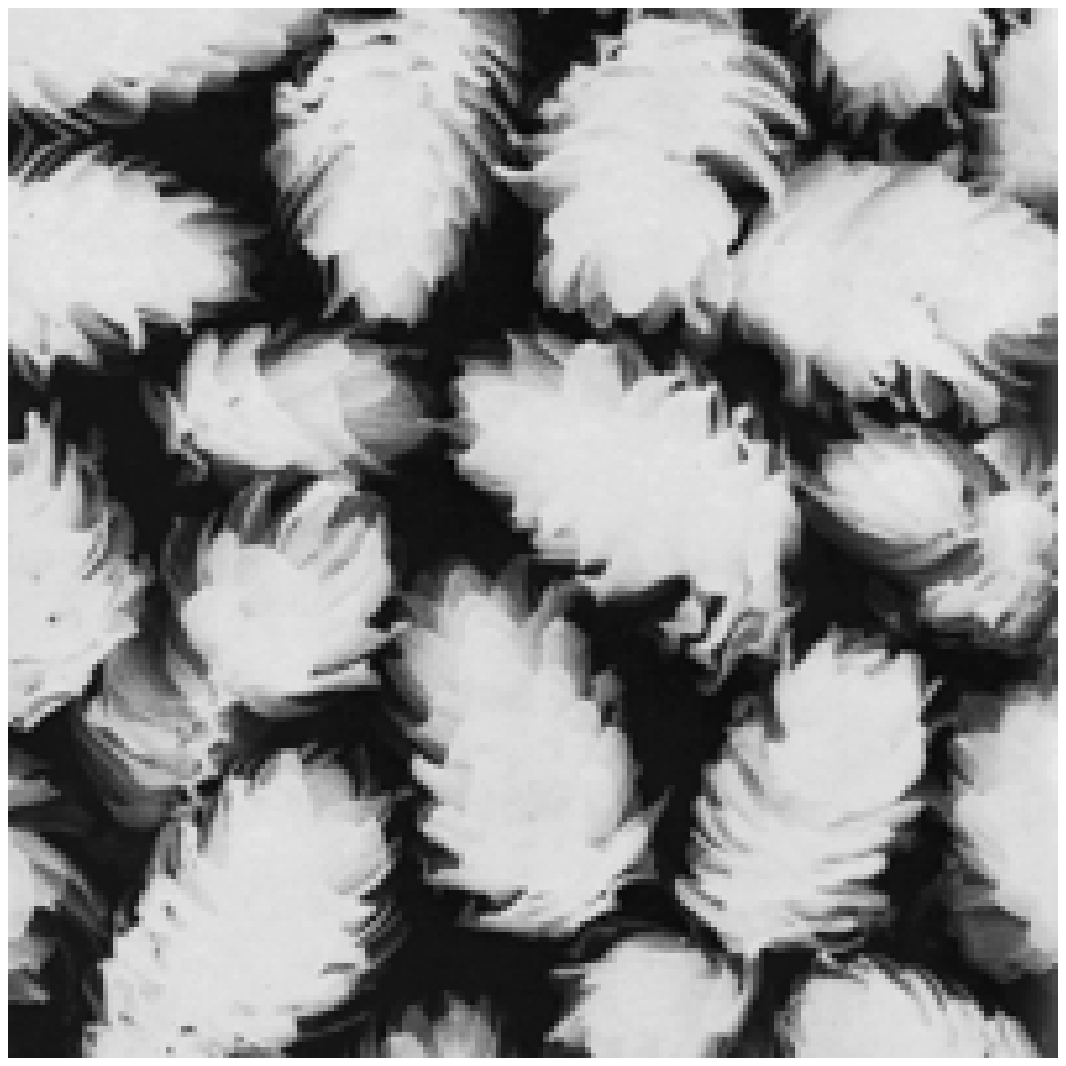}}&
{\includegraphics[width = \figwidths mm,height = \figheights mm]{D89}}&
{\includegraphics[width = \figwidths mm,height = \figheights mm]{D91}}&
{\includegraphics[width = \figwidths mm,height = \figheights mm]{D88}}\\ 
{(f) 58 vs. 89} & {(g) 30 vs. 88}& {(h) 7 vs. 89}& {(i) 31 vs. 91}& {(j) 31 vs. 88}\\
{\includegraphics[width = \figwidths mm,height = \figheights mm]{D30}}&
{\includegraphics[width = \figwidths mm,height = \figheights mm]{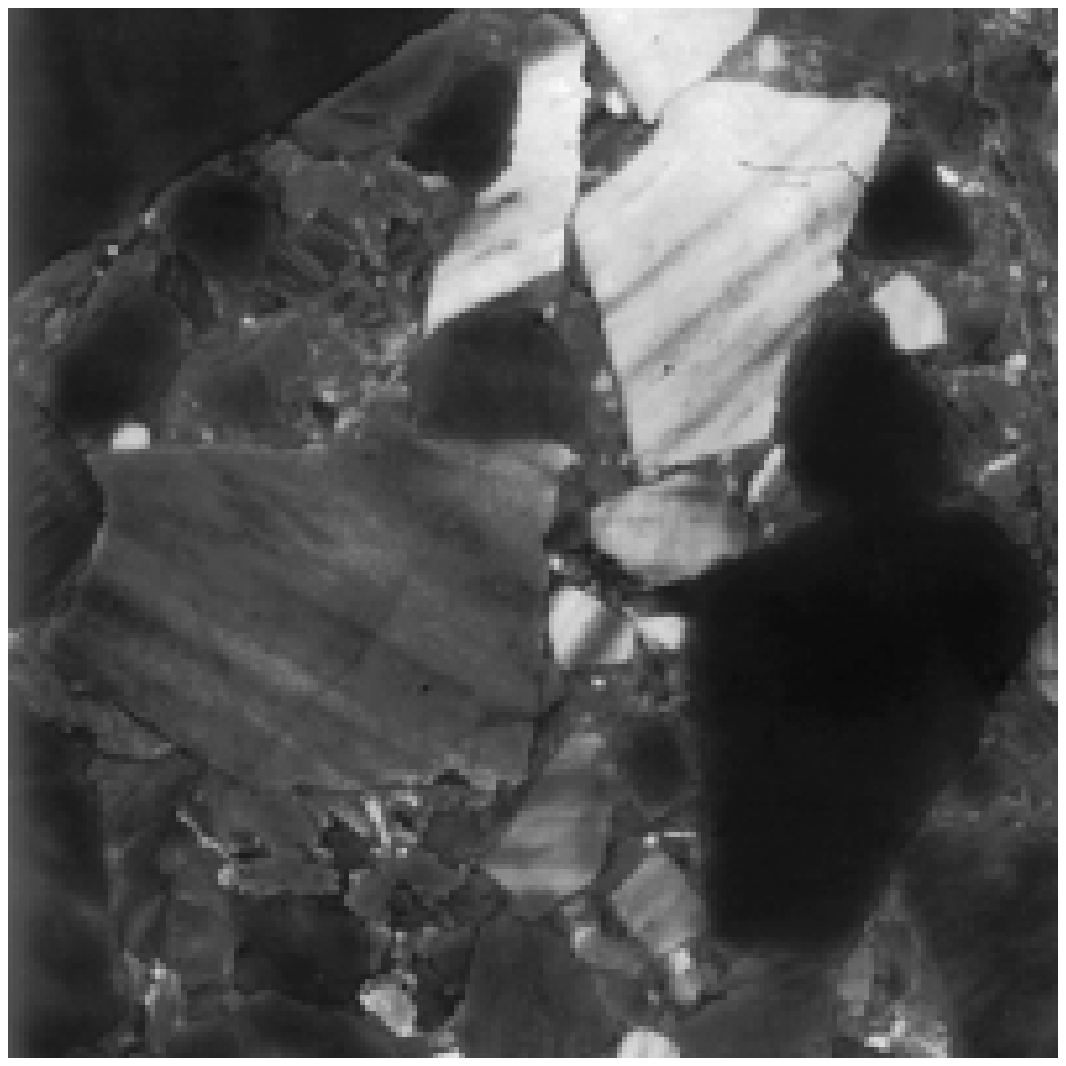}}&
{\includegraphics[width = \figwidths mm,height = \figheights mm]{D27}}&
{\includegraphics[width = \figwidths mm,height = \figheights mm]{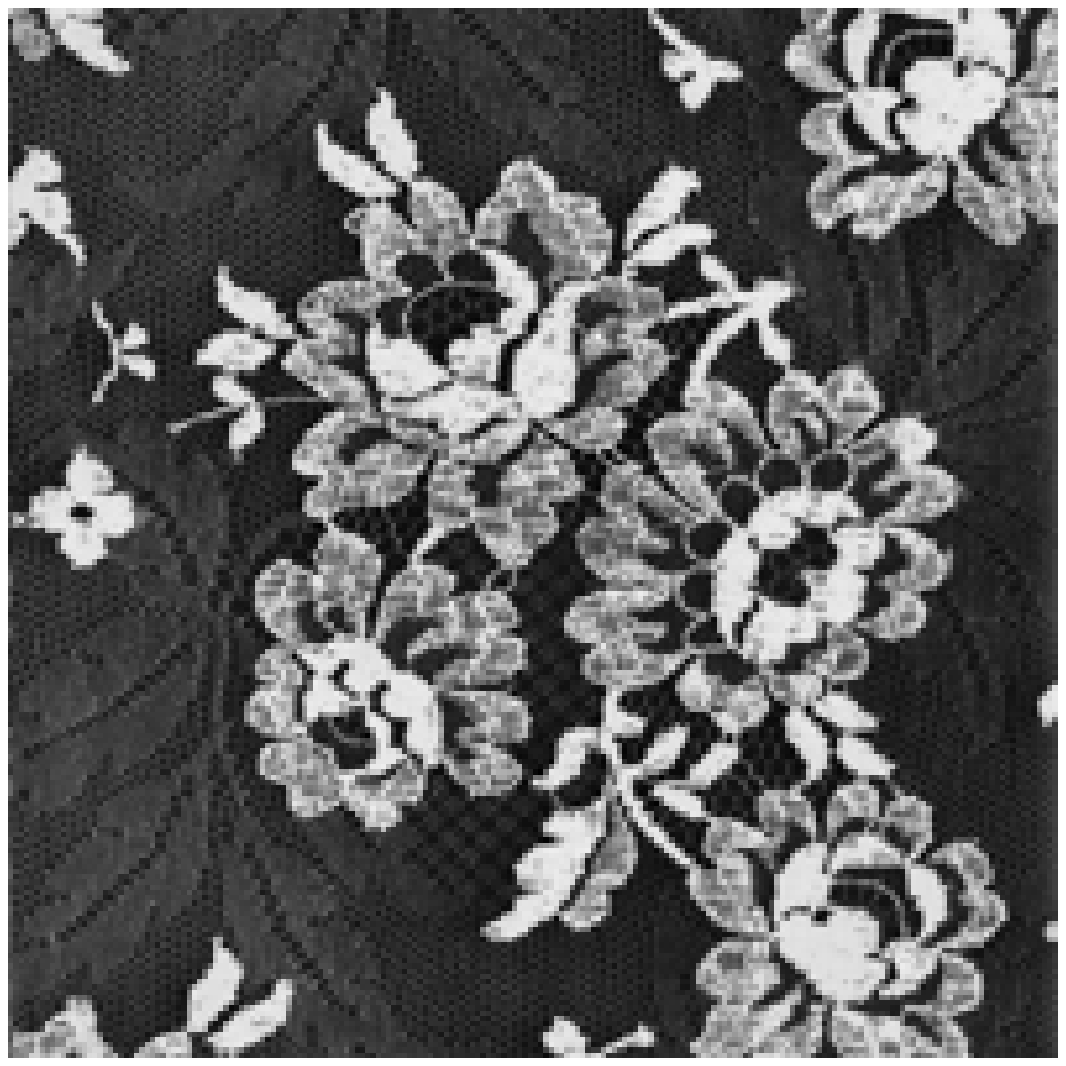}}&
{\includegraphics[width = \figwidths mm,height = \figheights mm]{D31}}\\ 
{\includegraphics[width = \figwidths mm,height = \figheights mm]{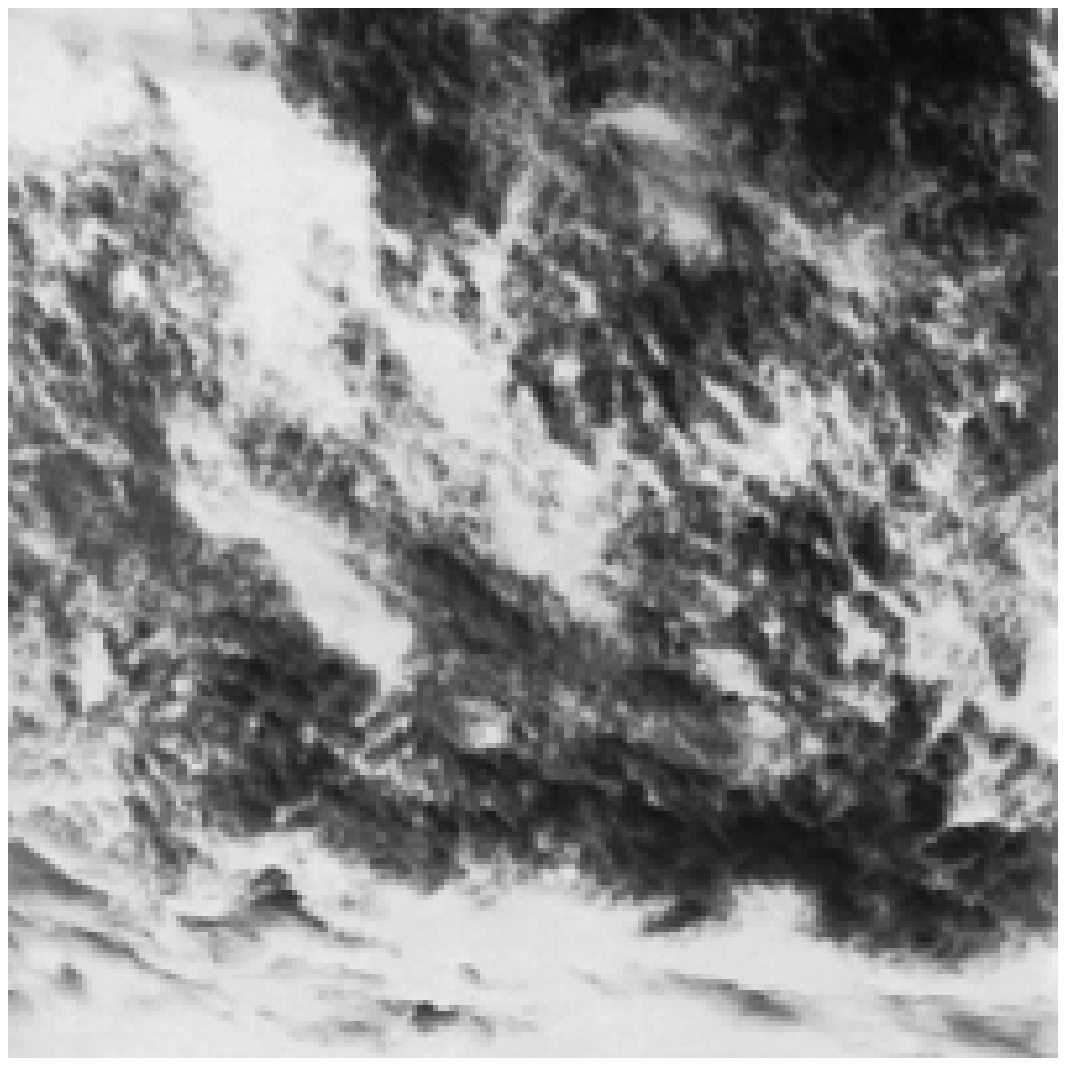}}&
{\includegraphics[width = \figwidths mm,height = \figheights mm]{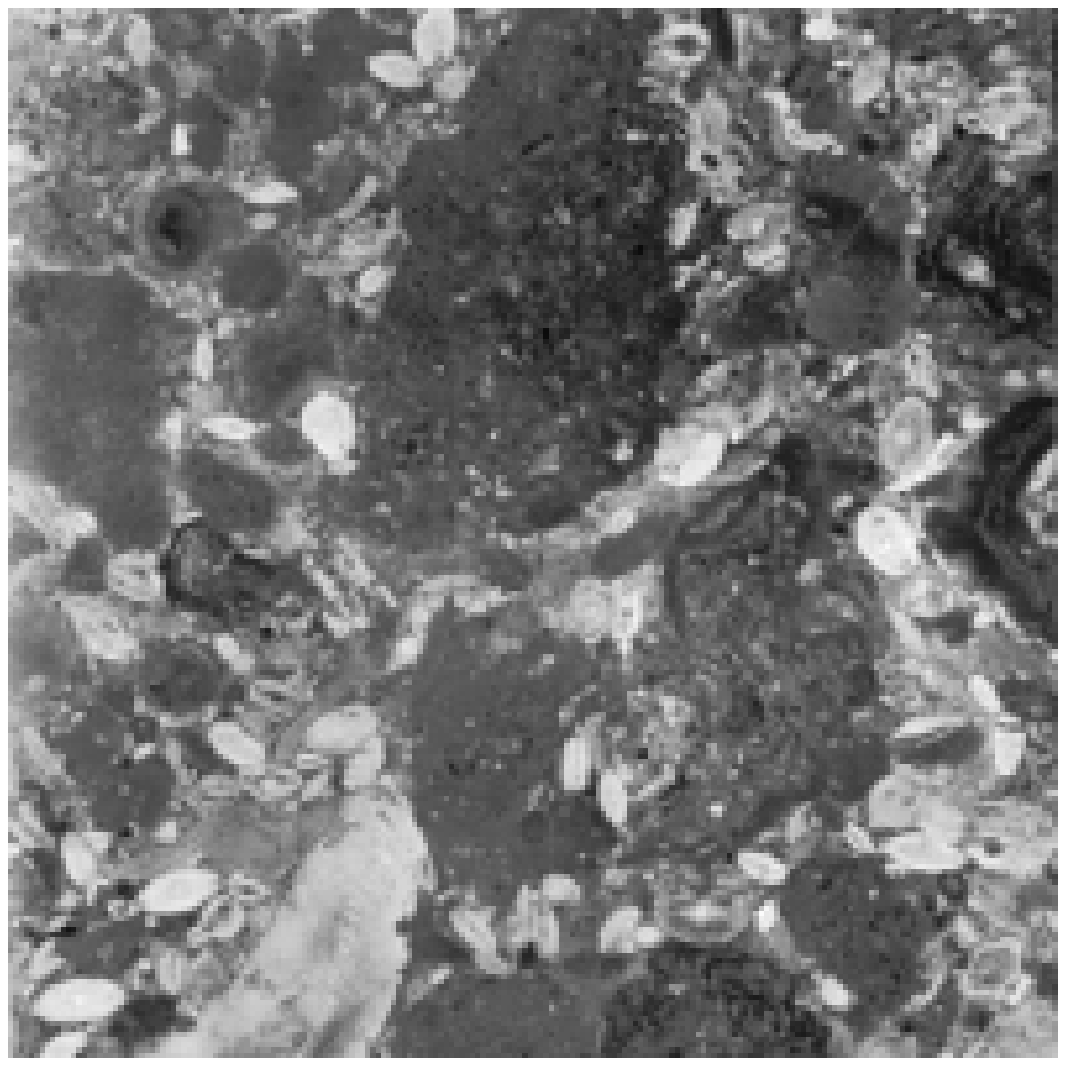}}&
{\includegraphics[width = \figwidths mm,height = \figheights mm]{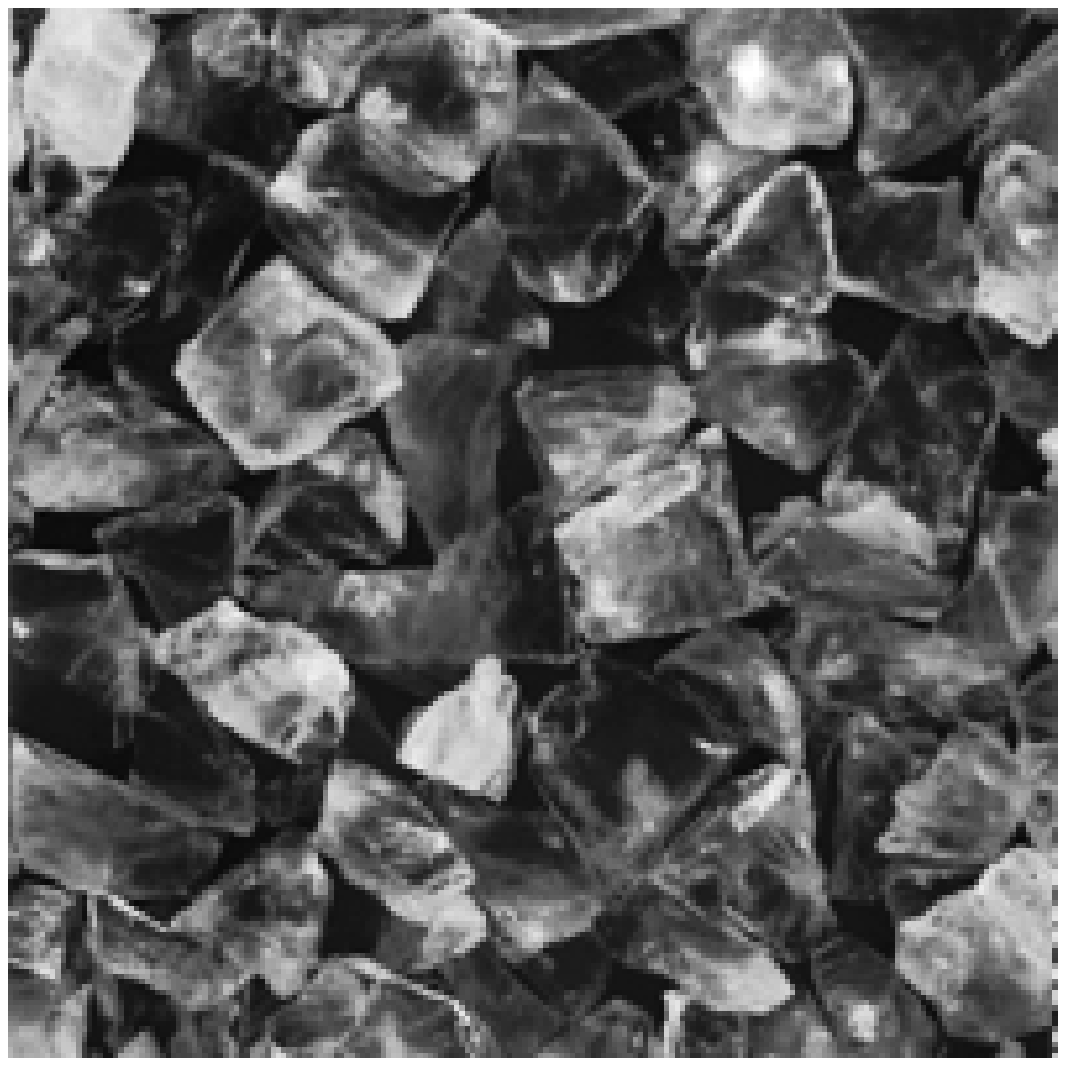}}&
{\includegraphics[width = \figwidths mm,height = \figheights mm]{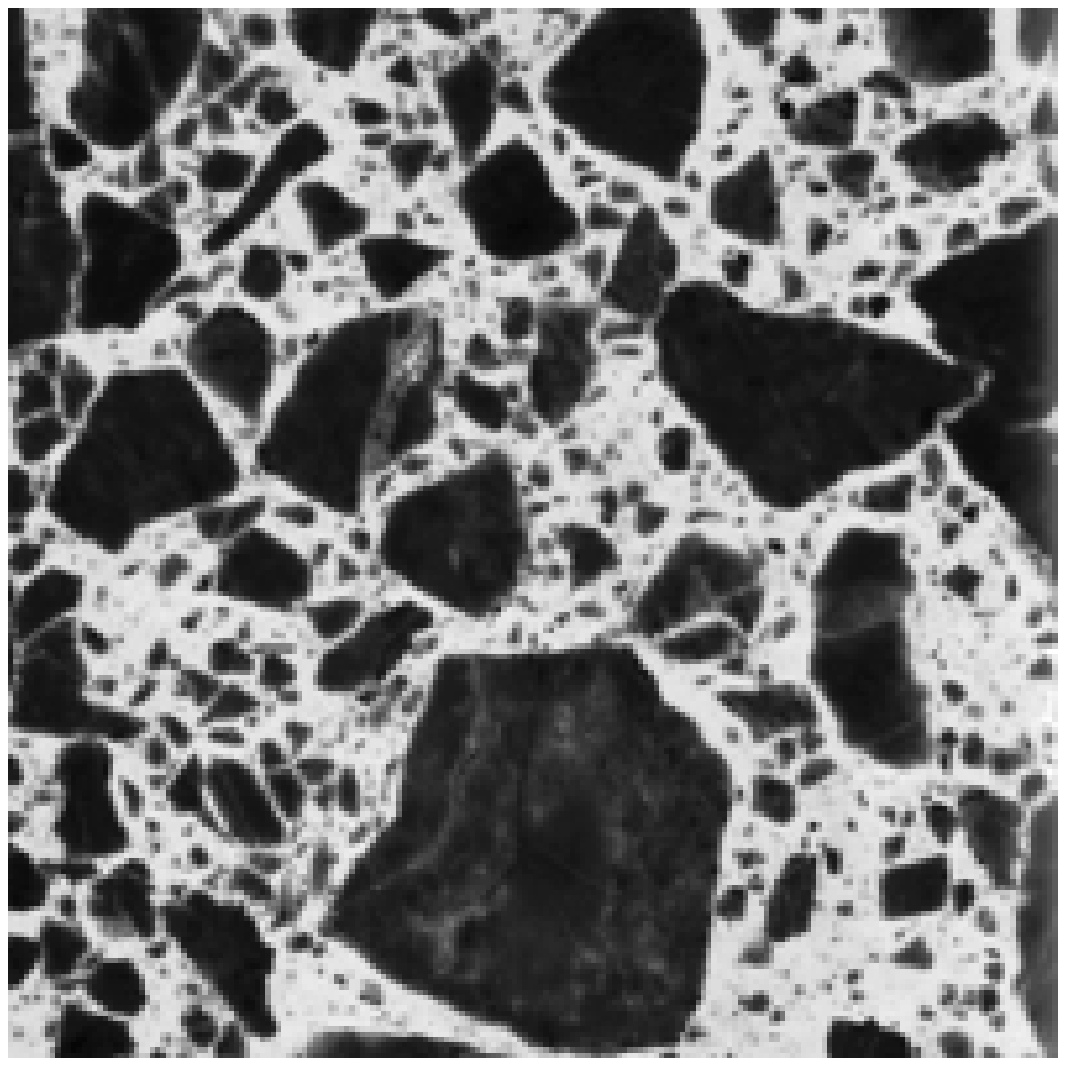}}&
{\includegraphics[width = \figwidths mm,height = \figheights mm]{D99}}\\ 
{(k) 30 vs. 90} & {(l) 59 vs. 61}& {(m) 27 vs. 99}& {(n) 42 vs. 62}& {(o) 31 vs. 99}\\
\end{tabular}
\end{center}
\caption{The most difficult $15$ pairs of texture classes with image IDs selected from Brodatz texture data set.} 
\label{fig:textureexample}
\end{figure}

\newpage

\section{An example of the learned adjustment parameter $\boldsymbol\alpha$ in various classification tasks}
\begin{figure}[!htb]
\begin{center}
{\includegraphics[width = \twomedfigwidths mm]{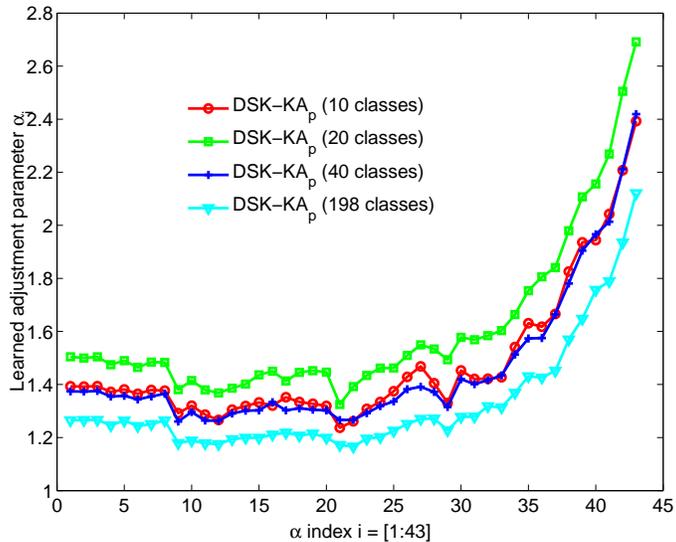}}\\
(a) The learned adjustment parameter $\boldsymbol\alpha$ when it is used as power.\\
{\includegraphics[width = \twomedfigwidths mm]{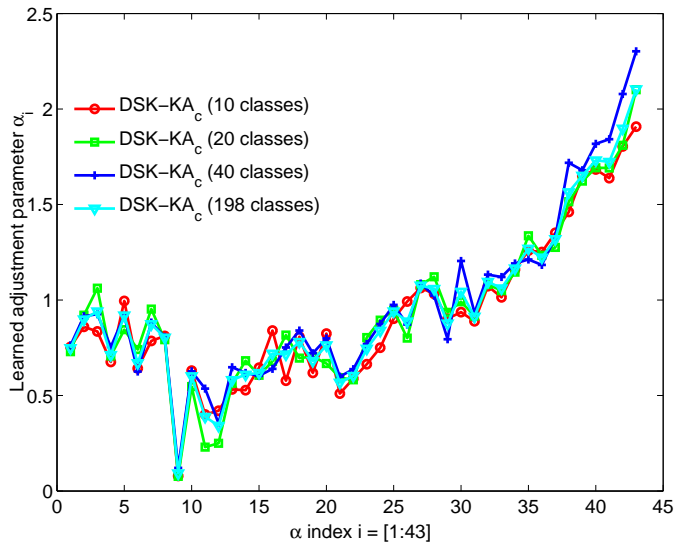}} \\
(b) The learned adjustment parameter $\boldsymbol\alpha$ when it is used as coefficient.
\end{center}
\caption{The learned adjustment parameter $\boldsymbol\alpha$ for the classification tasks involving $10$, $20$, $40$ and all $198$ classes from  'b' subset of FERET data set. The kernel alignment framework is used as an example.}
\label{fig:alpha}
\end{figure}

We compare the learned adjustment parameter $\boldsymbol\alpha$ in various classification tasks.  Fig.~\ref{fig:alpha}(a) plots the learned adjustment parameter $\boldsymbol\alpha$ when it is used as power, and Fig.~\ref{fig:alpha}(b) shows $\boldsymbol\alpha$ when it is used as coefficient. As seen, the components of $\boldsymbol\alpha$ learned in these tasks are very similar to each other, although the number of classes varies. This result suggests that the learned parameters are consistent across various tasks on the same data set. It indicates that the learned adjustment parameters may be able to capture the underlying discrimination.

\newpage

\section{Illustration of the rs-fMRI images and the constructed correlation matrices.}
\begin{figure}[!htb]
\begin{center}
{\includegraphics[width = \fMRItwofigwidths mm,height = \fMRIfigheights mm]{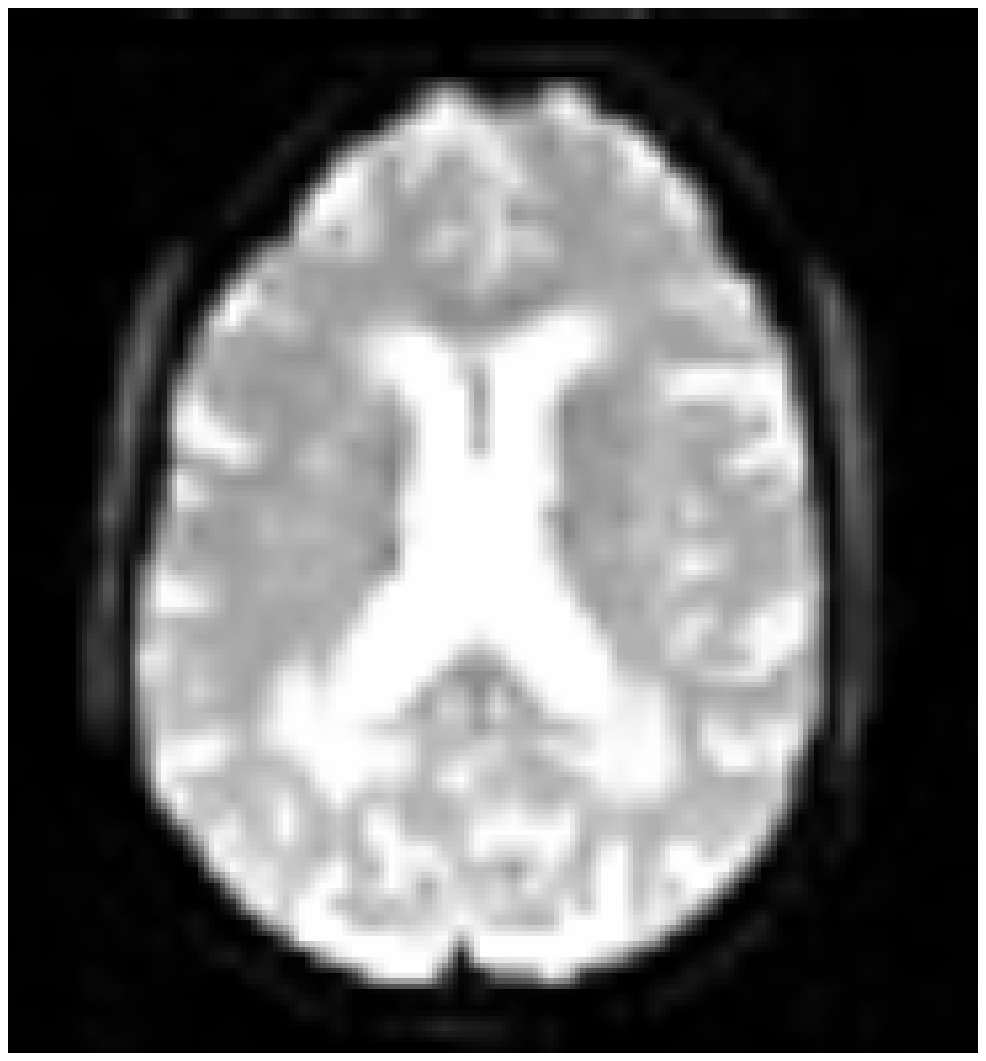}}
{\includegraphics[width = \fMRItwofigwidths mm,height = \fMRIfigheights mm]{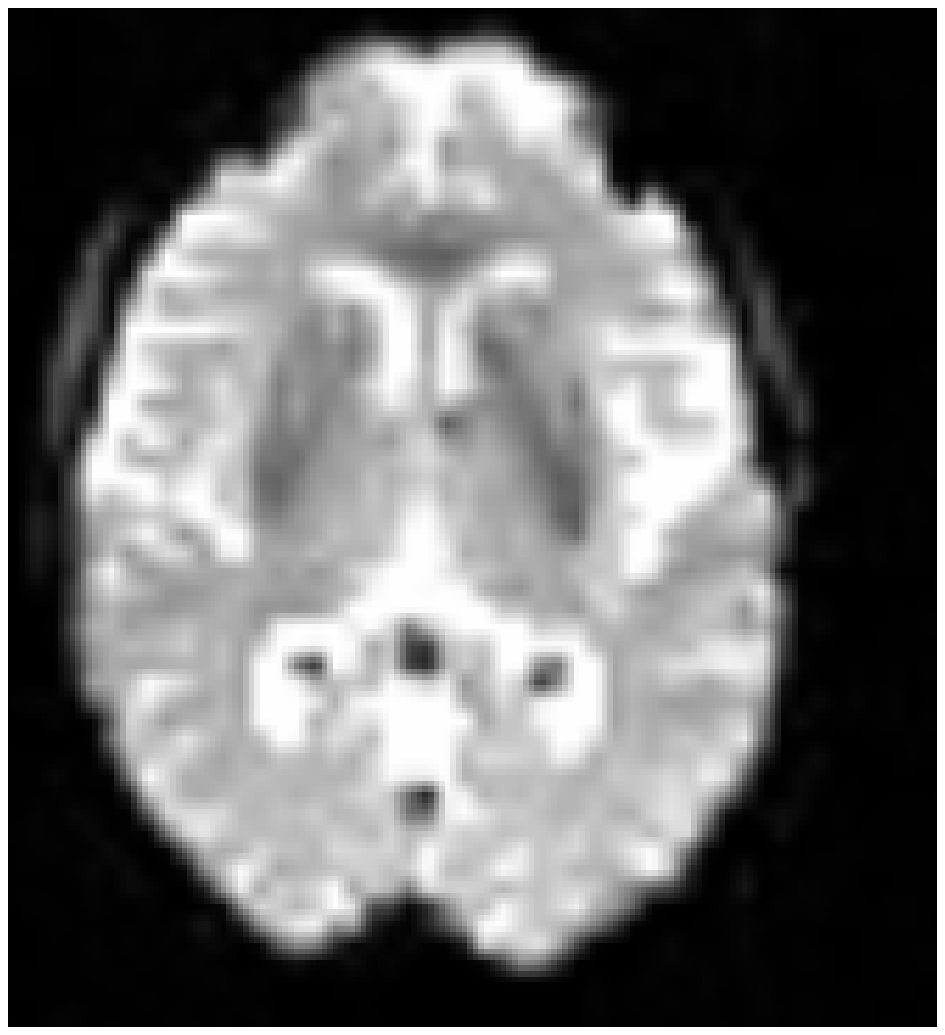}}
{\includegraphics[width = \fMRItwofigwidths mm,height = \fMRIfigheights mm]{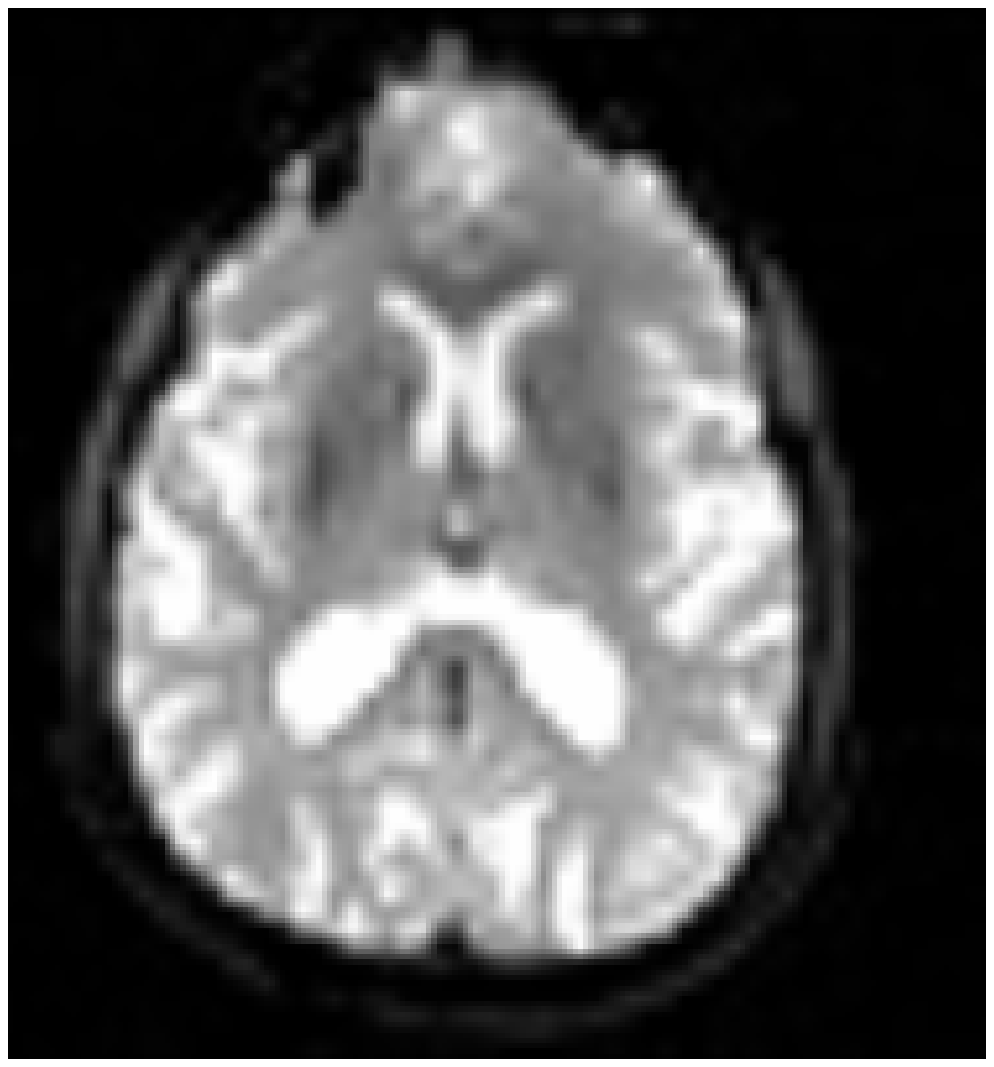}}
{\includegraphics[width = \fMRItwofigwidths mm,height = \fMRIfigheights mm]{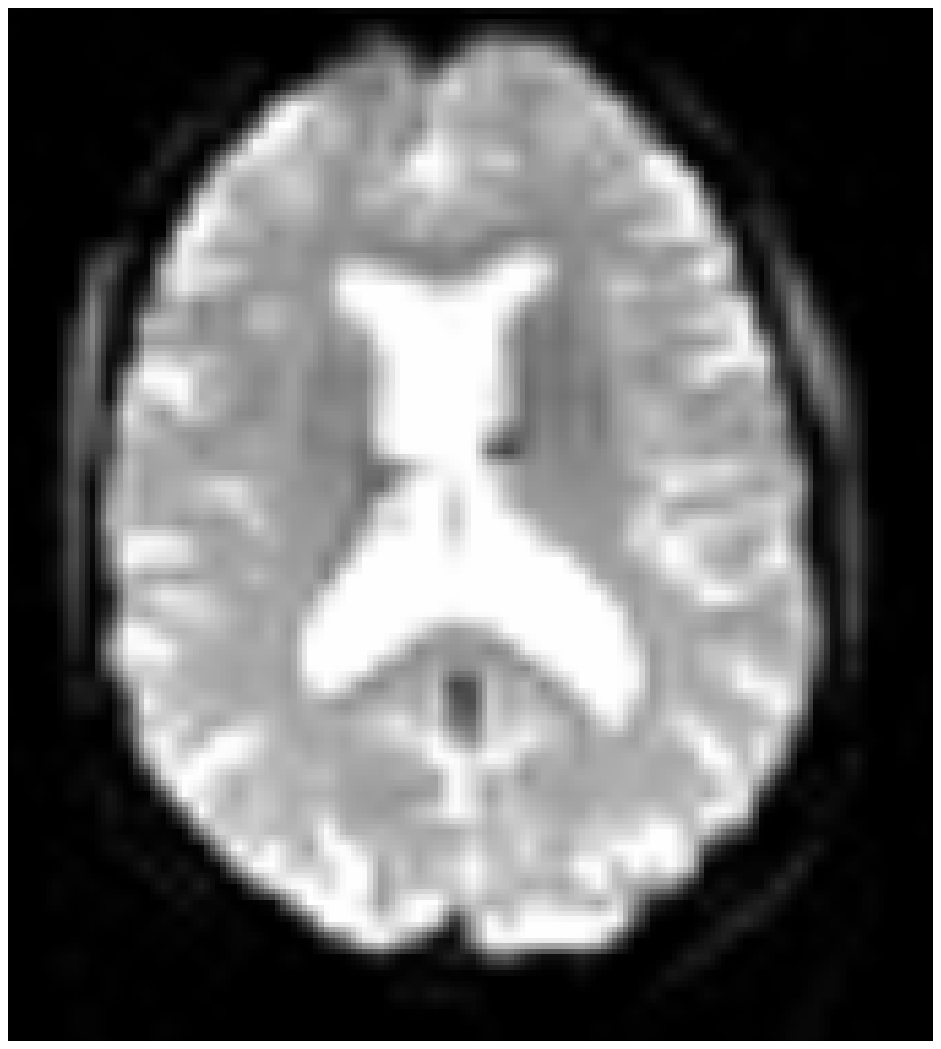}}\\
{\includegraphics[width = \fMRItwofigwidths mm,height = \fMRIfigheights mm]{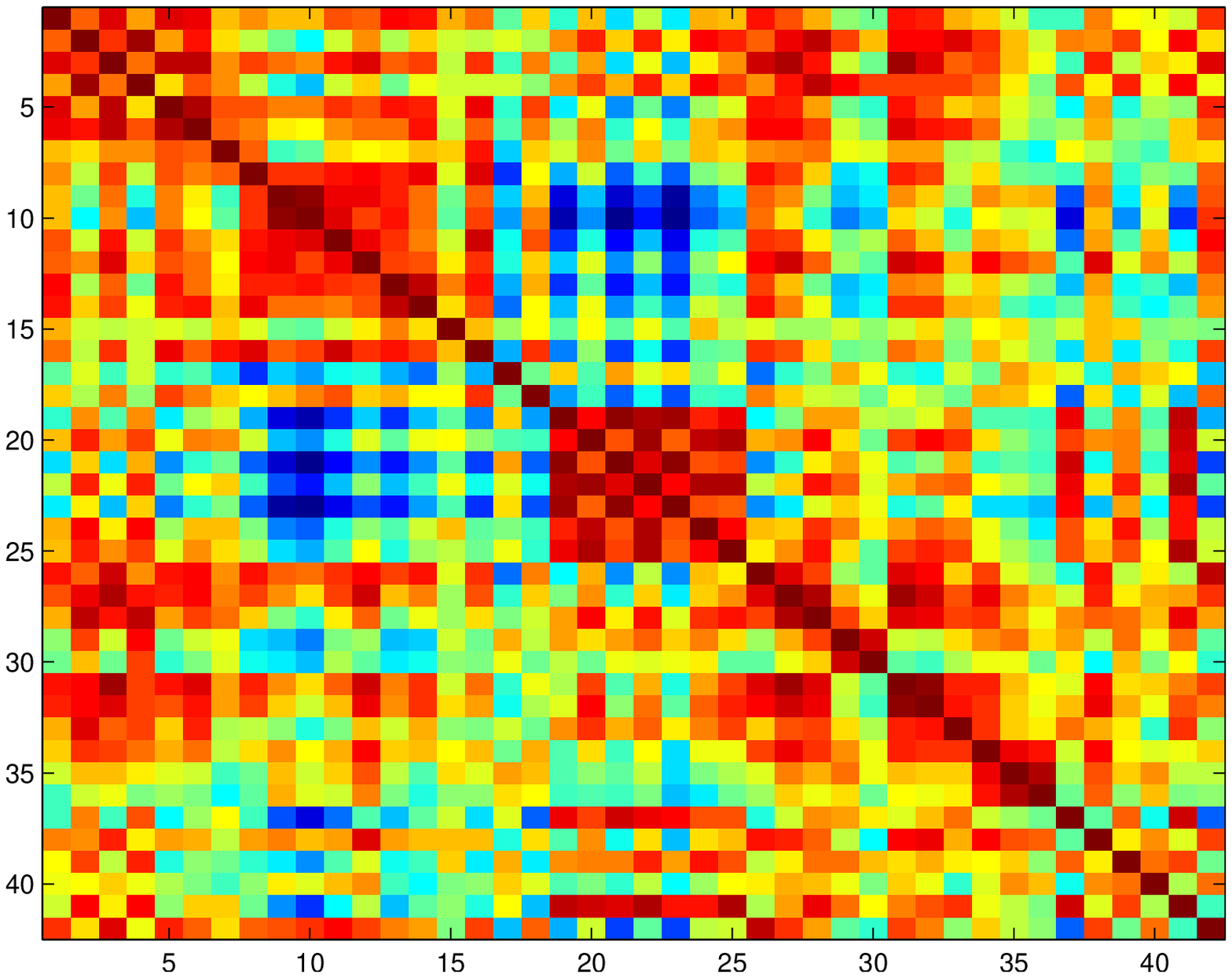}}
{\includegraphics[width = \fMRItwofigwidths mm,height = \fMRIfigheights mm]{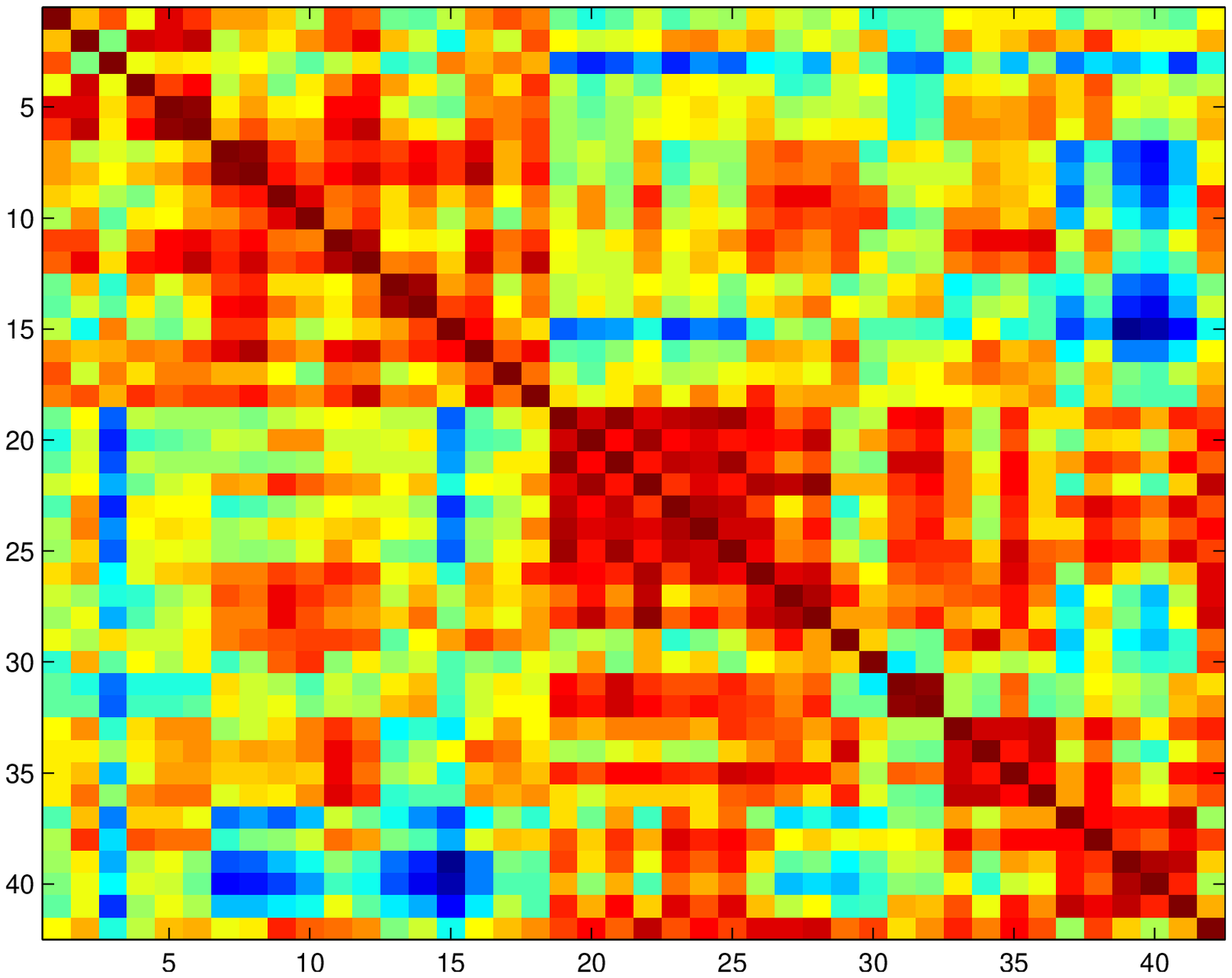}}
{\includegraphics[width = \fMRItwofigwidths mm,height = \fMRIfigheights mm]{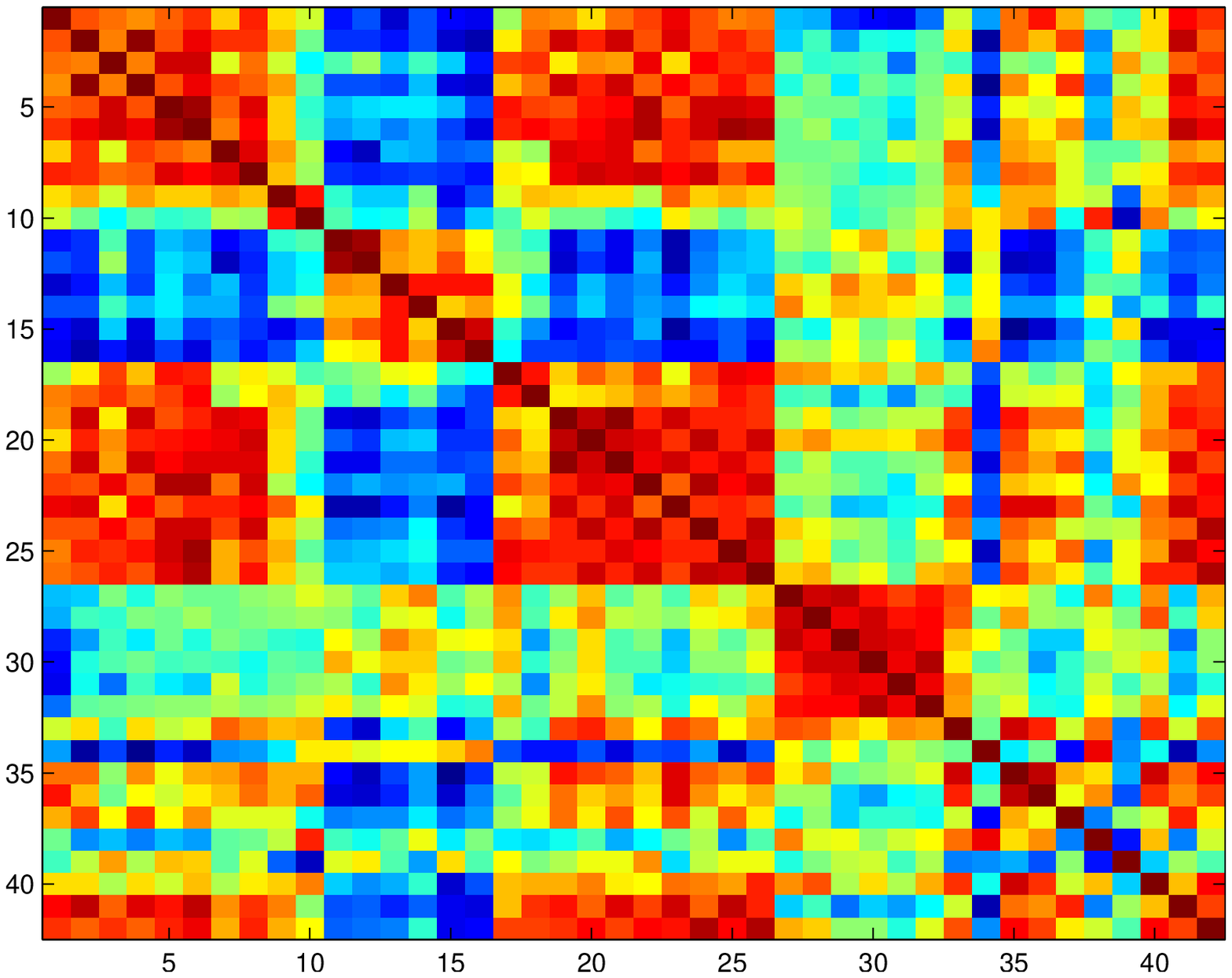}}
{\includegraphics[width = \fMRItwofigwidths mm,height = \fMRIfigheights mm]{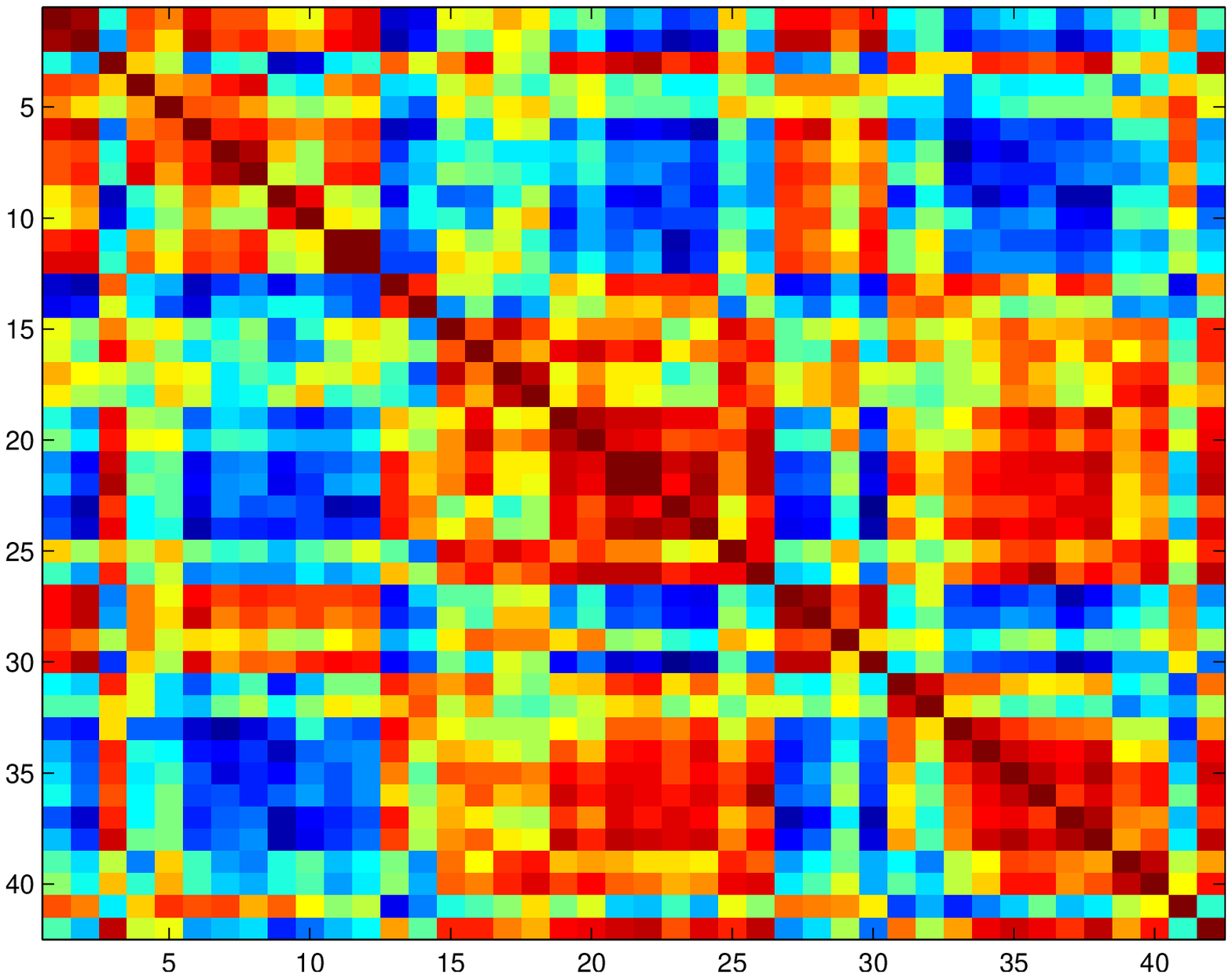}}
\end{center}
\caption{Illustration of rs-fMRI images (the top row) and the correlation matrices (the bottom row) as representations of brain networks.}
\label{fig:exfmri}
\end{figure}

{Fig.}~\ref{fig:exfmri} shows the illustration of rs-fMRI images (the top row) and the correlation matrices (the bottom row) as representations of brain networks.

\newpage

\section{Evolution of various criteria in the optimization of an example task on the Brodatz data set}
\begin{figure*}[!htb]
\begin{center}
\begin{tabular}{cc}
{\includegraphics[width = \fourfigwidth mm]{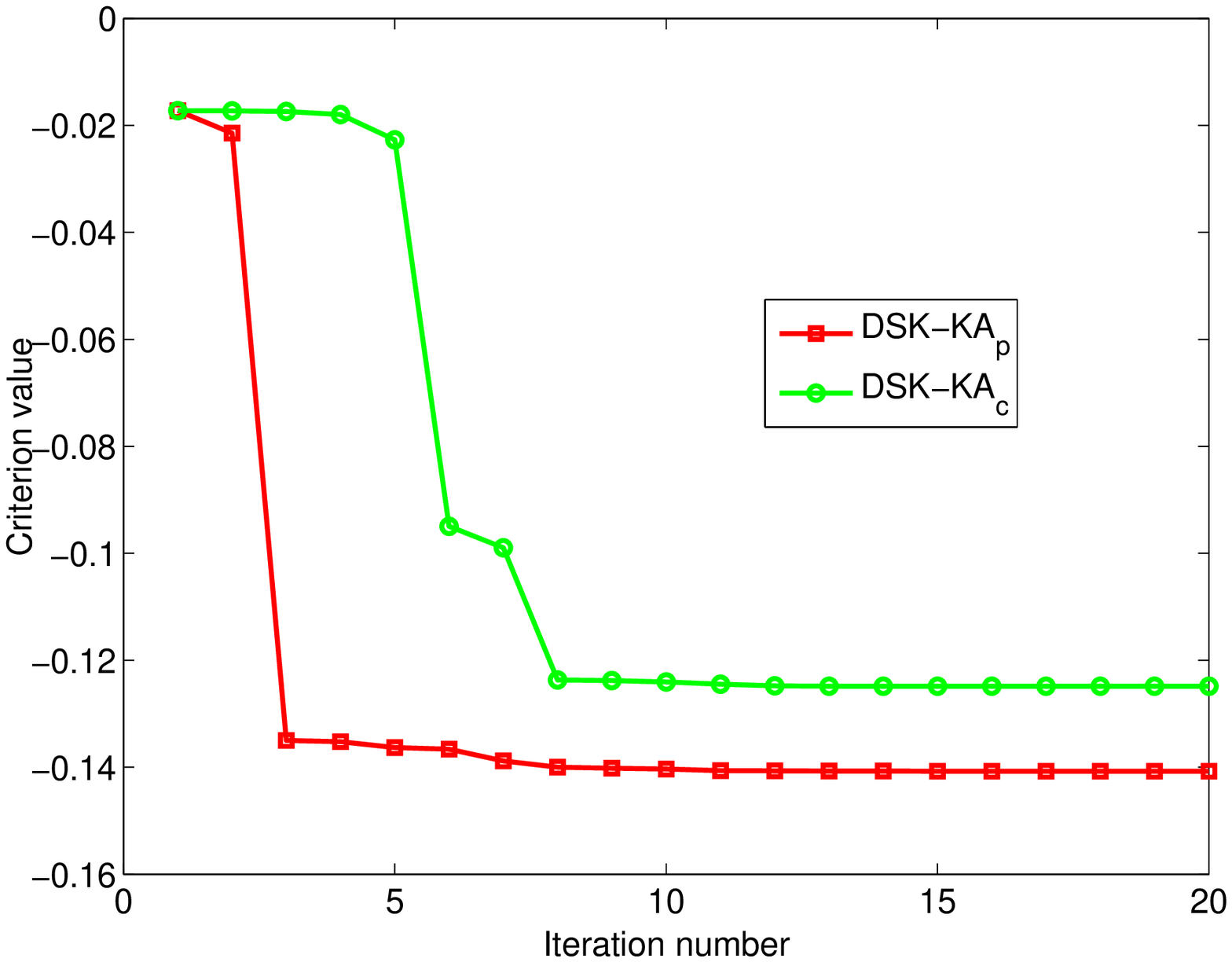}} &
{\includegraphics[width = \fourfigwidth mm]{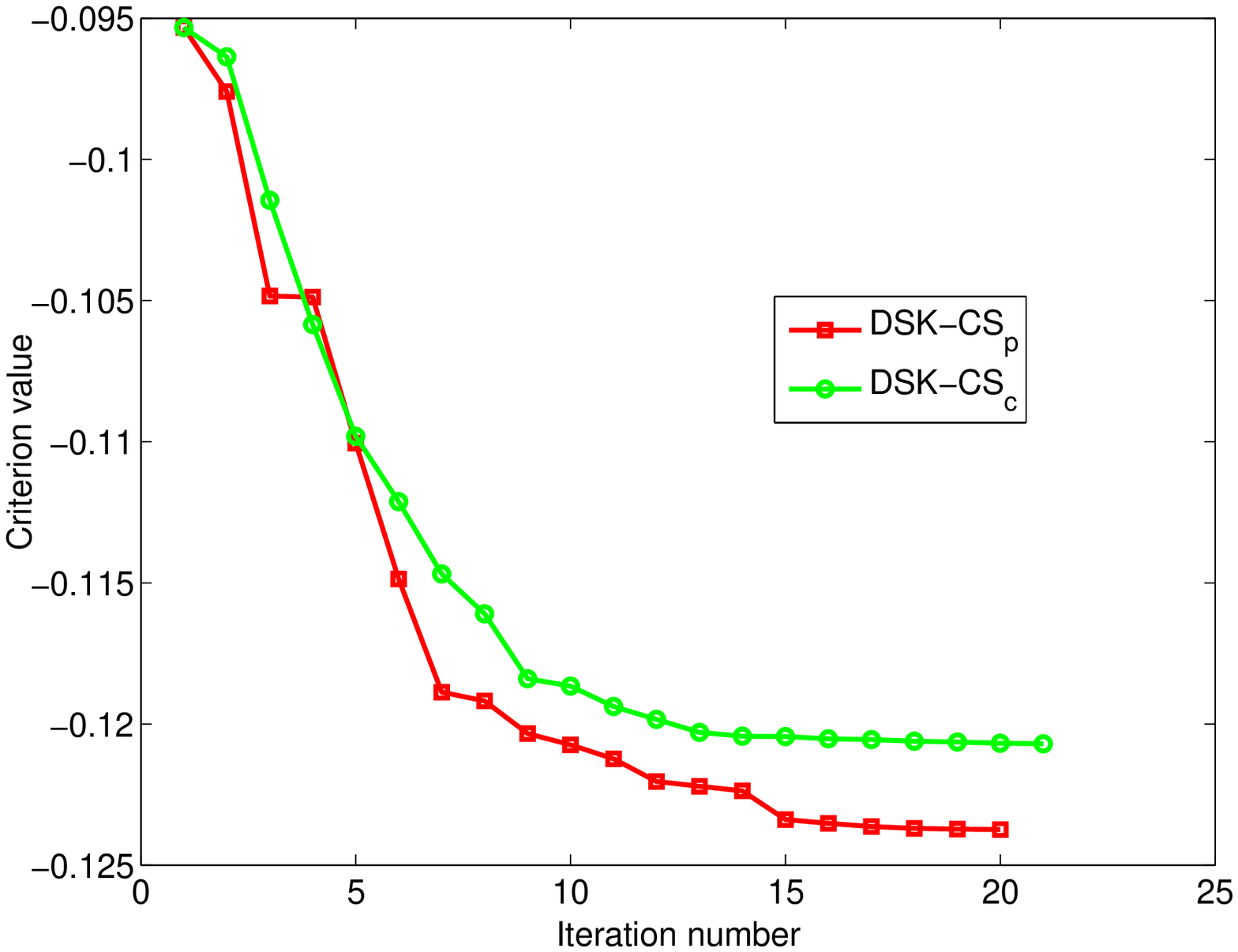}}\\
(a) kernel alignment & (b) Class separability\\
{\includegraphics[width = \fourfigwidth mm]{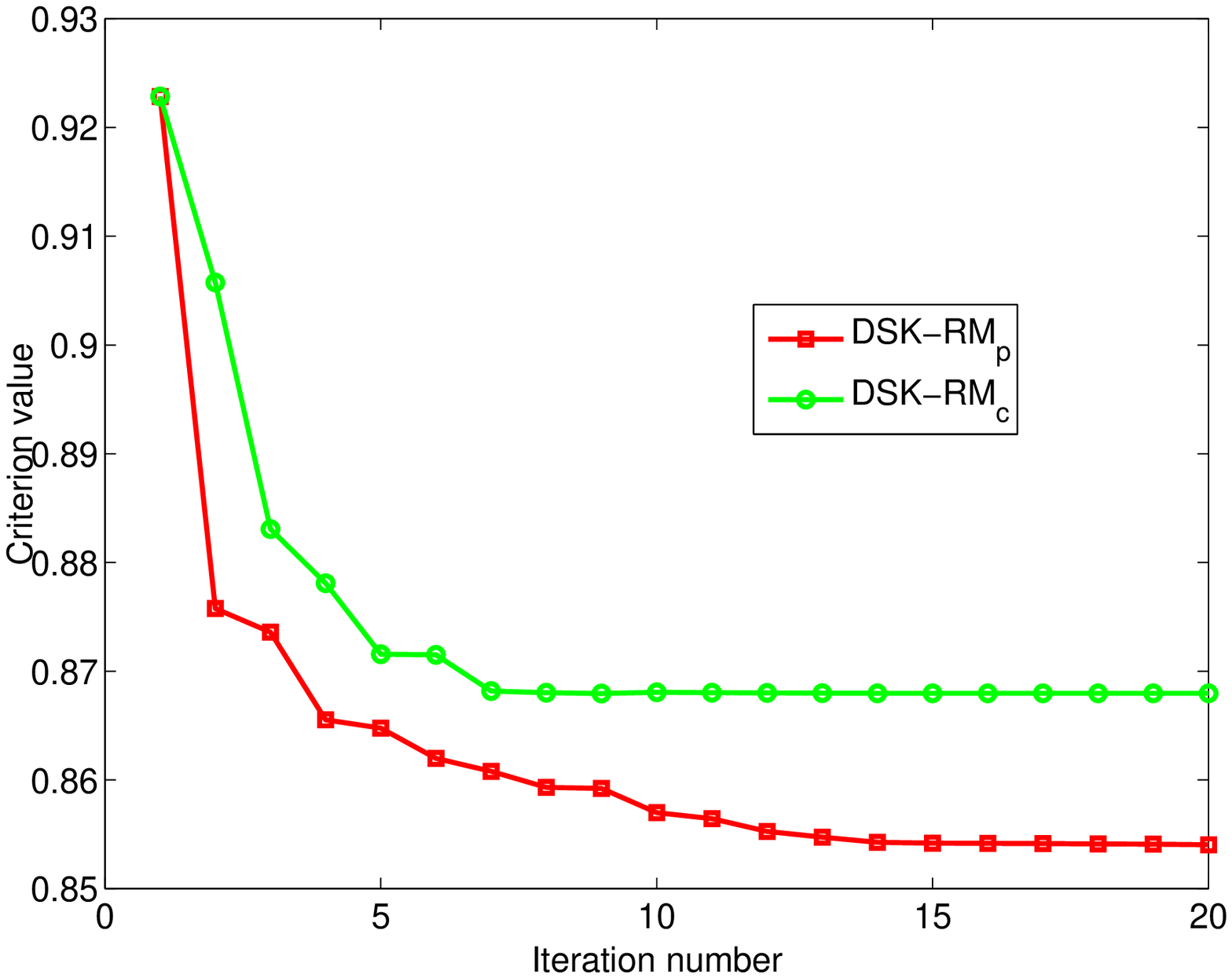}} &
{\includegraphics[width = \fourfigwidth mm]{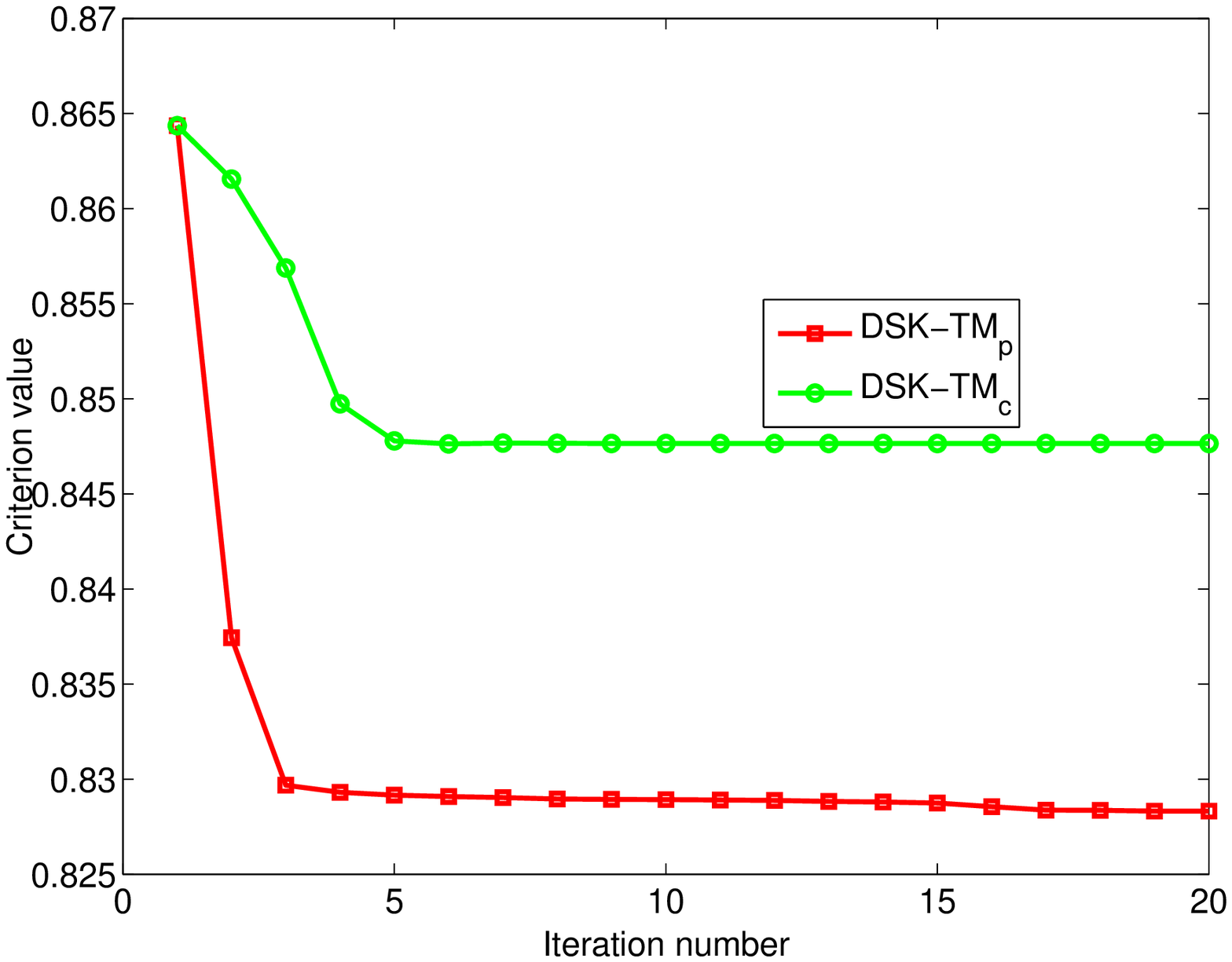}}\\
(c) Radius margin bound  & (d) Trace margin criterion\\
\end{tabular}
\end{center}
\caption{Evolution of various criteria in the optimization of an example task on the Brodatz data set. (a) Kernel alignment. (b) Class separability. (c) Radius margin bound. (d) Trace margin criterion}
\label{fig:optcre}
\end{figure*}

Fig.~\ref{fig:optcre} shows an example of the evolution of various criteria on the Brodatz data set. As shown, DSK using the kernel alignment (Fig.~\ref{fig:optcre}(a)), the radius margin bound (Fig.~\ref{fig:optcre}(c)) or the trace margin criterion (Fig.~\ref{fig:optcre}(d)) needs no more than $10$ iterations to converge, and DSK using the class separability (Fig.~\ref{fig:optcre}(b)) criterion converges in $15$ iterations. 
\newpage

\section{Proof of the affine invariance of the S-divergence }

\begin{equation}  
\begin{split}
 \text{Given}~~ S({\bm X},{\bm Y}) &= \log\left(\det\left(\frac{{\bm X} + {\bm Y}}{2}\right)\right) - \frac{1}{2} \log\left(\det({\bm X}{\bm Y})\right)\\
 \text{For an invertible matrix}~~ &\bm W : \\
 S({{\bm W}^{\top} \bm X \bm W},{{\bm W}^{\top} \bm Y \bm W}) &= \log\left(\det \left({\bm W}^{\top}\left(\frac{{\bm X} + {\bm Y}}{2}\right){\bm W}\right)\right) - \frac{1}{2} \log\left(\det\left({\bm W}^{\top}{\bm X}{\bm W}{\bm W}^{\top}{\bm Y}{\bm W}\right)\right)\\
 &= 2\log\left(\det \left(\bm W\right)\right) + \log\left(\det\left(\frac{{\bm X} + {\bm Y}}{2}\right)\right) - \frac{1}{2}*4 \log\left(\det(\bm W)\right) - \frac{1}{2} \log\left(\det({\bm X}{\bm Y})\right)\\
 &= \log\left(\det\left(\frac{{\bm X} + {\bm Y}}{2}\right)\right) - \frac{1}{2} \log\left(\det({\bm X}{\bm Y})\right)\\
 &=S({\bm X},{\bm Y})
\end{split}
\end{equation}$\hfill{} \blacksquare$

Note that a SPD matrix can be eigen-decomposed by $\bm{X} = \bm{U}_{\bm{X}}\bm{\Lambda}_{\bm{X}} \bm{U}_{\bm{X}}^\top$. In this case, if ${\bm W}^{\top}$ is applied to adjust the eigenvectors $\bm{U}_{\bm{X}}$ by ${\bm W}^{\top}\bm{U}_{\bm{X}}$, the S-divergence between the adjusted SPD matrices will remain the same, because $S\left({\left({\bm W}^{\top}\bm{U}_{\bm{X}}\right)}\bm{\Lambda}_{\bm{X}} {\left({\bm W}^{\top}\bm{U}_{\bm{X}}\right)}^{\top},{\left({\bm W}^{\top}\bm{U}_{\bm{Y}}\right)}\bm{\Lambda}_{\bm{Y}} {\left({\bm W}^{\top}\bm{U}_{\bm{Y}}\right)}^{\top}\right) = S({{\bm W}^{\top} \bm X {\bm W}},{{\bm W}^{\top} \bm Y {\bm W}})=S({\bm X},{\bm Y})$. Therefore, for Stein kernel, adjusting the eigenvalues only (rather than including the matrix of eigenvectors) may have been sufficient.

\end{document}